\documentclass[10pt,twocolumn,letterpaper]{article}

\usepackage{cvpr}              % To produce the CAMERA-READY version
\definecolor{cvprblue}{rgb}{0.21,0.49,0.74}
\usepackage[pagebackref,breaklinks,colorlinks,allcolors=cvprblue]{hyperref}
\usepackage{colortbl}  % For coloring table cells
\usepackage{xcolor}    % For color definitions
\usepackage{amsmath}   % For math symbols
\usepackage{graphicx}
\usepackage{booktabs}
\usepackage{etoc}
\usepackage{titletoc}
\usepackage{float}

\usepackage[most]{tcolorbox}
\def\paperID{14422} % *** Enter the Paper ID here
\def\confName{CVPR}
\def\confYear{2025}

\title{Learning from Massive Human Videos for Universal Humanoid Pose Control}

\author{
    Jiageng Mao$^{1*}$ \quad
    Siheng Zhao$^{1*}$ \quad
    Siqi Song$^{1*\dag}$ \quad
    Tianheng Shi$^{1}$ \quad
    Junjie Ye$^{1}$ \quad
    Mingtong Zhang$^{1}$ \\ 
    Haoran Geng$^{2}$ \quad
    Jitendra Malik$^{2}$ \quad
    Vitor Guizilini$^{3}$ \quad
    Yue Wang$^{1}$\\
    \\
    $^{1}$University of Southern California \quad $^{2}$UC Berkeley \quad
    $^{3}$Toyota Research Institute \\
    \href{https://usc-gvl.github.io/UH-1/}{ \tt\small https://usc-gvl.github.io/UH-1}
}

\begin{document}

\twocolumn[{
\renewcommand\twocolumn[1][]{#1}
\maketitle
\vspace*{-0.25in}
\centering
\captionsetup{type=figure}\includegraphics[width=\textwidth]{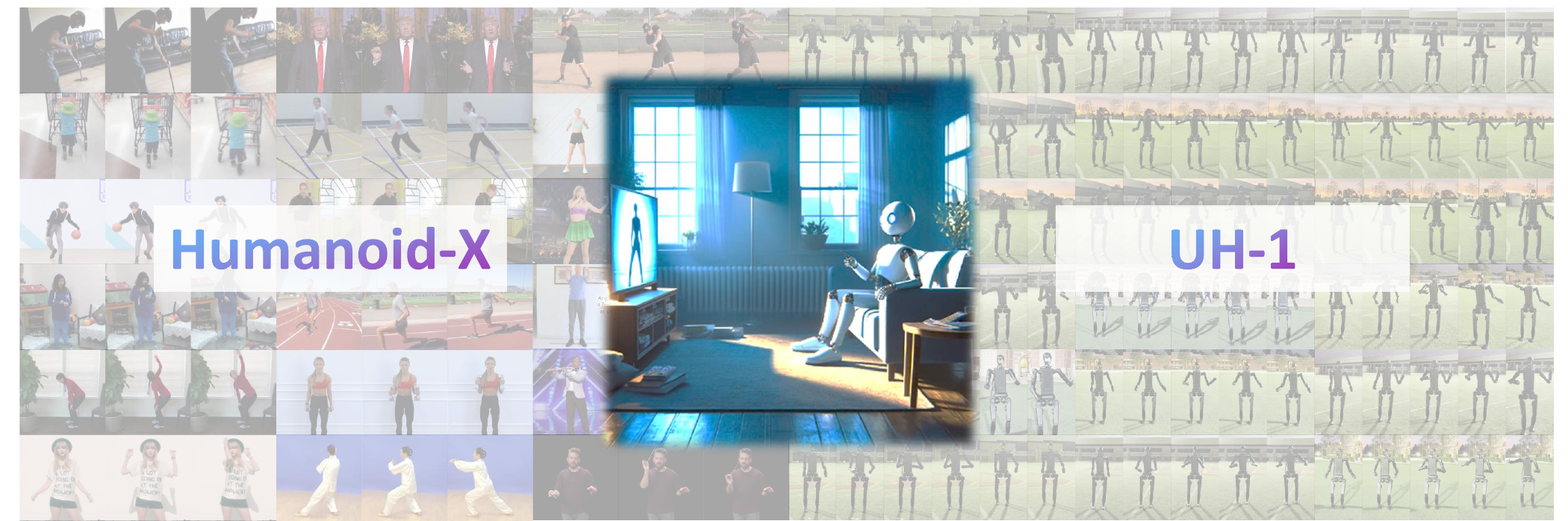}
\vspace*{-7mm}
\captionof{figure}{\textbf{Overview.} We introduce \emph{Humanoid-X}, a large-scale dataset to facilitate humanoid robot learning from massive human videos. On top of \emph{Humanoid-X}, we introduce \emph{UH-1}, a large humanoid model for universal language-conditioned pose control of humanoid robots.}
\label{fig:teaser}
\vspace{4mm}
}]

\maketitle

\renewcommand{\thefootnote}{\fnsymbol{footnote}}
\footnotetext[0]{\textsuperscript{*}equal contribution (in alphabetical order). \textsuperscript{\dag}work done while at USC.}
% \begin{figure*}[!ht]
%   \centering
%    \includegraphics[width=0.98\linewidth]{fig/teaser_v3.pdf}
%    \caption{Learning from Massive Human Videos for Universal Humanoid Robot Control.}
%    \label{fig:teaser}
% \end{figure*}

\begin{abstract}

Scalable learning of humanoid robots is crucial for their deployment in real-world applications. 
While traditional approaches primarily rely on reinforcement learning or teleoperation to achieve whole-body control, they are often limited by the diversity of simulated environments and the high costs of demonstration collection. In contrast, human videos are ubiquitous and present an untapped source of semantic and motion information that could significantly enhance the generalization capabilities of humanoid robots. This paper introduces Humanoid-X, a large-scale dataset of over 20 million humanoid robot poses with corresponding text-based motion descriptions, designed to leverage this abundant data. Humanoid-X is curated through a comprehensive pipeline: data mining from the Internet, video caption generation, motion retargeting of humans to humanoid robots, and policy learning for real-world deployment. With Humanoid-X, we further train a large humanoid model, UH-1, which takes text instructions as input and outputs corresponding actions to control a humanoid robot. Extensive simulated and real-world experiments validate that our scalable training approach leads to superior generalization in text-based humanoid control, marking a significant step toward adaptable, real-world-ready humanoid robots.

\end{abstract}    
\section{Introduction}
\label{sec:intro}

Scalability is crucial in deep learning. Recent advances in computer vision have demonstrated that scaling up training data leads to more powerful foundation models for visual recognition~\cite{clip, sam, dinov2} and generation~\cite{SD, SVD}. In robotics, researchers follow a similar paradigm and build foundation models for robotic manipulation~\cite{openvla, rt1, rt2, rtx} by collecting massive robotic demonstrations. Nevertheless, in contrast to images and videos that are abundant and easily accessible, collecting large-scale robotic demonstrations is expensive and time-consuming, which limits the scalability of current robot learning methods. This raises the question: \textit{Can we use videos as demonstrations to improve the scalability of robot learning?}

To address this challenge, many efforts have been made, such as learning affordances~\cite{vrb, ram, egoexo4d}, flows~\cite{imflow2act, genflow}, and world models~\cite{unisim} from natural videos, which enable more generalizable robotic manipulation. However, when it comes to humanoid robots, learning such action representations from videos remains an open problem. Unlike robotic arms, humanoid robots have distinct kinematic structures and more degrees of freedom (DoFs), making them harder to control. Existing works~\cite{rl-humanoid-locomotion, rl-humanoid-jump, humanoid-gym, humanoid-worldmodel, humanoid-lipschitz, expressive, humanoid-loco-terrain} leverage large-scale reinforcement learning to learn robust humanoid control policies, but they only focus on limited robotic skills such as locomotion or jumping, making them less generalizable for handling everyday tasks. Other works~\cite{omnih2o, humanplus, h2o, humanoid-teleop} control humanoid robots through teleoperation, but they require human labor to collect robotic data, which is less scalable. In contrast to these previous works, learning a universal action representation from massive videos will greatly improve the scalability of humanoid robot learning and enable more generalizable humanoid pose control.

To bridge this gap in humanoid robot learning, we introduce Humanoid-X, a large-scale dataset curated from a massive and diverse collection of videos for universal humanoid pose control. Humanoid-X utilizes natural language as an interface to connect human commands and humanoid actions, so humans can talk to their humanoid robots to control their actions. The natural language representations are extracted from videos via captioning tools and are used to describe the actions of humanoid robots. For action representations, Humanoid-X leverages both robotic keypoints for high-level control and robotic target DoF positions for direct position control. To extract humanoid actions from human videos, we first reconstruct 3D humans and their motions from videos. Then, we leverage motion retargeting to transfer motions from 3D humans to humanoid robots, resulting in robotic keypoints for high-level humanoid pose control. Finally, we learn a universal RL-based control policy that maps keypoints to low-level humanoid target DoF positions that can be deployed in real robots. We collect over 160,000 human-centric videos from academic datasets and the Internet, covering diverse action categories. We further transform these videos into text-action pairs, resulting in over 20 million humanoid actions with corresponding text descriptions. Humanoid-X paves the way for developing more generalizable and scalable humanoid robotic control guided by natural language.

On top of the Humanoid-X dataset, we further investigate how to learn a universal humanoid pose control model using large-scale text-action pairs. We introduce \textbf{U}niversal \textbf{H}umanoid-1 (UH-1), a large humanoid model for universal language-conditioned humanoid pose control. UH-1 leverages the scalability of the Transformer architecture to handle vast amounts of data efficiently. We begin by discretizing 20 million humanoid actions into action tokens, creating a vocabulary of motion primitives. Then, given a text command as input, the Transformer model auto-regressively decodes a sequence of these tokenized humanoid robotic actions. For cases where the action representation involves robotic keypoints, we transform these into robotic DoF positions using an additional action decoder. Finally, we utilize a proportional-derivative (PD) controller to convert the DoF positions into motor torques, enabling us to control humanoid robots and deploy them in the real-world.

% To validate the effectiveness of the Humanoid-X dataset and the UH-1 model, we conducted extensive experiments using both simulators and real humanoid robots. The experimental results demonstrate that scaling up training data collected from massive videos allows our model to successfully ground text into humanoid actions and generate diverse humanoid movements based on textual commands. Moreover, our model maintains robustness and can be reliably deployed in a real humanoid robot.
To validate the effectiveness of the Humanoid-X dataset and the UH-1 model, we conducted extensive experiments across both simulated and real humanoid platforms. Our results reveal that leveraging vast amounts of video data enables our model to seamlessly translate textual commands into diverse and contextually accurate humanoid actions. Notably, the UH-1 model demonstrates strong robustness, proving reliable in real-world deployment.
To summarize, our key contributions are as follows:

$\cdot$ We introduce Humanoid-X, a pioneering large-scale dataset tailored for learning universal humanoid control from massive Internet video data. 

$\cdot$ We introduce UH-1, a powerful, scalable model for language-conditioned control of humanoid poses. Our approach supports two flexible control modes that are interchangeable, depending on task requirements. We also provide extensive ablation study for our design choices.  

$\cdot$ Our experiments confirm that training on massive video data enables a level of generalizability in humanoid control that was previously unattainable. 
% \yue{add a reference to our figure/table?}

\section{Related Works}
\label{sec:relatedworks}

\begin{figure*}
  \centering
   \includegraphics[width=1.0\linewidth]{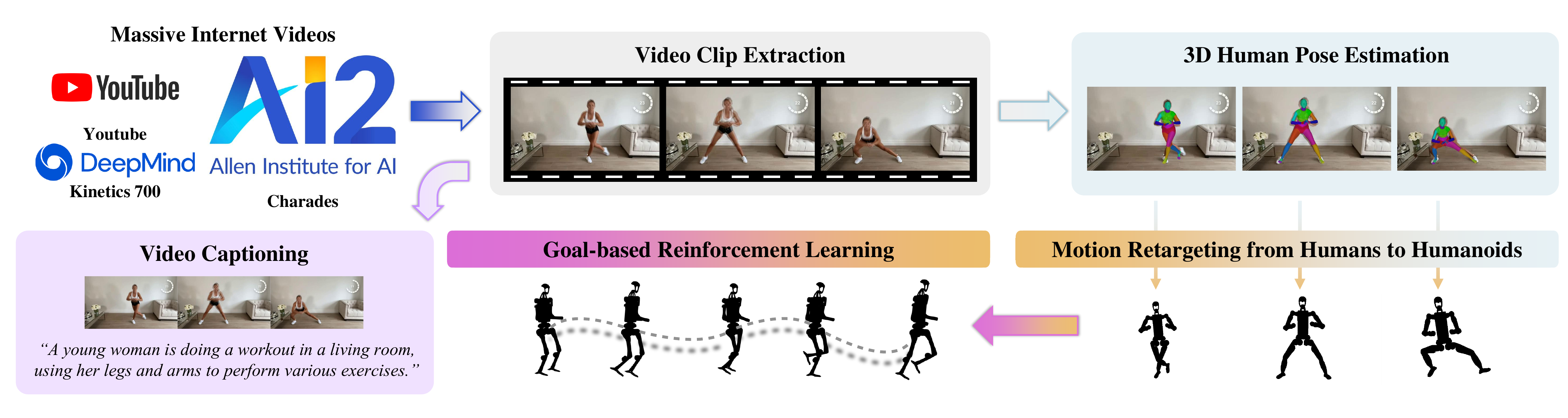}
   \vspace{-7mm}
   \caption{\textbf{Learning Humanoid Pose Control from Massive Videos}. We mine massive human-centric video clips $\mathcal{V}$ from the Internet. We then extract text-based action descriptions $\mathcal{T}$ and 3D human poses $\mathcal{P}_{human}$ from the video clips. Next, we retarget the motions from humans to humanoid robots, resulting in humanoid keypoints $\mathcal{P}_{robot}$ for high-level control. Finally, we employ reinforcement learning to generate physically deployable humanoid actions $\mathcal{A}_{robot}$. In this manner, we collect 163,800 pairs of motion samples $\langle \mathcal{V}, \mathcal{T}, \mathcal{P}_{human}, \mathcal{P}_{robot}, \mathcal{A}_{robot} \rangle$ from Internet videos, which are leveraged to distill a universal humanoid pose control policy.}
   \label{fig:method}
   \vspace{-5mm}
\end{figure*}

\quad\textbf{Robot Learning from Internet Data.} Many endeavors have been made to learn scalable robot learning policies from non-robotic data, especially Internet videos. The key idea is to learn valuable representations from massive visual data and transfer them to robotic tasks. The learned representations include pre-trained visual features from videos~\cite{r3m, MVP, VIP, rpt} and transferable action representations such as affordances~\cite{vrb, human2robot} and object-centric flows~\cite{imflow2act, genflow}. Other works~\cite{structured-worldmodel, unisim, vlp} attempt to learn world models from Internet videos. However, most of these works focus on robotic manipulation. Since robot arms have totally different kinematic structures from humanoid robots, the learned visual and action representations for robotic manipulation are not transferable to humanoid robot control.  In contrast, we investigate how to learn universal pose control for humanoid robots from massive videos.

\textbf{Humanoid Robot Learning.} Extensive work has been dedicated to learning policies that enable robust control of humanoid robots. Some works focus on humanoid locomotion using large-scale reinforcement learning~\cite{humanoid-gym, humanoid-lipschitz, rl-humanoid-locomotion, rl-humanoid-jump, humanoid-worldmodel} or imitation learning~\cite{humanoid-next-token, humanmimic}. Other works learn humanoid manipulation via imitation learning~\cite{humanoid-okami, humanoid-dp3}. Notably, some works~\cite{h2o, omnih2o, expressive, humanplus, humanoid-hover} learn humanoid teleoperation by transferring motions from 3D humans to humanoid robots. However, these works rely on well-calibrated motion capture data, limiting their generalization ability to unseen motions. In contrast, our method operates as a fully autonomous agent that learns from massive Internet videos and performs generalizable humanoid pose control based on arbitrary text commands.

\textbf{3D Human Motion Generation.} Many works are attempting to generate diverse 3D human motions via Transformers~\cite{t2m-gpt, motiongpt} or diffusion models~\cite{mdm, motiondiffuse, mdm-prior, intergen, omnicontrol}. Also, some works~\cite{physdiff, phys-humanoid, perpetual-huamnoid, ase, amp, calm, scalable-characters, humansin4d} are trying to generate realistic motions to animate physics-based virtual characters. However, humanoid robots are essentially different from digital humans in many aspects: (1) they have different joint structures and degrees of freedom; (2) humanoid robots cannot access privileged information like linear velocities, which is readily available when controlling virtual humans; (3) humanoid robots have physical constraints such as motor torque limits, whereas 3D virtual humans do not have these limitations. An alternative solution for generalizable humanoid pose control is to first generate 3D human motions and then retarget them to humanoid robots~\cite{omnih2o, harmon}. Compared to these approaches, our UH-1 model offers a more streamlined solution by directly mapping text commands into executable humanoid actions without intermediate steps. Furthermore, unlike human motion generation models trained on expensive motion capture data, learning from massive videos significantly enhances the generalization ability of our method.

\section{Humanoid-X Dataset}
\label{sec:dataset}
\begin{figure*}
  \centering
   \includegraphics[width=1.0\linewidth]{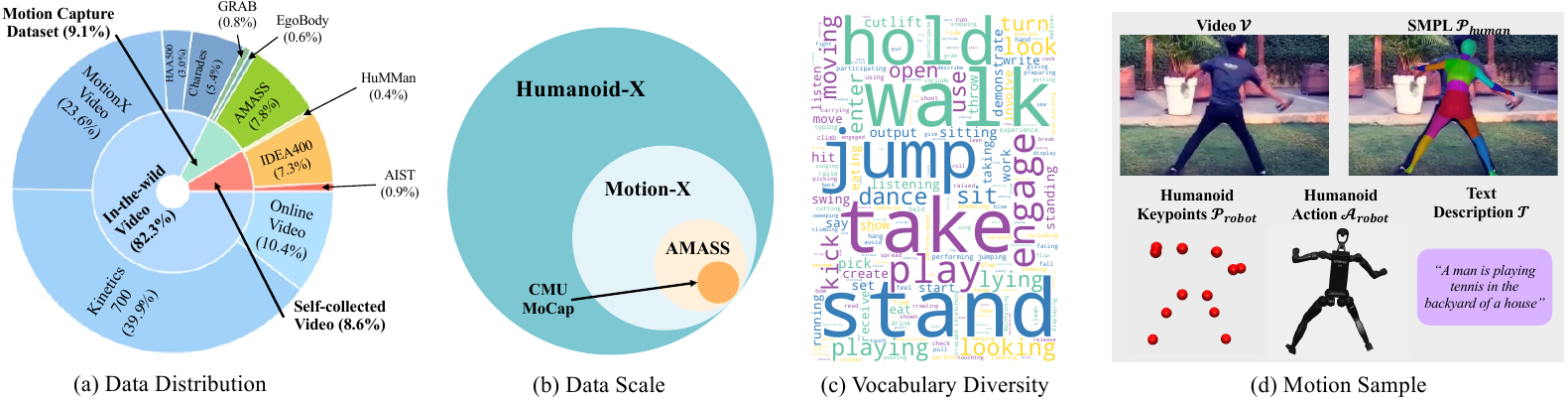}
   \caption{\textbf{Dataset Statistics}. Humanoid-X features extensive scale, diverse sources, a rich action vocabulary, and multiple data modalities.}
   \label{fig:word_cloud}
   \vspace{-5mm}
\end{figure*}

\subsection{Overview}
\label{dataset_overview}
To scale up humanoid robot learning using massive human videos, we introduce Humanoid-X, the largest humanoid robot dataset to date compiled from a vast and diverse collection of videos for universal humanoid pose control. Humanoid-X consists of 163,800 motion samples covering a comprehensive set of action categories. Each motion sample in the dataset contains 5 data modalities: an original video clip $\mathcal{V}$, a text description $\mathcal{T}$ of the action in the video, a sequence of SMPL~\cite{smpl}-based human poses $\mathcal{P}_{human}$ estimated from the video, a sequence of humanoid keypoints $\mathcal{P}_{robot}$ for high-level robotic control, and a sequence of humanoid actions $\mathcal{A}_{robot}$ representing target DoF positions for low-level robotic position control. Humanoid-X encompasses over 20 million frames, totaling approximately 240 hours of data. Beyond its extensive scale across multiple data modalities, which is essential for scalable humanoid policy training, Humanoid-X also features a large and diverse text-based action vocabulary, as shown in \cref{fig:word_cloud} (c). This diversity supports universal and text-conditioned humanoid pose control. In the next section, we will discuss how to obtain these motion samples $\langle \mathcal{V}, \mathcal{T}, \mathcal{P}_{human}, \mathcal{P}_{robot}, \mathcal{A}_{robot} \rangle$ from massive videos.

% integrating five distinct data modalities (text, video clip, SMPL human model, robot pose, robot action) that can be used to train three types of models: text-to-human-motion, text-to-humanoid-pose, and text-to-humanoid-action generation. 
% \ourdataset consists of 163,800 pairs of task descriptions in natural language along with corresponding video clips, SMPL human model sequences, humanoid robot pose trajectories, and humanoid robot rollout (pairs of observation and action) trajectories. 
% As detailed in \cref{tab:dataset_statistics}, \ourdataset encompasses over 20 millions frames, amounting to a total of 240 hours of data.
% Beyond its extensive scale across multiple data modalities, which is essential for humanoid policy training, \ourdataset also features a large and varied natural language vocabulary, as shown in \cref{fig:word_cloud}. This diversity supports a wide range of humanoid pose generation and control tasks, advancing human-humanoid interaction and enhancing text-conditioned model training.

\subsection{Learning from Massive Videos}

% Our dataset is primarily sourced from three types of collections: existing MoCap and non-MoCap SMPL human model datasets (such as AIST~\cite{tsuchida2019aist}, AMASS~\cite{guo2022generating}, EgoBody~\cite{zhang2022egobody}, GRAB~\cite{taheri2020grab}, HAA500~\cite{chung2021haa500}, HuMMan~\cite{cai2022humman}, IDEA400~\cite{lin2024motion}, and MotionX~\cite{lin2024motion}), action recognition video datasets (including Charades~\cite{sigurdsson2016hollywood} and Kinetics700~\cite{carreira2019short}), and a curated set of in-the-wild YouTube videos. 
% To collect these YouTube videos, we designed over 400 unique search terms (details are provided in the supplementary material) covering a range of human activities from daily tasks and fitness routines to professional sports, and then utilize the Google Cloud API\footnote{\href{https://console.cloud.google.com/apis/library/youtube.googleapis.com}{YouTube Data API v3}} to retrieve the top 20 videos for each specified search term.

To process large-scale, in-the-wild raw video data, we developed a fully automated data annotation pipeline comprising five modules, as illustrated in \cref{fig:method}. 
The pipeline includes (1)~a video processing module that mines and extracts video clips $\mathcal{V}$ from noisy Internet videos, (2)~a video captioning model that generates text description of human actions $\mathcal{T}$, (3)~a human pose detection module that estimates parametric 3D human poses $\mathcal{P}_{human}$ from video clips, (4)~a motion retargeting module to generate humanoid robotic keypoints $\mathcal{P}_{robot}$ by transferring motions from humans to humanoid robots, and (5)~a goal-conditioned reinforcement learning policy to learn physically-deployable humanoid actions $\mathcal{A}_{robot}$ by imitating humanoid keypoints. 
% Notably, for data originating from SMPL human model datasets, the initial video clip extraction and 3D human pose detection stages are excluded.

\noindent\textbf{Video Mining and Processing.} The first step of our approach is to collect a large number of human-centric videos that encompass a wide variety of action types. To this end, we mine massive informative video clips from 3 sources: academic datasets for digital human research~\cite{tsuchida2019aist, guo2022generating, zhang2022egobody, taheri2020grab, chung2021haa500, cai2022humman, lin2024motion}, datasets for video action understanding~\cite{sigurdsson2016hollywood, carreira2019short}, and Internet videos from YouTube. To collect Internet videos, we designed over 400 unique search terms covering a range of human activities from daily tasks to professional sports, and then utilized the Google Cloud API\footnote{\href{https://console.cloud.google.com/apis/library/youtube.googleapis.com}{YouTube Data API v3}} to retrieve the top 20 videos for each specified search term.

Original videos are often noisy, including segments with no humans, multiple humans, or a stationary individual, which makes them unsuitable for humanoid control. 
To obtain meaningful video clips, we begin by downsampling each video to a standardized 20 frames per second (FPS) to ensure consistency across the dataset. 
Next, we employ an object detector~\cite{yolov8} for single-human detection, selecting frames with precisely one visible person. 
Following detection, we apply motion detection by calculating the pixel-wise grayscale difference between consecutive frames to keep frames showing significant movement. 
We then compile sequences of at least 64 consecutive frames that satisfy the above single-human motion criterion into video clips, resulting in 163,800 video clips $\mathcal{V}$ in total.

\noindent\textbf{Video captioning.}
Language bridges human commands and humanoid actions. To associate humanoid actions with semantic meaning and enable language-conditioned humanoid control, we employ a video captioning model~\cite{cheng2024videollama} to generate fine-grained action descriptions $\mathcal{T}$ from videos:
\begin{equation}
    \mathcal{T} = F_{caption}(\mathcal{V}),
\end{equation}
where $F_{caption}$ is the video captioning model.
To avoid irrelevant text descriptions, we carefully design prompts to guide the model to describe human actions instead of physical appearance, resulting in action-centric text descriptions.

\noindent\textbf{3D Human Pose Estimation.} 
Humanoid robots inherently share a similar skeleton with humans, which allows for learning control policies for humanoid robots based on human motion data. To this end, we first need to extract human poses from videos. To accurately track and estimate human poses in video clips, we adopt a video-based 3D human parametric model estimator~\cite{kocabas2020vibe}, which estimates SMPL~\cite{smpl}-based humans and camera parameters for each frame. We further extract global human motions, \textit{i.e.}, root translations, using the estimated camera parameters. The process can be formulated as:
\begin{equation}
    \mathcal{P}_{human}(\beta, \theta, t_{root}) = F_{pose}(\mathcal{V}),
\end{equation}
where $F_{pose}$ is the human pose estimation model.
Finally, we obtain per-frame 3D human pose: $\mathcal{P}_{human}(\beta, \theta, t_{root})$, where $\beta$ controls the human shapes, $\theta$ controls the joint rotations, and $t_{root}$ controls the global root translations.  
% However, \cite{kocabas2020vibe} does not provide global human motions, \textit{i.e.}, root translations, which is essential to control humanoid robots. 
% To bridge the gap and estimate root translations in the 3D space, we first convert the weak perspective camera parameters from cropped image coordinates back to the original image coordinates, which are represented by the camera scaling factor $s$ and position $p=[p_x, p_y]$. 
% The 3D root translation $T$ in camera space is then computed as:
% \begin{equation}
%  T = [p_x, p_y, \frac{2f}{W \cdot s}]
%   \label{eq:3.1}
% \end{equation}
% where $f$ is the focal length and $W$ represents the image width.

% \paragraph{Video Captioning} 
% Effective training of a language-conditioned humanoid control model requires a natural language description for each human action video. 
% To generate these descriptions, we employ VideoLLaMA2-7B~\cite{cheng2024videollama} for video captioning, and use a carefully crafted prompt to steer the model towards describing human actions rather than their physical appearance (details are provided in the supplementary materials).

\noindent\textbf{Motion Retargeting from Humans to Humanoid Robots.} Since humans and humanoid robots have similar skeletons, we can track the human joint positions across frames and map them to the corresponding joints in a humanoid robot, resulting in humanoid keypoints $\mathcal{P}_{robot}$ for high-level control. In particular, we chose 12 joints that exist in both humans and humanoid robots: left and right hips, knees, ankles, shoulders, elbows, and wrists. The joint positions $\mathcal{P}_{joints}$ can be obtained via forward kinematics $F_{fk}$:
\begin{equation} \label{eq:retarget}
    \mathcal{P}_{joints} = F_{fk}(\mathcal{P}_{human}(\beta, \theta, t_{root})).
\end{equation} 
Since humans have different shapes from humanoid robots, following~\cite{h2o}, we first optimize the human shape parameters $\beta$ to ensure that resized human shapes closely resemble those of a humanoid robot. Specifically, we first obtain joint positions in the humanoid robot under a standard T-shaped pose: $\mathcal{P}^{T}_{robot}$. Then, under the same T-shaped pose, we optimize $\beta$ to make human joint positions $\mathcal{P}^T_{joints}$ the same as the corresponding humanoid joint positions $\mathcal{P}^{T}_{robot}$:
\begin{align}
    & \min_{\beta} \, ||\mathcal{P}^T_{joints} - \mathcal{P}^{T}_{robot}||_2, \\
    & \text{s.t.} \quad \mathcal{P}^T_{joints} = F_{fk}(\mathcal{P}_{human}(\beta, \theta^T, t_{root})),
\end{align}
where $\theta^T$ denotes the standard T pose. For each frame of human pose, we replace the original $\beta$ with the optimal $\beta^{\prime}$ in $\mathcal{P}_{human}$, and following Eq.~\ref{eq:retarget} we can obtain the adjusted joint positions $\mathcal{P}^{\prime}_{joints}$. Finally, we directly set humanoid robotic keypoints as the adjusted human joint positions:
\begin{equation} \label{eq:keypoint}
    \mathcal{P}_{robot} := \mathcal{P}^{\prime}_{joints}.
\end{equation}
To effectively control humanoid robots, we also extract the motor DoF positions $q_{robot}$ in the humanoid robot via inverse kinematics $F_{ik}$:
\begin{equation} \label{eq:ik}
    q_{robot} = F_{ik}(\mathcal{P}_{robot}).
\end{equation}
We use the Adam optimizer~\cite{adam} to solve the inverse kinematics problem. A smoothing term is added to the optimization to regularize changes in $q_{robot}$ across frames.

\noindent\textbf{Goal-conditioned Humanoid Control Policy.} The retargeted humanoid keypoints $\mathcal{P}_{robot}$ and DoF positions $q_{robot}$ accurately reflect humanoid motions, but they cannot be directly deployed to the real robot. This is because they lack the necessary safety guarantees and robustness needed to handle real-world variability and constraints effectively. To address this, we develop a goal-conditioned control policy $\pi$ that adapts these motions while ensuring safe and reliable deployment on the physical robot:
\begin{equation} \label{eq:rlpolicy}
    \pi:\mathcal{G}\times\mathcal{O}\mapsto\mathcal{A}_{robot}.
\end{equation}
The inputs to the policy $\pi$ include two parts: the goal space $\mathcal{G}$ and the observation space $\mathcal{O}$. The goal space $\mathcal{G}$ contains humanoid keypoints $\mathcal{P}_{robot}$, DoF positions $q_{robot}$, and root movement goals derived from $t_{root}$. The observation space $\mathcal{O}$ contains robot proprioception information such as root orientation, angular velocity, and current motor DoF positions. The output action space $\mathcal{A}_{robot}$ are target DoF positions of each joint for controling the humanoid robot, which can be further transformed into motor torque signals through a proportional-derivative (PD) controller.

We train the control policy, $\pi$, using large-scale reinforcement learning with PPO~\cite{ppo} for policy optimization. The reward function includes multiple terms: motion rewards to encourage imitation of the retargeted humanoid keypoints $\mathcal{P}_{robot}$ and DoF positions $q_{robot}$; root tracking rewards to follow target root orientations and linear velocities from $t_{root}$; and stability rewards to help the robot maintain balance and prevent falls during movement. The resulting policy $\pi$ and robotic actions $\mathcal{A}_{robot}$ enable the humanoid robot to operate safely in the physical world while maintaining the desired motions.

Finally, we collect a large number of motion samples $\langle \mathcal{V}, \mathcal{T}, \mathcal{P}_{human}, \mathcal{P}_{robot}, \mathcal{A}_{robot} \rangle$ from massive videos. In the next section, we investigate how to train a universal humanoid pose control policy using massive motion samples.

\begin{figure}
  \centering
   \includegraphics[width=1.0\linewidth]{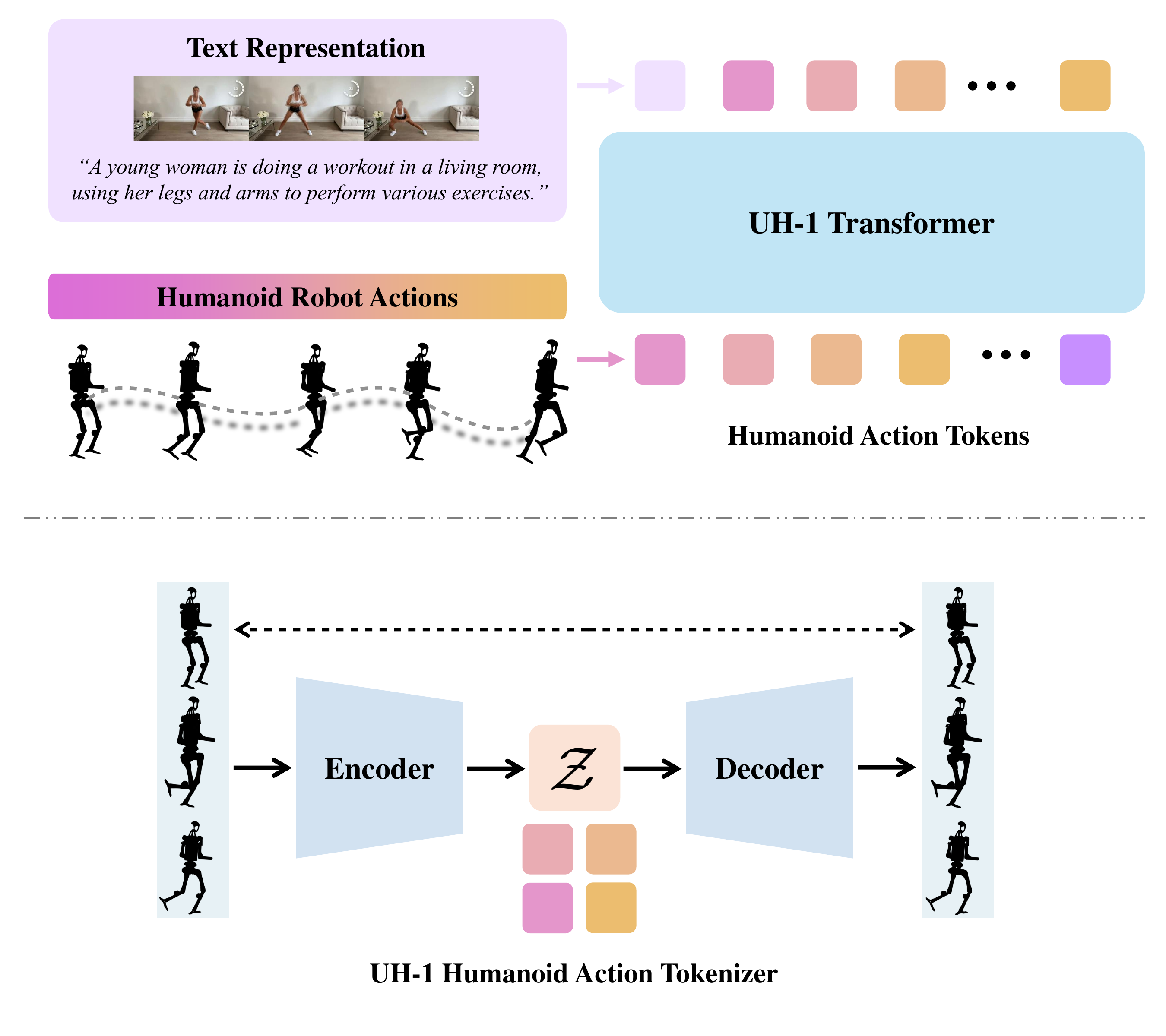}
   \caption{\textbf{UH-1 Model Architecture}. UH-1 leverages the Transformer for scalable learning. Humanoid actions are first tokenized into discrete action tokens. Then, we train the UH-1 Transformer that takes text commands as inputs and auto-regressively generates the corresponding humanoid action tokens.}
   \label{fig:uh-1-model}
\end{figure}

\section{UH-1 for Universal Humanoid Pose Control}
\begin{figure}
  \centering
   \includegraphics[width=1.0\linewidth]{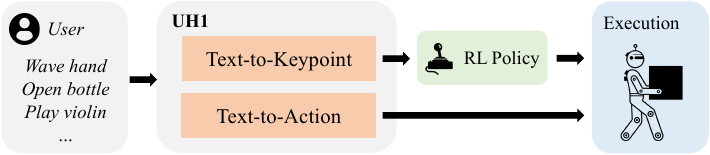}
   \caption{\textbf{Text-to-keypoint and text-to-action control modes}. UH-1 can either generate high-level humanoid keypoints (text-to-keypoint) for the goal-conditioned policy $\pi$ to control the humanoid robot in closed-loop, or generate robotic actions $\mathcal{A}_{robot}$ for direct open-loop control (text-to-action).}
   \label{fig:conrtol_mode}
  \vspace{-5mm}
\end{figure}

Learning from massive videos enables us to distill a universal humanoid pose control policy from large-scale motion samples $\langle \mathcal{V}, \mathcal{T}, \mathcal{P}_{human}, \mathcal{P}_{robot}, \mathcal{A}_{robot} \rangle$. We introduce UH-1, a large language-conditioned humanoid model that takes natural language commands $\mathcal{T}$ and generates corresponding humanoid robotic actions $\{\mathcal{P}_{robot}, \mathcal{A}_{robot}\}$:
\begin{equation}
    \pi_{\scriptscriptstyle{UH\text{-}1}}:\mathcal{T} \mapsto \{\mathcal{P}_{robot}, \mathcal{A}_{robot}\},
\end{equation}
where $\pi_{\scriptscriptstyle{UH\text{-}1}}$ denotes the UH-1 model. Notably, as illustrated in Fig.~\ref{fig:conrtol_mode}, our model can either generate high-level humanoid keypoints $\mathcal{P}_{robot}$, which are then fed into the goal-conditioned policy $\pi$ to control the humanoid robot in closed-loop, or generate robotic actions $\mathcal{A}_{robot}$ for direct open-loop control. Our model bridges the gap between semantic language commands and physically deployable robotic actions, enabling more generalizable humanoid robotic control using text instructions. For simplicity, in the following section, we use $\mathcal{A}_{robot}$ as an example to illustrate our method; $\mathcal{P}_{robot}$ can be generated in the same manner.

We adopt the Transformer~\cite{Transformer} as our main model architecture due to its scalability to large-scale data. As shown in~\cref{fig:uh-1-model}, to enable efficient learning, we first train an action tokenizer using~\cite{van2017neural} to discretize humanoid motions into a vocabulary of action tokens. Then, we train the Transformer to auto-regressively decode action tokens, resulting in executable humanoid actions.

% Drawing on insights from digital human motion generation research~\cite{tevet2023human, zhang2023generating}, which leverages text-conditioned generative models to generate human parametric models, we introduce \ourmethod, a unified model architecture for text-conditioned humanoid robot pose generation and humanoid action generation. 
% The generated pose can also function as the imitation target for a goal-conditioned RL policy. 
% Mainly following T2M-GPT~\cite{zhang2023generating}, \ourmethod integrates a Vector Quantized Variational AutoEncoder~\cite{van2017neural} (VQ-VAE) to convert input representations into discrete code sequences and a Transformer~\cite{vaswani2017attention} to generate these codes conditioned on user language instructions, as depicted in \cref{fig:model_architecture}. 
% With the trained decoder of VQ-VAE, code sequences generated by the Transformer can be converted back to the original input space. The only difference between different generation modalities lies in the input/output space. For humanoid robot pose generation, this space encompasses both robot joint values and root poses, whereas for humanoid action generation, it includes only the target joint values.

% \begin{figure*}
%   \centering
%    \includegraphics[width=1.0\linewidth]{fig/uh1.pdf}
%    \caption{UH-1 model architecture}
%    \label{fig:model_architecture}
% \end{figure*}

\noindent\textbf{UH-1 Action Tokenizer.} We follow~\cite{van2017neural} and map $T$ frames of actions $\mathcal{A}_{robot} = [a_1, \dots, a_T]$ into a sequence of discrete action tokens $\mathcal{Z}_{token} = [z_1, \dots, z_{T/K}]$ via an encoder $F_{encode}$ and quantization $F_{quant}$:
\begin{equation} \label{eq:tokenize}
    \mathcal{Z}_{token} = F_{quant}(F_{encode}(\mathcal{A}_{robot})),
\end{equation}
where $F_{encode}$ and $F_{quant}$ are standard operations in \cite{van2017neural}. 
The action tokens $\mathcal{Z}_{token}$ come from a shared action vocabulary, and each token can be viewed as a motion primitive that is learned and shared across all data samples.
Notably, different from language tokenization, humanoid actions won't change much in adjacent frames. To maintain the temporal smoothness in humanoid actions, we encode a short clip with $K$ frames of actions $[a_{iK}, \dots a_{(i+1)K}]$ into a single action token $z_i$, rather than encoding each frame individually. This approach not only preserves smooth transitions but also eases the learning process.

The decoder of VQ-VAE $F_{decode}$ tries to reconstruct the original action sequence with the latent embeddings associated with the action tokens:
\begin{equation} \label{eq:decode}
    \mathcal{A}^{\prime}_{robot} = F_{decode}(\mathcal{Z}_{token}).
\end{equation}
We denote the reconstructed action sequence as $\mathcal{A}^{\prime}_{robot} = [a^{\prime}_1, \dots, a^{\prime}_T]$. The reconstruction loss is formulated as
\begin{equation}
    L_{recon} = \sum^T_{i} (|a^{\prime}_{i} - a_{i}| + |(a^{\prime}_{i+1} - a^{\prime}_{i}) - (a_{i+1} - a_{i})|),
\end{equation}
where the first term is the $L_1$ reconstruction loss in \cite{van2017neural} and the second term encourages the first-order similarity of original and reconstructed action sequences. Additionally, we add regularization terms on latent embeddings as in \cite{van2017neural}.

\textbf{UH-1 Transformer.} We formulate the task of language-conditioned humanoid pose control as auto-regressively decoding action tokens $\mathcal{Z}_{token}$ conditioning on text commands $\mathcal{T}$. Formally, let $\mathcal{Z}_{token} = [z_1, \dots z_{T/K}]$ denote the target action token sequence, where $z_i$ is the current step to predict, and $z_{1:i-1}$ represent the preceding context of action tokens, and $l$ denote the text embedding by encoding the text command $\mathcal{T}$ with the CLIP~\cite{radford2021learning} encoder. The UH-1 Transformer is then trained to model the conditional probability distribution $P(z_i| z_{1:i-1}, l)$. A special [\texttt{End}] token is incorporated into the vocabulary to signal the termination of sequence generation. During training, we first tokenize each $\mathcal{A}_{robot}$ into $\mathcal{Z}_{token}$ using Eq. \ref{eq:retarget}. Then, we feed the language embedding $l$ into the UH-1 transformer, and the transformer auto-regressively decodes action tokens. The learning objective is to minimize the negative log-likelihood over the whole training dataset $\mathcal{D}$:
\begin{equation}
\mathcal{L}_{learn} = - \underset{\mathcal{Z} \in \mathcal{D}}{\sum} \log \underset{i = 1}{\overset{|\mathcal{Z}|}{\Pi}} p(z_i|z_{1:i-1}, l).
\label{eq:trans}
\end{equation}
During inference, using Eq. \ref{eq:decode}, the generated action tokens are decoded into $\mathcal{A}_{robot}$ for controlling the humanoid robot. The Transformer architecture and auto-regressive modeling ensure scalable learning of humanoid robot pose control.
\begin{figure*}
  \centering
   \includegraphics[width=1.0\linewidth]{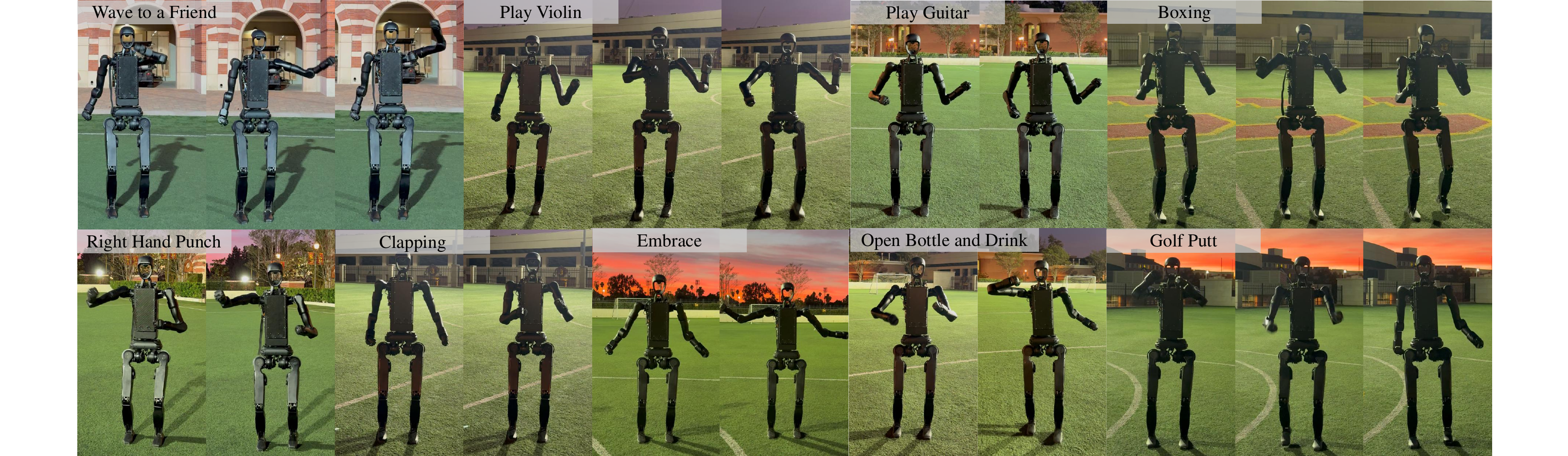}
   \vspace{-5mm}
   \caption{\textbf{Real robot experiment}. UH-1 model can be reliably deployed on the real humanoid robot with a nearly 100\% success rate.}
   \label{fig:real_robot_exp}
   \vspace{-3mm}
\end{figure*}

% \begin{table*}
%   \centering
%   \begin{tabular}{@{}l|ccc ccc@{}}
%     \toprule
%     \textbf{Methods}  & \textbf{FID} $\downarrow$ & \textbf{MM Dist} $\downarrow$ & \textbf{Diversity} $\uparrow$ & \multicolumn{3}{c}{\textbf{R Precision} $\uparrow$} \\
%     \cmidrule(lr){5-7} & & & & Top 1 & Top 2 & Top 3 \\
%     \midrule
%     Ground Truth  & $0.005^{\pm.001}$ &  $3.140^{\pm.010}$ &  $9.846^{\pm.062}$ &  $0.488^{\pm.003}$ & $0.681^{\pm.004}$ & $0.780^{\pm.003}$ \\
%     \midrule
% Motion Diffusion &  $0.582^{\pm.051}$ &  $5.921^{\pm.034}$ &  ${10.122^{\pm.078}}$ &  $0.409^{\pm.008}$ & $0.521^{\pm.006}$ & $0.617^{\pm.007}$ \\
% Text-to-Motion GPT &  $0.667^{\pm.109}$  &  $3.401^{\pm.017}$  &  $\mathbf{10.328^{\pm.099}}$ & $0.461^{\pm.005}$ &  $0.648^{\pm.005}$  &  $0.734^{\pm.004}$  \\
%      \midrule
%      UH-1 (ours)  &  $\mathbf{0.445^{\pm.078}}$  &  $\mathbf{3.249^{\pm.016}}$  &  ${10.157}^{\pm.106}$ & $\mathbf{0.479^{\pm.005}}$ &  $\mathbf{0.663^{\pm.004}}$  &  $\mathbf{0.761^{\pm.003}}$ \\
%     \bottomrule
%   \end{tabular}
%   \caption{\textbf{Comparison of model performance across key metrics on a commonly used benchmark}~\cite{guo2022generating}. We calculate standard metrics following~\cite{guo2022generating}, repeating each evaluation 20 times and reporting the average along with the 95\% confidence interval. The \textbf{bolded} values highlight the best results. The results indicate that UH-1 attains the highest performance across nearly all metrics and achieves comparable performance on the \textit{Diversity} metric.}
%   \label{tab:model_exp1}
% \end{table*}

\begin{table}
  \centering
  \resizebox{1.0\linewidth}{!}{
  \begin{tabular}{@{}l|cccc@{}}
    \toprule
    \textbf{Methods}  & \textbf{FID} $\downarrow$ & \textbf{MM Dist} $\downarrow$ & \textbf{Diversity} $\uparrow$ & {\textbf{R Precision} $\uparrow$} \\
    \midrule
    Oracle  & $0.005^{\pm.001}$ &  $3.140^{\pm.010}$ &  $9.846^{\pm.062}$ & $0.780^{\pm.003}$ \\
    \midrule
MDM~\cite{tevet2023human} &  $0.582^{\pm.051}$ &  $5.921^{\pm.034}$ &  ${10.122^{\pm.078}}$  & $0.617^{\pm.007}$ \\
T2M-GPT~\cite{zhang2023generating} &  $0.667^{\pm.109}$  &  $3.401^{\pm.017}$  &  $\mathbf{10.328^{\pm.099}}$  &  $0.734^{\pm.004}$  \\
     \midrule
     UH-1 (ours)  &  $\mathbf{0.445^{\pm.078}}$  &  $\mathbf{3.249^{\pm.016}}$  &  ${10.157}^{\pm.106}$  &  $\mathbf{0.761^{\pm.003}}$ \\
    \bottomrule
  \end{tabular}
  }
  \caption{\textbf{Comparisons of model performances on the HumanoidML3D benchmark}. We calculate standard metrics following~\cite{guo2022generating}, repeating each evaluation 20 times and reporting the average along with the 95\% confidence interval. The results indicate that UH-1 attains the highest performance across most metrics and achieves comparable performance on the \textit{Diversity} metric.}
  \label{tab:model_exp1}
  \vspace{-5mm}
\end{table}

\section{Experiments}

In this section, we conduct extensive experiments to investigate the following research questions:
(1) \textit{Universal Pose Control with UH-1}: Does our UH-1 model enable universal humanoid robot pose control based on text commands?
(2) \textit{Scalability and Generalization with Humanoid-X}: Does the large-scale Humanoid-X dataset facilitate scalable training and improve the generalization ability of our UH-1 model?
(3) \textit{Real-World Deployment of UH-1}: Can our UH-1 model be deployed on real humanoid robots to enable reliable robotic control in real-world environments?

\subsection{Universal Humanoid Pose Control with UH-1}

We conduct extensive experiments to validate the generalization ability of the UH-1 model. An alternative solution to text-to-humanoid action generation is a two-stage pipeline: generating 3D human motions first and then retargeting the human motions to humanoid robots. To this end, we compare our method with two important baselines for text-to-human motion generation: Motion Diffusion Model (MDM)~\cite{tevet2023human} and Text-to-Motion GPT (T2M-GPT)~\cite{zhang2023generating}. For fair comparisons, We choose the commonly used HumanML3D~\cite{guo2022generating} benchmarks and transform the humans in this dataset into humanoid robots, resulting in a new benchmark called HumanoidML3D. Similarly, we adopt the same motion retargeting method as in this paper to transform the human motions generated by the baselines into humanoid actions. We adopt the metrics in~\cite{guo2022generating} to evaluate the humanoid motions from different aspects: (1) Quality: The \textit{Frechet Inception Distance (FID)} evaluates the dissimilarity between feature distributions of generated and ground truth humanoid poses. (2) Diversity: The \textit{Diversity} metric evaluates the variability within the generated humanoid pose distribution, calculated as the average Euclidean distance between 300 randomly sampled pairs of humanoid poses. (3) Reliability: The \textit{Multi-modal Distance (MM Dist)} measures the Euclidean distance between motions and corresponding texts, and the \textit{R Precision} assesses the accuracy of text and humanoid pose matches in the Top 3 rankings.

\cref{tab:model_exp1} shows the results of our UH-1 model compared against the baselines. The results indicate that UH-1 attains the highest performance across nearly all metrics, showing an over 23\% reduction in the critical \textit{FID} metric, while also maintaining comparable performance on the \textit{Diversity} metric. 
The first-order similarity loss proposed in this paper greatly enhances the quality and reliability of the generated outputs.
The results suggest that UH-1 is a streamlined model and performs better than the two-stage methods.

\subsection{Scalable Learning with Humanoid-X}

\begin{table}
  \centering
  \resizebox{1.0\linewidth}{!}{
  \begin{tabular}{@{}l|cccc@{}}
    \toprule
    \textbf{Dataset} & \textbf{FID} $\downarrow$ & \textbf{MM Dist} $\downarrow$ & \textbf{Diversity} $\uparrow$ & {\textbf{R Precision} $\uparrow$} \\
    \midrule
    Oracle & $0.005^{\pm.001}$ &  $3.140^{\pm.010}$ &  $9.846^{\pm.062}$  & $0.780^{\pm.003}$\\
    \midrule
     HumanoidML3D &  ${0.445}{^{\pm.078}}$  &  ${3.249}{^{\pm.016}}$  &  ${10.157^{\pm.106}}$ & $0.760^{\pm.003}$ \\

    Humanoid-X  &  $\mathbf{0.379^{\pm.046}}$  &  $\mathbf{{3.232}^{\pm.008}}$  &  $\mathbf{10.221^{\pm.100}}$ & $\mathbf{0.761^{\pm.003}}$ \\
    \bottomrule
  \end{tabular}
  }
  \caption{\textbf{Dataset quality evaluation.}. Training on the Humanoid-X dataset greatly improves the quality and reliability of humanoid actions, compared to training on the HumanoidML3D dataset.}
  \label{tab:model_exp2}
  % \vspace{-5mm}
\end{table}

In this section, we investigate whether scaling up training data with the large-scale Humanoid-X dataset can improve the generalization ability of our model. To explore this, we first pre-trained our UH-1 model on the Humanoid-X dataset and then finetuned and evaluated the performance on the HumanoidML3D benchmark. \cref{tab:model_exp2} shows the performance comparison with training only on HumanoidML3D. We found that pre-training on the Humanoid-X dataset greatly improves the quality, reliability, and diversity of humanoid actions, with an \textit{FID} improvement from $0.445$ to $0.379$, a \textit{MM Dist} score improvement from $3.249$ to $3.232$, and a \textit{Diversity} improvement from $10.157$ to $10.221$.

\begin{figure}
  \centering
   \includegraphics[width=1.0\linewidth]{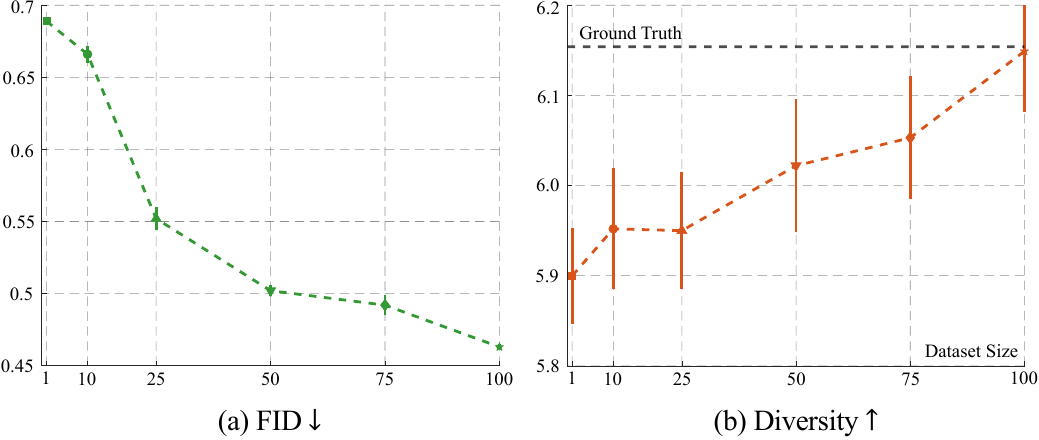}
   \caption{\textbf{Effectiveness of scaling up training data}. Points indicate the mean values, and error bars indicate the 95\% confidence interval. Increasing the dataset size from 1\% to 100\% leads to significant improvements in both \textit{FID} and \textit{Diversity} metric.}
   \label{fig:scaling_exp}
  \vspace{-5mm}
\end{figure}

In addition, we also study how scaling up training data affects the model performance. To this end, we train our UH-1 model on varying proportions of the Humanoid-X dataset, specifically 1\%, 10\%, 25\%, 50\%, 75\%, and 100\%. The results shown in \cref{fig:scaling_exp} indicate that scaling up training data from 1\% to 100\% leads to a significant performance improvement in all metrics (\textit{FID} from $0.689$ to $0.463$ and \textit{Diversity} from $5.900$ to $6.149$). This suggests that by learning from massive videos, we successfully scale up the training data of humanoid robots and attain better performance.

\subsection{Real-World Deployment of UH-1} \label{sec:5.3}

To investigate whether our UH-1 model, trained on the Humanoid-X dataset, can generate reliable humanoid actions that are physically deployable on humanoid robots, we designed 12 distinct language commands, as shown in \cref{tab:robot_exp1}, and evaluated them on a real humanoid robot. We use \textsc{Unitree H1-2}\footnote{\href{https://www.unitree.com/h1}{https://www.unitree.com/h1}} as our test embodiment. For the experiments, we evaluated each language command 10 times and controlled the robot in different places. Notably, for text-to-humanoid actions, we found that open-loop control can only work for upper-body control, so in this control mode, we use a pre-trained locomotion policy for controlling the lower-body of the humanoid robot. \cref{fig:real_robot_exp} shows the demos of real-robot experiments. \cref{tab:robot_exp1} measures the task success rate for each language command. Our experimental results demonstrate that our UH-1 model can be reliably deployed on the real humanoid robot, achieving a success rate of nearly 100\% across all evaluated language instructions.

\subsection{Empirical Studies}
\begin{figure}
  \centering
   \includegraphics[width=0.99\linewidth]{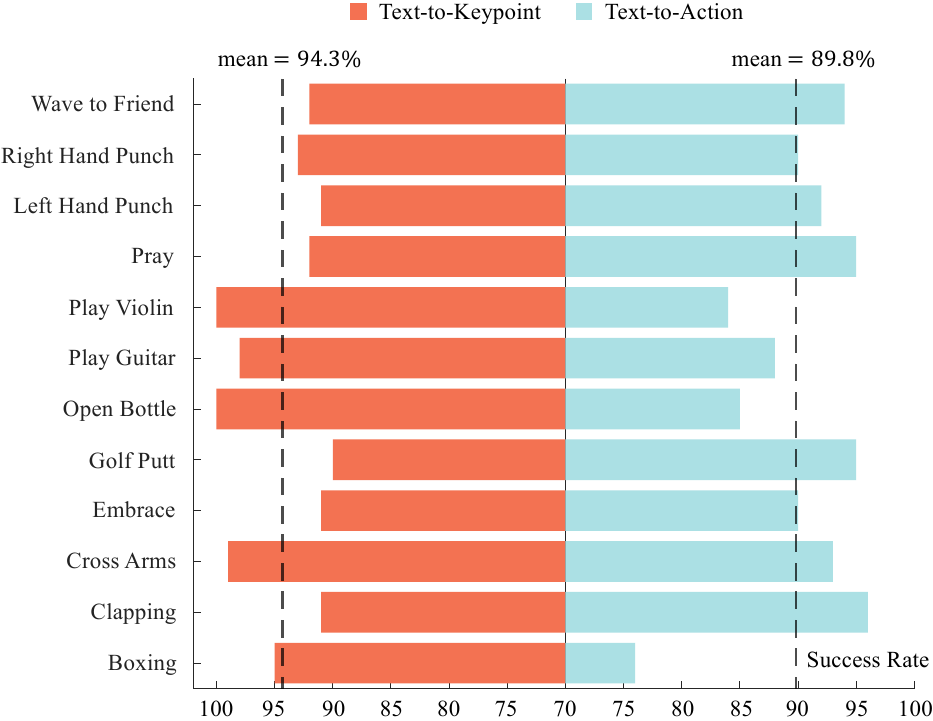}
   \caption{\textbf{Simulated experiments on the UH-1 control modes}. Bars indicate success rates for specific commands and dash lines show the mean success rate on 12 different text instructions. While \textit{Text-to-Action} mode with a separate locomotion policy is sufficient for most tasks, \textit{Text-to-Keypoint} mode shows greater robustness.}
   \label{fig:control_mode_analysis}
\end{figure}

\textbf{Analysis of two control modes.} UH-1 can either produce high-level humanoid keypoints for a goal-conditioned, closed-loop policy or directly generate robotic actions for open-loop control. To investigate the effectiveness of these two control modes, we randomly generate 100 keypoint sequences and 100 action sequences for each task, as illustrated in \cref{fig:control_mode_analysis}, and apply them in simulated robot control. The findings indicate that both modes can achieve an average success rate exceeding 89\%, suggesting that text-to-action open-loop control with a separate locomotion policy is sufficient for most tasks. Moreover, the text-to-keypoint control mode, benefiting from the whole-body control policy, demonstrates slightly better robustness.

\begin{table}
  \centering
  \resizebox{0.4\textwidth}{!}{%
    \begin{tabular}{@{}c|cc@{}}
      \toprule
      \textbf{Instruction}  & \textbf{Text-to-Keypoint}  & \textbf{Text-to-Action}  \\
      % \cmidrule(lr){2-3} \cmidrule(lr){4-5} 
      % & Sim. & Real. & Sim. & Real. \\
      \midrule
\rowcolor{gray!20} Boxing & 90\% & 70\%  \\
    Clapping & 100\% & 100\% \\
\rowcolor{gray!20} Cross Arms & 80\% & 80\% \\
       Embrace & 100\% & 100\% \\
\rowcolor{gray!20} Golf Putt & 90\% & 100\% \\
      Open Bottle \& Drink & 100\% & 100\%  \\
\rowcolor{gray!20}  Play Guitar & 100\% & 100\% \\
      Play Violin & 100\% & 80\% \\
\rowcolor{gray!20} Pray & 100\% & 100\% \\
      Left Hand Punch & 100\% & 100\% \\
\rowcolor{gray!20} Right Hand Punch & 100\% & 90\% \\
      Wave to Friend & 100\% & 100\% \\
      \bottomrule
    \end{tabular}%
  }
  \caption{\textbf{Task success rate on a real humanoid robot}. Both \textit{Text-to-Keypoint} and \textit{Text-to-Action} modes can reach a success rate of nearly 100\% across all evaluated language instructions. }
  \label{tab:robot_exp1}
  \vspace{-5mm}
\end{table}

\begin{figure}
  \centering
   \includegraphics[width=1.0\linewidth]{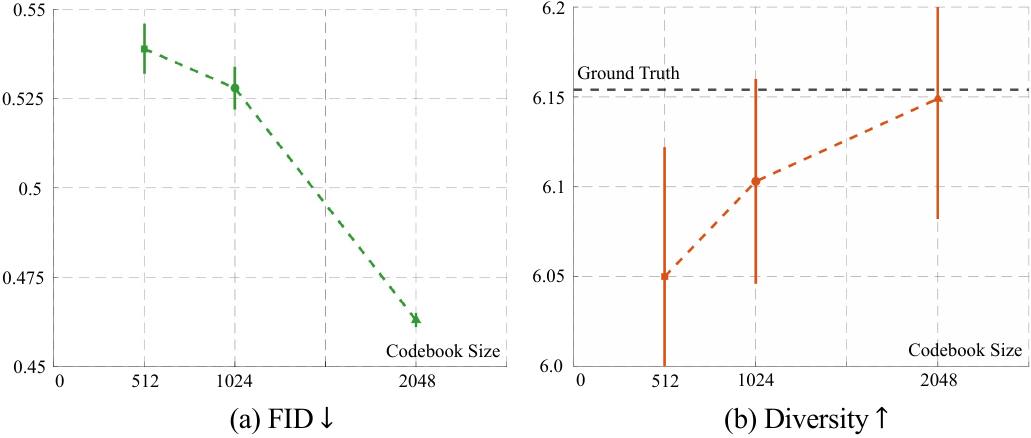}
   \caption{\textbf{Ablation on the vocabulary sizes of the UH-1 action tokenizer}. Increasing the vocabulary size of the action tokenizer provides more motion primitives for humanoid robots and thus leads to an improvement in both \textit{FID} and \textit{Diversity} metric.}
   \label{fig:codebook_ablation}
  \vspace{-5mm}
\end{figure}

\noindent\textbf{Ablation study on the action tokenizer.} We conduct an ablation study to investigate the impact of different vocabulary sizes of the UH-1 action tokenizer on model training. We selected the vocabulary sizes of 512, 1024, and 2048, and reported the model performances on the Humanoid-X dataset. As illustrated in \cref{fig:codebook_ablation}, increasing the vocabulary size up to 2048 leads to an improvement in \textit{FID} metric from $0.539$ to $0.463$ and brings an improvement in \textit{Diversity} metric from $6.050$ to $6.149$. This indicates that increasing the number of motion primitives learned in the action tokenizer results in more diverse humanoid motion generation. Due to the limited computational resources, we didn't try a larger vocabulary. We will leave this for future works.

\noindent\textbf{Ablation study on the model architecture.} A key consideration for generation tasks is selecting the appropriate model architecture, such as the Transformer or diffusion model. To explore this, we trained a text-controlled humanoid motion diffusion model on the Humanoid-X dataset and compared its performance with the original Transformer-based UH-1 model. The results in \cref{tab:arch} show that the Transformer architecture used in UH-1 is more scalable to large-scale training data and achieves better performance, with a lower \textit{FID} and \textit{MM Dist} score compared to the diffusion-based model.

\begin{table}
  \centering
  \resizebox{0.98\linewidth}{!}{
  \begin{tabular}{@{}l|cccc@{}}
    \toprule
    \textbf{Methods} & \textbf{FID} $\downarrow$ & \textbf{MM Dist} $\downarrow$ & \textbf{Diversity} $\uparrow$ & {\textbf{R Precision} $\uparrow$} \\
    \midrule
    Oracle & $0.005^{\pm.001}$ &  $3.140^{\pm.010}$ &  $9.846^{\pm.062}$  & $0.780^{\pm.003}$\\
    \midrule
Diffusion model & $0.624^{\pm.074}$ & $5.536^{\pm.029}$  &  $\mathbf{10.281^{\pm.096}}$ & $0.630^{\pm.007}$ \\

    Transformer  &  $\mathbf{0.379^{\pm.046}}$  &  $\mathbf{{3.232}^{\pm.008}}$  &  ${10.221^{\pm.100}}$ & $\mathbf{0.761^{\pm.003}}$ \\
    \bottomrule
  \end{tabular}
  }
  \vspace{-3mm}
  \caption{\textbf{Diffusion model vs. Transformer as the UH-1 model.} We found that the Transformer architecture is more scalable to large-scale training data and exhibits better performance.}
  \label{tab:arch}
  % \vspace{-5mm}
\end{table}

% \begin{figure}
%   \centering
%    \includegraphics[width=1.0\linewidth]{fig/rl_ablation.pdf}
%    \caption{\textbf{Ablation on different RL policies}, measured by task cumulative reward value. The solid line represents the mean return value, while the shaded regions correspond to the standard deviation, both calculated across five different random seeds. Our retargeted training data enhances the performance of the RL policy in tracking the imitation of body keypoints, joint positions, root orientation and root linear velocity.}
%    \label{fig:rl_policy_ablation}
% \end{figure}
\vspace{-2mm}
\section{Conclusion}

We introduce Humanoid-X, a large-scale dataset that facilitates scalable humanoid robot learning from massive videos. On top of Humanoid-X, we trained a large humanoid model, UH-1, for generalizable humanoid pose control based on language commands. Extensive experiments demonstrate that scalable training enables UH-1 to generate generalizable and reliable humanoid actions following language commands, and the UH-1 model can be effectively deployed on the real humanoid robot.

\textbf{Limitations.} In this paper, we only study the humanoid pose control. Humanoid manipulation is not in the scope of this paper. In future works, we plan to investigate learning humanoid loco-manipulation from Internet videos. 
\section*{Acknowledgement}
The USC Geometry, Vision, and Learning Lab acknowledges generous supports from Toyota Research Institute, Dolby, and Google DeepMind. Yue Wang is also supported by a Powell Research Award. 

% \clearpage
{
    \small
    \bibliographystyle{ieeenat_fullname}
    \bibliography{main}
}

\clearpage
\setcounter{page}{1}
\appendix

% Title
{
\noindent\Large\textbf{Appendix}
}
% Table of Contents Settings
% \let\oldcolor\color
% {
% \hypersetup{linkcolor=black}

% \titlecontents{subsection}[2em]
% {\vspace{0.1em}}
% {\thecontentslabel\quad} 
% {}
% {\titlerule*[0.8em]{.}\contentspage}
% [\vspace{0.1em}]

% \startcontents[appendix]
% \printcontents[appendix]{}{1}{}
% }

\setcounter{figure}{0}
\setcounter{table}{0}
\setcounter{equation}{0}

\startcontents % Start collecting content for the partial TOC here
{
    \hypersetup{linkcolor=black}
    \printcontents{}{1}{}
}
\newpage

% \vfill % Fill the remaining space in the column
% \columnbreak % Move to the next column
% \vspace*{\fill}

\section{Ethics Statement}
\label{sec:ethics}
This paper presents Humanoid-X, a large-scale dataset that facilitates scalable humanoid robot learning from massive videos, and UH-1, a large humanoid model for generalizable humanoid pose control based on language commands.
The Internet videos that Humanoid-X and UH-1 involve in the dataset and the pipeline are strictly for academic research and are not intended for commercial use.
On the privacy protection side, we apply face anonymization to all human subjects in the Internet videos involved in Humanoid-X and UH-1, making sure that the videos do not include any personal information.
In addition, we will not release the original Internet videos to protect copyright. 
In summary, we believe that Humanoid-X and UH-1 do not raise ethical concerns.

\section{Details on Humanoid-X Data Collection}
In this section, we will introduce more details on the whole data collection pipeline of the Humanoid-X dataset, including data source distribution, video mining and processing, video captioning, 3D human pose estimation, motion retargeting from humans to humanoid robots, and the goal-conditioned humanoid control policy.

\subsection{Data Source Distribution}
Humanoid-X consists of massive motion samples with diverse sources, and the detailed source of the data in our Humanoid-X dataset is shown in \cref{tab:dataset_statistics}.
Humanoid-X consists of 163.8K motion samples, spanning 240.3 hours of video footage, containing 20.7M frames of human and robotic motion data, with a vocabulary size of 3206 words.
Each motion video sample is expanded to the 5 data modalities $\langle \mathcal{V}, \mathcal{T}, \mathcal{P}_{human}, \mathcal{P}_{robot}, \mathcal{A}_{robot} \rangle$ of the motion sample in our Humanoid-X dataset.
The subsections below introduce details on the dataset building and data processing pipeline.

\begin{figure*}
  \centering
   \includegraphics[width=1.0\linewidth]{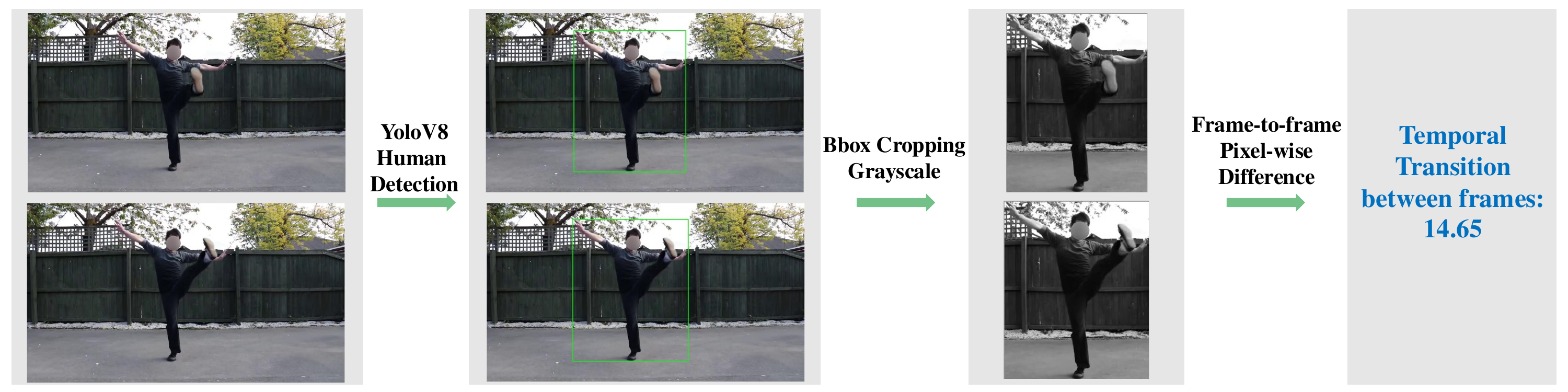}
   \caption{\textbf{Video Processing Pipeline}.}
   \label{fig:yolo_pipeline}
\end{figure*}

\subsection{Video Mining and Processing}
% \todo{Introduce the Internet videos search terms.}
To collect a dataset of videos featuring single-person movements, we first designed specific motion categories and then generated search prompts based on these categories. Using the phrase ``single person'' in searches often produced irrelevant results since the majority of the video titles would not specify whether the video is single person using the exact word ``single person''. So, activity-based terms were created to ensure relevant data retrieval. These categories included martial arts tutorials, fitness and exercise drills, sports techniques, dance practice, music performance tutorials, everyday movement patterns, animal-inspired movements, and rehabilitation exercises.

Martial arts tutorials included search terms for techniques, drills, and demonstrations across disciplines like Wushu, Taekwondo, Karate, and Kung Fu. Examples of generated terms are ``karate front kick training,'' ``taekwondo spinning hook kick demonstration,'' and ``wushu staff spin practice.'' Fitness and exercise drills focused on isolated movements like ``yoga handstand practice,'' and  ``calisthenics planche progression tutorial,''.

Sports techniques targeted individual actions in activities like baseball, tennis, archery, running, and parkour, with examples including ``tennis serve technique tutorial'' and ``running stride form analysis.'' Dance practice emphasized solo routines in styles such as salsa, hip hop, ballet, modern dance, and improvisation, using terms like ``salsa basic turn solo'' and ``ballet arabesque demonstration.'' Music performance tutorials captured movements involved in playing instruments such as guitar, violin, piano, and drums, with terms like ``guitar strumming while standing solo'' and ``violin bowing technique while standing demonstration.''

Everyday movement patterns focus on practical motions during daily activities, using terms like ``picking up an object while balancing,'' ``loading a dishwasher with proper form,'' and ``squatting to tie shoelaces,''. Animal-inspired movements were included to capture dynamic motion patterns with terms like ``bear crawl coordination movement,'' ``frog jump exercise,'' and ``flamingo balance on one leg.'' Rehabilitation and mobility exercises targeted balance, flexibility, and strength, focusing on slow and deliberate movements such as ``dynamic torso twist warm-up'' and ``hip flexor stretch technique breakdown.''

By designing categories and generating search terms from these, we ensured the collected videos focused on single-person movements while covering a wide range of activities. 
% \todo{Single human detector details.}

% \todo{Motion detection by calculating the pixel-wise grayscale difference.}
After collection of videos from the designed searching prompts, we designed a pipeline for detecting and extracting video segments featuring single-person movements. The process begins with the YOLOv8 model~\cite{yolov8}, which detects objects in each frame and identifies detected humans based on the class label corresponding to ``person''. Frames containing exactly one detected person are selected, ensuring the focus remains solely on single-person actions. Once a single-person frame is identified, a region of interest (ROI) is extracted using the bounding box of the detected individual from YOLOv8 detection result. To determine existence of motion, the pipeline calculates frame-to-frame differences in the grayscale ROI, assessing movement levels using predefined thresholds. This ensures only frames with significant motion are retained, while static or irrelevant segments are excluded.

To further refine the selection regarding motion, the pipeline employs a batch-based filtering process, analyzing sequences of frames to identify consistent motion patterns over time. Small movement threshold is applied to frame-to-frame and a larger threshold is applied to the frame batch, enabling the detection of subtle and significant activities by allowing relative small motions for several frames as long as large motion is detected in frames' batch. Such design would benefit continuity of the clips by keeping frames in between large motions. Frames that meet these criteria are grouped into chunks representing continuous motion, and only chunks exceeding a minimum duration are considered for clip generation.

\begin{table}
  \centering
    \resizebox{0.99\linewidth}{!}{%
    \begin{tabular}{@{}l|cccc@{}}
    \toprule
    Data Source & \# of Clips & \# of Frames & \# of Hours & Vocab. Size \\
    \midrule
    AIST & 1.5K & 0.3M & 3.2 & 590  \\
  \rowcolor{gray!20}  AMASS & 13.4K & 2.0M & 27.4 & 3942\\
    Charades & 9.3K & 1.0M & 1.0 & 813 \\
\rowcolor{gray!20}    EgoBody & 1.0K & 0.4M & 4.0 & 367\\
  GRAB & 1.3K & 0.4M & 3.8 & 565 \\
\rowcolor{gray!20}    HAA500 & 5.2K & 0.3M & 2.9 & 1754 \\
    HuMMan & 0.7K & 0.1M & 1.0 & 980 \\
    \rowcolor{gray!20} IDEA400 & 12.5K & 2.6M & 24.0 & 1715\\
    Kinetics700 & 68.6K & 5.2M & 72.4 & 3360 \\
    \rowcolor{gray!20} MotionX Video & 40.6K & 7.9M & 72.9 & 4021 \\
   Online Video & 17.8K & 2.3M & 32.6 & 2040 \\
     \midrule
    Total & \textbf{163.8K} & \textbf{20.7M} & \textbf{240.3} & \textbf{11897} \\
    \bottomrule
  \end{tabular}
  }
  \caption{\textbf{Dataset statistics}. Compiled from diverse data sources, Humanoid-X possesses an extensive scale of data modalities and a massive action vocabulary.}
  \label{tab:dataset_statistics}
\end{table}

\begin{figure*}
  \centering
   \includegraphics[width=1.0\linewidth]{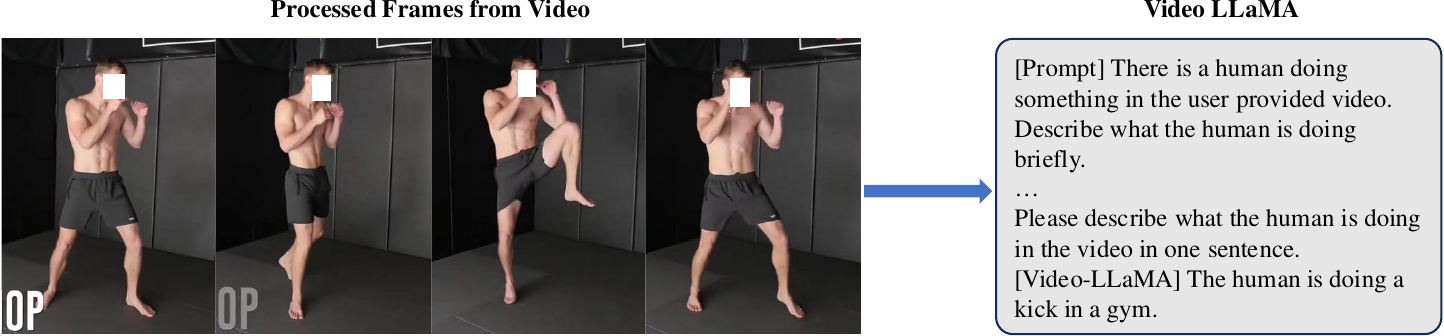}
   \caption{\textbf{Video captioning example by using Video-LLaMA}.}
   \label{fig:llama}
\end{figure*}

The output clips are processed to maintain consistent quality and playback speed. Frames within each chunk are down-sampled for efficiency, interpolated for smooth transitions, and standardized to 20 FPS. The resulting clips focus exclusively on single-person actions, discarding distractions such as multiple individuals or irrelevant frames. This approach ensures a precise and diverse dataset of single-person motion segments, suitable for applications in motion analysis, action recognition, and training of computer vision models. By integrating object detection, motion analysis, and sequence processing, the pipeline achieves high accuracy and relevance in isolating meaningful single-person movements.

\subsection{Video Captioning}

\begin{tcolorbox}[colback=blue!5!white,colframe=blue!75!black,title=Video-LLaMA Prompt]
There is a human doing something in the user provided video. Describe what the human is doing briefly.

You must follow the following rules:\par

1. Do not describe the appearance of the human. \par

2. You must at least answer ``a man/woman doing something [adverb]''\par

3. If applicable, you should describe the [item] the human is interacting with, the [body part] the human is using, or the [location] the human is in.

4. Your answer must be within one sentence, and do not begin with ``in the video''.

Please describe what the human is doing in the video in one sentence.
\end{tcolorbox}

% \todo{Video caption model parameter details.}
For video captioning, we implemented a video captioning pipeline using Video LLaMA~\cite{cheng2024videollama}, with a video processing framework which extracts visual information from input videos by sampling a fixed number of eight frames at regular intervals.

\begin{figure*}
  \centering
   \includegraphics[width=1.0\linewidth]{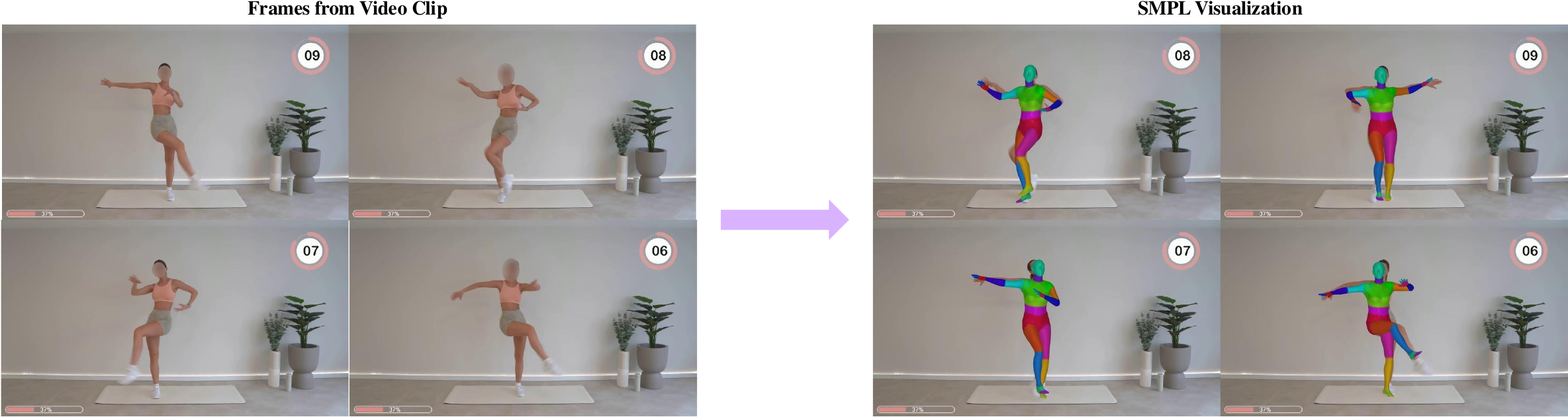}
   \caption{\textbf{SMPL 3D human model estimation example}.}
   \label{fig:smpl}
\end{figure*}

The prompts used for video captioning are designed to produce concise and action-focused descriptions. The main prompt directs the model to describe the actions of a person in the video in a single sentence, explicitly avoiding mentions of the person’s appearance. We used the query ``Please describe what the human is doing in the video in one sentence.'' with guidance of rules shown above. Such a query would guarantee a concise description of motion without any irrelevant information being collected. An example of such interaction with Video-LLaMA is shown in \cref{fig:llama}.

% \todo{Prompt design for video caption models.}

\subsection{3D Human Pose Estimation}

% \todo{SMPL estimator details}

% \todo{Root translation details}
The SMPL generation pipeline is designed to estimate 3D human pose and shape parameters from video frames. This process involves several key steps, including detecting the subject in video frames, estimating pose and shape parameters, and generating a 3D mesh representation. VIBE model~\cite{kocabas2020vibe} is used to infer SMPL parameters, such as body pose, global orientation, and shape coefficients, from video sequences. Bounding boxes are first detected for the subject, and these are used to crop and process the frames for subsequent steps. The final output includes SMPL parameters, root translations, and optional visualizations of the 3D mesh overlaid on the video frames.

The VIBE-based mesh regression model is used as video-based inference, which benefits from temporal consistency across frames. For the detected person in a video, the pipeline extracts bounding boxes and sequences of features from the video frames. VIBE processes these sequences to estimate the SMPL parameters, including pose rotations, shape coefficients, and camera parameters. The extracted parameters are then stored for further use in 3D visualization or downstream tasks. An example of SMPL visualization is shown in \cref{fig:smpl}.

To compute the root translation of the subject in 3D space, the bounding boxes and camera parameters from the mesh regression step are combined. The bounding box coordinates are converted to the original image coordinate system, accounting for resolution and aspect ratio. Using the weak-perspective camera parameters, including scale \(s\) and 2D translation \(\mathbf{t} = (t_x, t_y)\), the depth \(t_z\) is estimated based on a predefined focal length \(f\). The depth is computed as:

\begin{equation}
t_z = \frac{f}{s \cdot 0.5 \cdot W_{\text{img}}},
\end{equation}
where \(W_{\text{img}}\) represents the width of the input image. The root translation vector \(\mathbf{T}_{\text{root}}\) is then formed as:

\begin{equation}
\mathbf{T}_{\text{root}} = 
\begin{bmatrix} 
t_x \\ 
t_y \\ 
t_z 
\end{bmatrix},
\end{equation}
where \(\mathbf{t} = (t_x, t_y)\) corresponds to the 2D translations from the camera parameters, and \(t_z\) is the computed depth.

% \subsection{Motion Retargeting from Humans to Humanoid Robots}
\subsection{Motion Retargeting}
Our motion retargeting process mainly consists of two tasks: the optimization of human shape parameters $\beta$ to fit human shapes to those of a humanoid robot, and solve the humanoid motor DoF positions $q_{robot}$ from adjusted human joint positions with inverse kinematics.

\noindent\textbf{Optimization of human shape parameters $\beta$}.
Given the forward kinematics of human body models in \cref{eq:retarget}, we optimize the human shape parameters $\beta$ with the Adam optimizer~\cite{adam}, using the loss $\mathcal{L}(\beta)$:
\begin{align}
    & \mathcal{L}(\beta)=||\mathcal{P}^T_{joints} - \mathcal{P}^{T}_{robot}||_2, \\
    & \text{s.t.} \quad \mathcal{P}^T_{joints} = F_{fk}(\mathcal{P}_{human}(\beta, \theta^T, t_{root})).
\end{align}
To avoid overfitting on $\mathcal{P}^{T}_{robot}$ which leads to too much deformation on the human model T-shaped pose, we set a limit to the human shape parameters $\beta$:
\begin{equation}
    \forall i \in \{1, 2, \dots, n\}, \beta=(\beta_1, \beta_2, \dots, \beta_{n}), |\beta_i|<5,
\end{equation}
where $n$ denotes the size of the human shape parameters $\beta$.

\noindent\textbf{Solving humanoid motor DoF positions $q_{robot}$}.
With the optimal $\beta$ and \cref{eq:keypoint}, we need to extract the motor DoF positions $q_{robot}$ through inverse kinematics in \cref{eq:ik}.
The inverse kinematics problem is solved by optimimzation with the loss $\mathcal{L}_{ik}$:
\begin{equation} \label{eq:ikloss}
\mathcal{L}_{ik}=\mathcal{L}_{r}+\lambda\mathcal{L}_{s}.
\end{equation}
In \cref{eq:ikloss}, the retarget loss $\mathcal{L}_{r}$:
\begin{equation}
\mathcal{L}_{r}(q_{robot}, s_{root})=||F_{rk}(q_{robot}, s_{root}) - \mathcal{P}_{robot}||_{1},
\end{equation}
where $s_{root}$ denotes robot root states including root translation and root orientation, $F_{rk}$ denotes robot forward kinematics which maps from $q_{robot}, s_{root}$ to humanoid robot keypoint positions.
Also in \cref{eq:ikloss}, the smoothing term $\mathcal{L}_{s}$:
\begin{equation}
\mathcal{L}_{s}(q_{robot})=\sum_{i=1}^{n-2}(2q_{robot}[i]-q_{robot}[i-1]-q_{robot}[i+1]),
\end{equation}
where $n$ is the number of frames of one motion sample trajectory, with the index ranging from $0$ to $n-1$.
We use the Adam optimizer~\cite{adam} to solve the inverse kinematics problem, where the weight of smoothing term $\lambda=0.05$.

\begin{figure}
  \centering
   \includegraphics[width=1.0\linewidth]{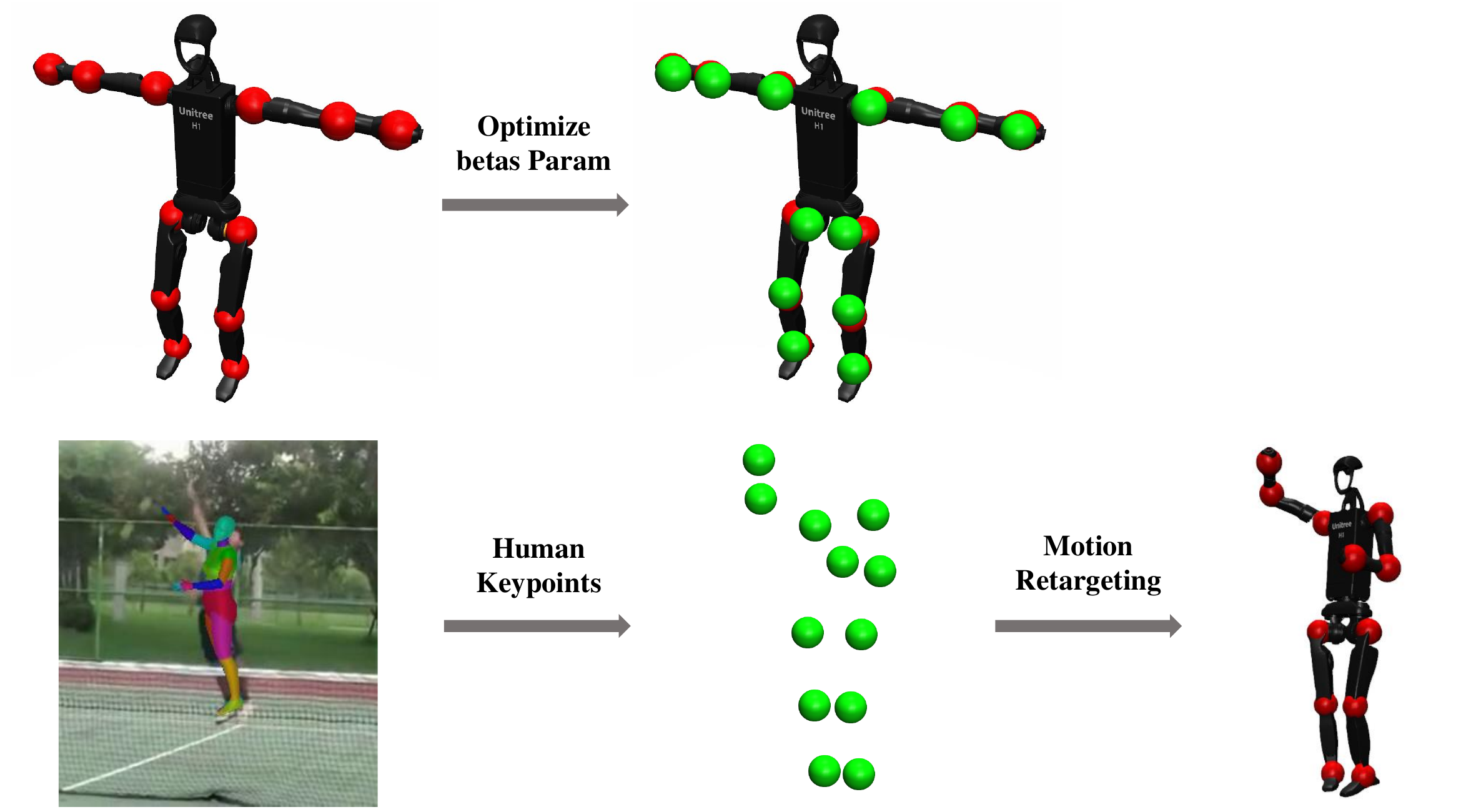}
   \caption{\textbf{Motion Retargeting}, including optimization of human shape parameters and solving humanoid motor DoF positions.}
   \label{fig:retargeting}
\end{figure}

% \subsection{Goal-conditioned Humanoid Control Policy}
\subsection{Goal-conditioned Control Policy}
We use massively parallel simulation to train our goal-conditioned humanoid RL control policy with Isaac Gym. In this subsection, we will introduce our training data, our policy, our training rewards and training parameters.

\noindent\textbf{Training Data}.
We selectively used a portion of the CMU MoCap dataset in AMASS~\cite{AMASS}, in the form of SMPL models.
We exclude motions that involve physical interactions with others, heavy objects, or rough terrain.
We retarget from the training data to humanoid robot motion with the method introduced above, including humanoid keypoint joint positions $\mathcal{P}_{robot}$, humanoid robot DoF positions $q_{robot}$ and humanoid robot root states $s_{root}$.
We can estimate the corresponding linear or angular velocities of humanoid DoFs and humanoid root joint from the humanoid motion data across frames.

\begin{table}
  \centering
  \begin{tabular}{@{}l|c|c@{}}
    \toprule
    Term & Reward Expression & Weight \\
    \midrule
    DoF Position & $\exp(-0.7|\mathbf{q}_\text{tar}-\mathbf{q}|)$ & 3.0 \\
\rowcolor{gray!20}    Keypoint Position & $\exp(-|\mathbf{t}_\text{tar}-\mathbf{t}|)$ & 2.0 \\
    Root Linear Velocity & $\exp(-4.0|\mathbf{v}_\text{tar}-\mathbf{v}|)$ & 6.0 \\
   \rowcolor{gray!20} Root Roll \& Pitch & $\exp(-|\Omega^{\phi\theta}_\text{tar}-\Omega^{\phi\theta}|)$ & 1.0 \\
    Root Yaw & $\exp(-|\Delta y|)$ & 1.0 \\
    \bottomrule
  \end{tabular}
  \caption{\textbf{Imitation Rewards}.}
  \label{tab:imitation_rewards}
\end{table}

\begin{table}
  \footnotesize
  \centering
  \begin{tabular}{@{}l|c|c@{}}
    \toprule
    Term & Reward Expression & Weight \\
    \midrule
    Height & $\text{max}(|\textbf{h}_\text{feet}|-0.2,0)$ & 2.0 \\
\rowcolor{gray!20}    Time in Air & $\sum t_i^\text{air}\times\mathrm{1}_\text{new contact}$ & 10.0 \\
    Drag & $\sum|\textbf{v}_i^\text{foot}|\times~\mathrm{1}_\text{new contact}$ & -0.1 \\
   \rowcolor{gray!20} Contact Force & $\mathrm{1}\{|F_i^z|\ge F_\text{th}\}\times(|F_i^z|- F_\text{th})$ & -3e-3 \\
    Stumble & $\mathrm{1}\{\exists i,|\mathbf{F}_i^{xy}|>4|F_i^{z}|\}$ & -2.0 \\
  \rowcolor{gray!20}  DoF Acceleration & $|\ddot{\mathbf{q}}|^2$ & -3e-7 \\
    Action Rate & $|\mathbf{a}_{t-1}-\mathbf{a}_t|$ & -0.1 \\
   \rowcolor{gray!20} Energy & $|\dot{\mathbf{q}}|^2$ & -1e-3 \\
    Collision & $\mathrm{1}_\text{collision}$ & -10.0 \\
   \rowcolor{gray!20} DoF Limit Violation & $\mathrm{1}_{q_i>q_\text{max}||q_i<q_\text{min}}$ & -0.1 \\
    DoF Deviation & $|\mathbf{q}_\text{default}^\text{low}-\mathbf{q}^\text{low}|^2$ & -10.0 \\
   \rowcolor{gray!20} Vertical Linear Velocity & $v_z^2$ & -1.0 \\
    Horizontal Angular Velocity & $|\mathbf{\omega}_\text{xy}|^2$ & -0.4 \\
   \rowcolor{gray!20} Projected Gravity & $|\mathbf{g}_\text{xy}|^2$ & -2.0 \\
    \bottomrule
  \end{tabular}
  \caption{\textbf{Regularization Rewards}.}
  \label{tab:reg_rewards}
\end{table}

\noindent\textbf{RL Control Policy}.
Our goal is to track the root movement goal for the whole body and the target expression goal for upper body, and our training data is introduced above.
The humanoid control policy is defined with \cref{eq:rlpolicy}.
The goal space can be formulated as $\mathcal{G}=\mathcal{G}^{e}\times\mathcal{G}^{m}$, where $\mathcal{G}^{e}$ includes joint angles and keypoint translations from the retargeting process above and the goal space for robot movement control $\mathcal{G}^{m}=\langle\mathbf{v}, rpy, h\rangle$ where $\mathbf{v}\in\mathbb{R}^3$ is the linear velocity, $rpy\in\mathbb{R}^3$ is the robot pose in terms of row/pitch/yaw and $h$ is the body height.
The observation $\mathcal{O}$ includes robot proprieoception information $o_{t}=[\omega_t,r_t,p_t,\Delta y, q_t, \dot{q}_t, \mathbf{a}_{t-1}]^{T}$ where $\omega_t$ is robot root angular velocity, $r_t,p_t$ is roll and pitch, $\Delta y=y_t-y$ is the difference between current and desired yaw angle, $q_t$ and $\dot{q}_t$ is the joint position and angular velocity and $\mathbf{a}_{t}\in\mathbb{R}^{27}$ is the target position of the joint proportional-derivative (PD) controllers.

\noindent\textbf{Training Rewards}.
In each step, the reward from the environment consists of motion rewards, root tracking rewards and regularization terms.
To protect the fragile ankle roll joints on the robot hardware, we set the actions of the two joints to zero every simulation step.
Motion rewards include DoF position reward and keypoint position reward, and root tracking rewards include root linear velocity reward, root roll \& pitch reward and root yaw reward.

The imitation rewards, including motion rewards and root tracking rewards, are listed in \cref{tab:imitation_rewards}, where $\mathbf{q}_\text{tar}, \mathbf{q} \in \mathbb{R}^9$ are the target and actual upper body DoF positions, $\mathbf{t}_\text{tar}, \mathbf{t} \in \mathbb{R}^{18}$ are the target and actual upper body keypoint positions, $\mathbf{v}_\text{tar}, \mathbf{v} \in \mathbb{R}$ are the target and actual root velocity, $\Omega^{\phi\theta}_\text{tar}, \Omega^{\phi\theta}$ are the target and actual body roll and pitch.

The regularization rewards are listed in \cref{tab:reg_rewards}, where $h_\text{feet}$ is feet height, $t_i^\text{air}$ denotes the duration for which each foot remains in the air, $\mathrm{1}_\text{new contact}$ means new foot contact with the ground, $\mathbf{F}_i^{xy}, F_i^{z},  F_\text{th}$ are foot contact force in horizontal plane and along the z-axis, and the contact force threshold respectively, $\dot{\mathbf{q}}, \ddot{\mathbf{q}}$ are joint velocity and acceleration, $\mathbf{a}_t$ is action at timestep $t$, $\mathrm{1}_\text{collision}$ denotes self-collision, $q_\text{max}, q_\text{min}$ are limits for joint positions, and $\mathbf{g}_\text{xy}$ is gravity vector projected on horizontal plane.

\begin{table}
  \centering
  \begin{tabular}{@{}l|c@{}}
    \toprule
    Hyperparameter & Value \\
    \midrule
    Discount Factor & 0.99 \\
\rowcolor{gray!20}    GAE parameter & 0.95 \\
    Timesteps per Rollout & 21 \\
 \rowcolor{gray!20}   Epochs per Rollout & 5 \\
    Minibatches per Epoch & 4 \\
   \rowcolor{gray!20} Extropy Bonus ($\alpha_2$) & 0.01 \\
    Value Loss Coefficient ($\alpha_1$) & 1.0 \\
  \rowcolor{gray!20}  Clip Range & 0.2 \\
    Reward Normalization & Yes \\
  \rowcolor{gray!20}  Learning Rate & 1e-3 \\
    \# Environments & 6192 \\
  \rowcolor{gray!20}  Optimizer & Adam \\
    \bottomrule
  \end{tabular}
  \caption{\textbf{Training Parameters}.}
  \label{tab:training_param}
\end{table}

\noindent\textbf{Training Parameters}.
We use PPO with hyperparameters listed in \cref{tab:training_param} to train the policy.

\section{Details on Humanoid-X Dataset}
In this section, we will introduce the Humanoid-X dataset. We will introduce the data format and structure and show several examples of the dataset.
\subsection{Data Format and Structure}
For each motion sample in Humanoid-X, we expand them to the 5 data modalities introduced in Sec. 3.1, where they are described with $\langle \mathcal{V}, \mathcal{T}, \mathcal{P}_{human}, \mathcal{P}_{robot}, \mathcal{A}_{robot} \rangle$.
Visualization of part of the data samples in the dataset will be shown in \cref{dataset_vis}.

\noindent\textbf{Motion Video Clip $\mathcal{V}$}.
The video clips are collected in MP4 format at a frame rate of 20 frames per second (fps).

\noindent\textbf{Text Description $\mathcal{T}$}.
The text descriptions are stored in plain text (\textit{.txt}) format.

\noindent\textbf{Human Poses $\mathcal{P}_{human}$}.
The human poses are sequences of SMPL model parameters with a frame rate of 20 fps.
We stored the collected data for each motion sample in a NumPy (\textit{.npy}) file.

\noindent\textbf{Humanoid Keypoints $\mathcal{P}_{robot}$}.
The humanoid keypoints include humanoid robot DoFs $q_{robot}$ and humanoid robot root states $s_{root}$.
Each frame of the data contains 27 DoFs of the robot configuration and a 7-dimensional root state vector, consisting of 3-DoF root translation and 4-DoF quaternion representation for root orientation.
The humanoid keypoints are recorded with a frame rate of 20 fps.
We stored the collected data (27 robot DoFs and 7-DoF root state) for each motion sample in a NumPy (\textit{.npy}) file for efficient data management and processing.

\noindent\textbf{Humanoid Actions $\mathcal{A}_{robot}$}.
The humanoid actions are sequences of target DoF positions.
The data is collected and stored at 50 fps, with each frame containing 27 robot DoFs that correspond to the robot's physical configuration.
We stored the collected data for each motion sample in a NumPy (\textit{.npy}) file.

\subsection{Data Statistics}
\begin{figure}[htbp] % Use [htbp] for positioning
  \centering
  % First subfigure
  \begin{subfigure}[b]{0.48\linewidth} % Adjust width as needed
    \centering
    \includegraphics[width=\linewidth]{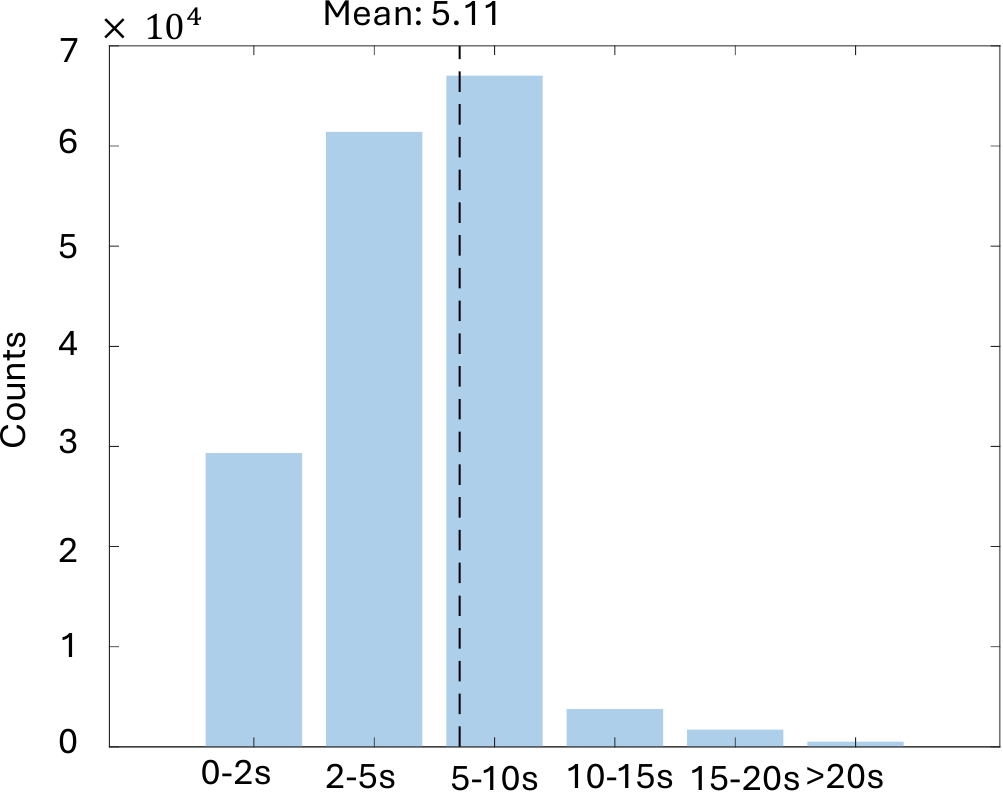}
    \caption{{Video Length}}
    \label{fig:vid_distribution}
  \end{subfigure}
  % Second subfigure
  \begin{subfigure}[b]{0.48\linewidth} % Adjust width as needed
    \centering
    \includegraphics[width=\linewidth]{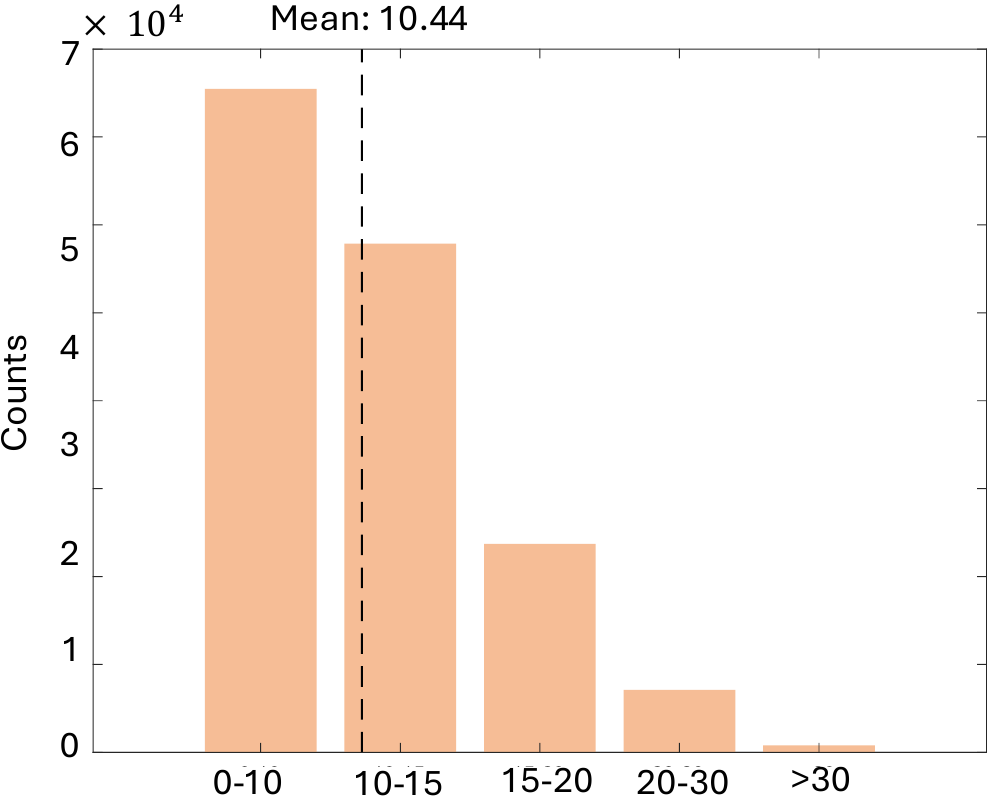}
    \caption{{Captioning Sentence Length}}
    \label{fig:sentence_distribution}
  \end{subfigure}
  \caption{Distribution of video length (in seconds) and captioning sentence length (in words), with the dotted line representing the average length.}
  \label{fig:combined_distribution}
\end{figure}

% \todo{siqi can ask tianheng to provide some images: distribution of sentence length, distribution of video length, word counts of different part of speech and their corresponding word cloud}
\noindent\textbf{Sequence Length Analysis}.
We conduct comprehensive statistical analysis on both video sequence durations and their corresponding caption lengths, as illustrated in \cref{fig:vid_distribution} and \cref{fig:sentence_distribution}.
The analysis reveals that the majority of video clips are relatively short, with durations less than 10 seconds. This distribution pattern stems from our video segmentation strategy, where clips are specifically extracted when significant or meaningful motion patterns are detected within the continuous recordings.
This approach naturally results in shorter, more focused segments, making longer clips relatively rare in our dataset.
Regarding the textual descriptions, the distribution of caption lengths shows that most sentences contain fewer than 20 words. This concise nature of captions aligns with our guidelines, which emphasized brevity while maintaining descriptive accuracy.
% Such moderate-length descriptions strike a balance between providing sufficient detail to capture the essential motion characteristics and maintaining clarity for potential downstream applications.

\noindent\textbf{Vocabulary Analysis}.
To gain deeper insights into the linguistic composition of video captions, we conduct a comprehensive analysis of different parts of speech, focusing on nouns, verbs, adjectives, and adverbs. This grammatical categorization helps understand how motions and actions are described in our dataset. \cref{tab:vocab_sizes} presents the vocabulary size distribution across these grammatical categories, providing a quantitative view of the linguistic diversity in our annotations. The analysis reveals the richness of descriptive elements used in capturing robot motions and their contextual information.

\begin{figure*}
  \centering
   \includegraphics[width=1.0\linewidth]{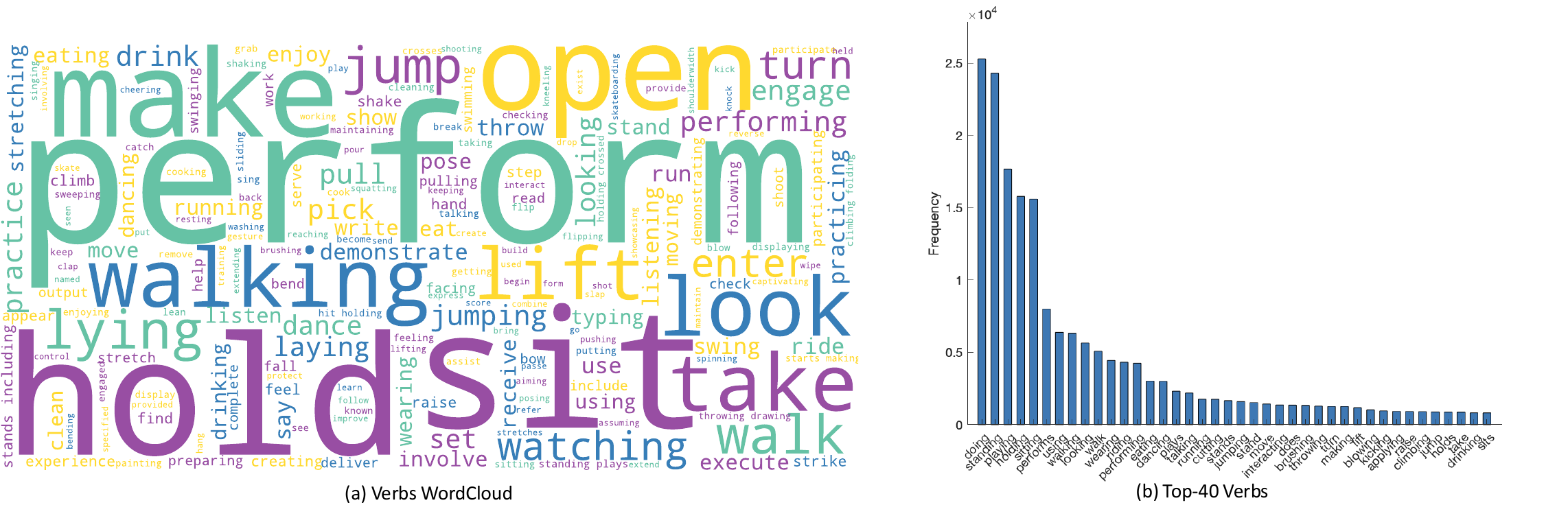}
   \caption{{\textit{Verbs} Word Cloud and Top-40 Frequent \textit{Verbs}}.}
   \label{fig:verb_freq}
\end{figure*}

For verbs, the word cloud and the top-40 frequent words are shown in \cref{fig:verb_freq}. It can be seen that verbs like ``doing'', ``standing'', ``playing'', ``holding'' and ``performing'' occur with a relatively high frequency. This implicitly matched the expectation since these words are heavily used as the prompt for video collection.

\begin{figure*}
  \centering
   \includegraphics[width=1.0\linewidth]{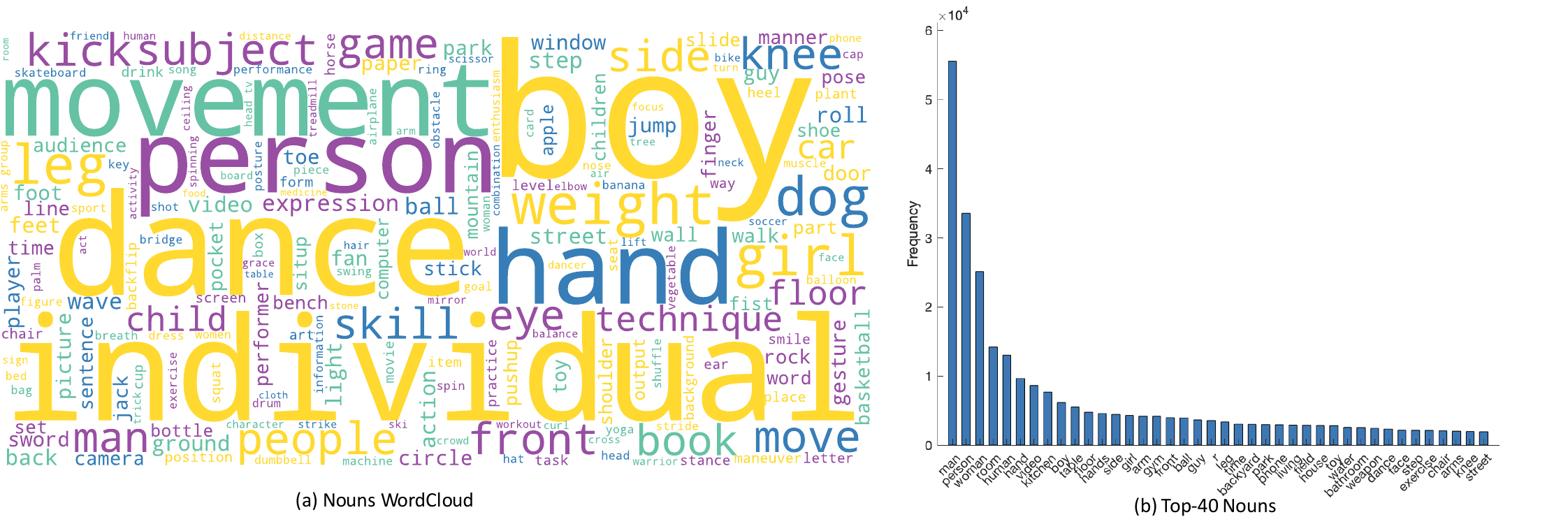}
   \caption{{\textit{Nouns} Word Cloud and Top-40 Frequent \textit{Nouns}}.}
   \label{fig:noun_cloud}
\end{figure*}

For nouns, the word cloud and the top-40 frequent words are shown in \cref{fig:noun_cloud}. The top 3 frequent words occurred are ``man'', ``person'' and ``woman''. This is also expected given the prompt for caption since it is specifically mentioned that the description should indicate what the man or woman is doing.

\begin{figure*}
  \centering
   \includegraphics[width=1.0\linewidth]{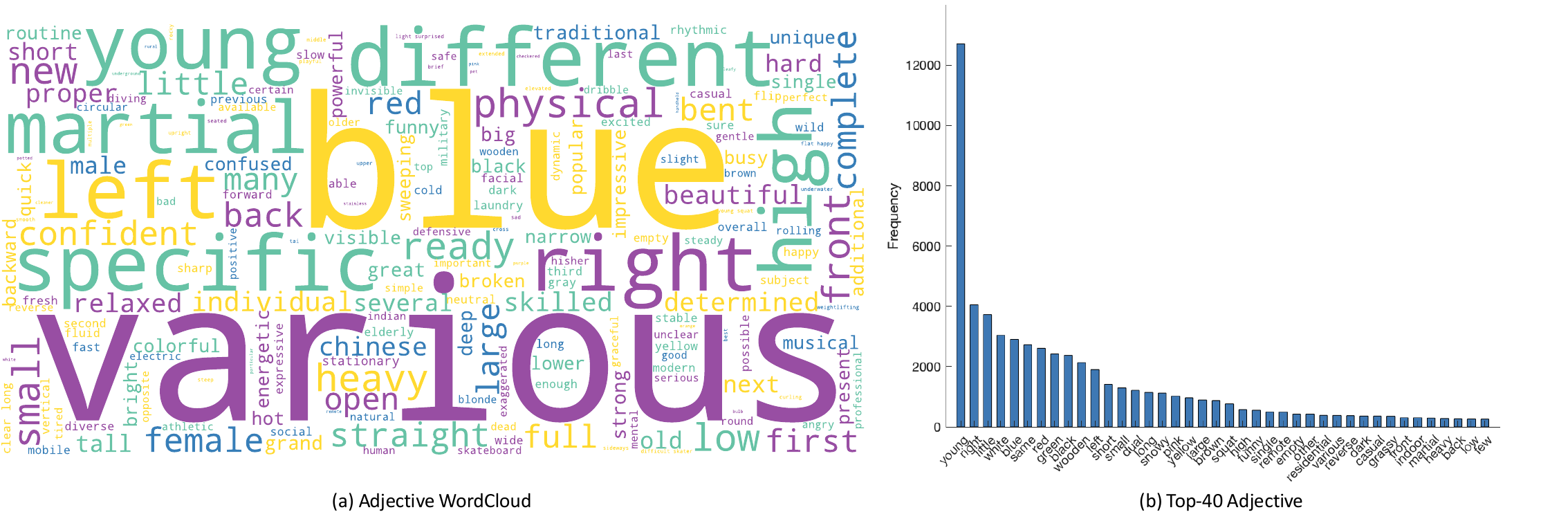}
   \caption{{\textit{Adjectives} Word Cloud and Top-40 Frequent \textit{Adjectives}}.}
   \label{fig:adj_cloud}
\end{figure*}

\begin{figure*}
  \centering
   \includegraphics[width=1.0\linewidth]{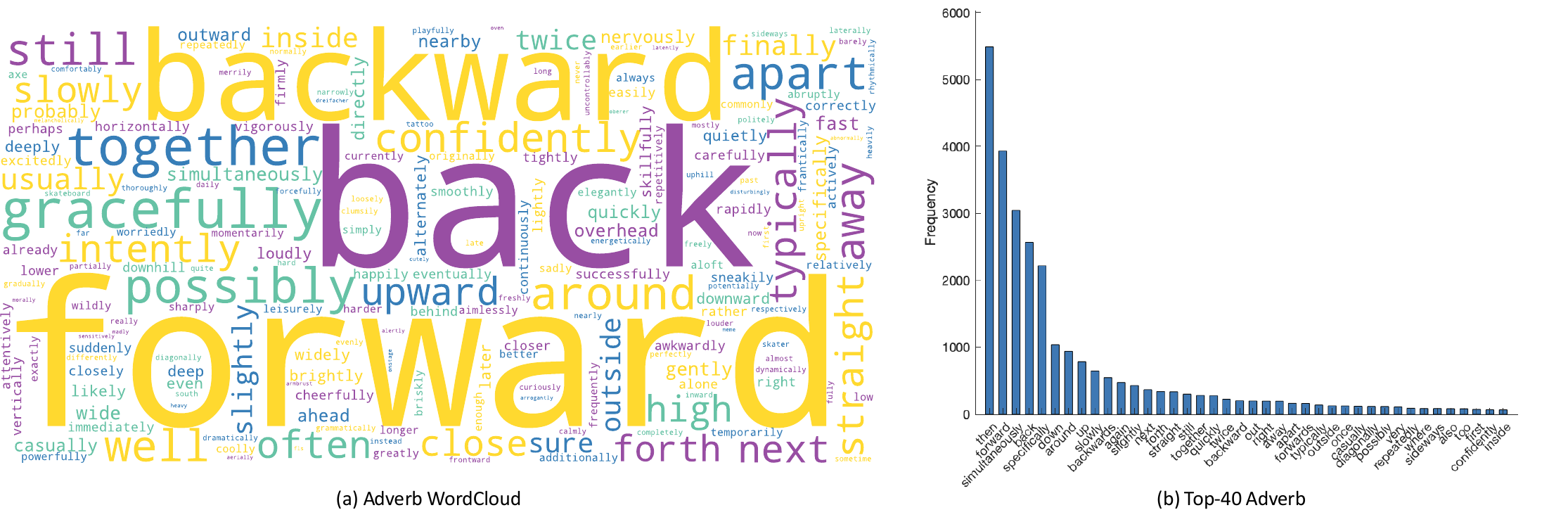}
   \caption{{\textit{Adverbs} Word Cloud and Top-40 Frequent \textit{Adverbs}}.}
   \label{fig:adv_cloud}
\end{figure*}

For adjectives and adverbs, their word cloud and top-40 frequent words are shown in \cref{fig:adj_cloud} and \cref{fig:adv_cloud}. It can been seen that most frequent adverbs are mostly about direction of motions and most frequent adjectives are mostly above color. This is cause by the fact that we explicitly instruct the Video-LLaMA to be concise so that there would not be redundant words for non-motion-related contents.

% \begin{figure}
%   \centering
%    \includegraphics[width=1.0\linewidth]{fig/vid_length_cropped.pdf}
%    \caption{\textbf{Video Length Distribution}.}
%    \label{fig:vid_distribution}
% \end{figure}
% \begin{figure}
%   \centering
%    \includegraphics[width=1.0\linewidth]{fig/sentence_length_cropped.pdf}
%    \caption{\textbf{Captioning Sentence Length Distribution}.}
%    \label{fig:sentence_distribution}
% \end{figure}
\begin{table}
  \centering
  \begin{tabular}{@{}l|c@{}}
    \toprule
    Part of Speech & Vocabulary Size \\
    \midrule
    Verbs & 3206 \\
\rowcolor{gray!20}    Nouns & 6048 \\
    Adjectives & 1526 \\
   \rowcolor{gray!20} Adverbs & 590 \\
    Others & 527 \\
    \midrule
    Total & 11897 \\
    \bottomrule
  \end{tabular}
  \caption{Vocabulary Sizes for Each Part of Speech.}
  \label{tab:vocab_sizes}
\end{table}

\subsection{Data Preparation and Release}
We will fully release our data and code in the future, without violating the ethics concerns stated in \cref{sec:ethics}.


\subsection{Data examples from Humanoid-X Dataset}
\label{dataset_vis}
\begin{figure*}[h]
  \centering
   \includegraphics[width=0.95\linewidth]{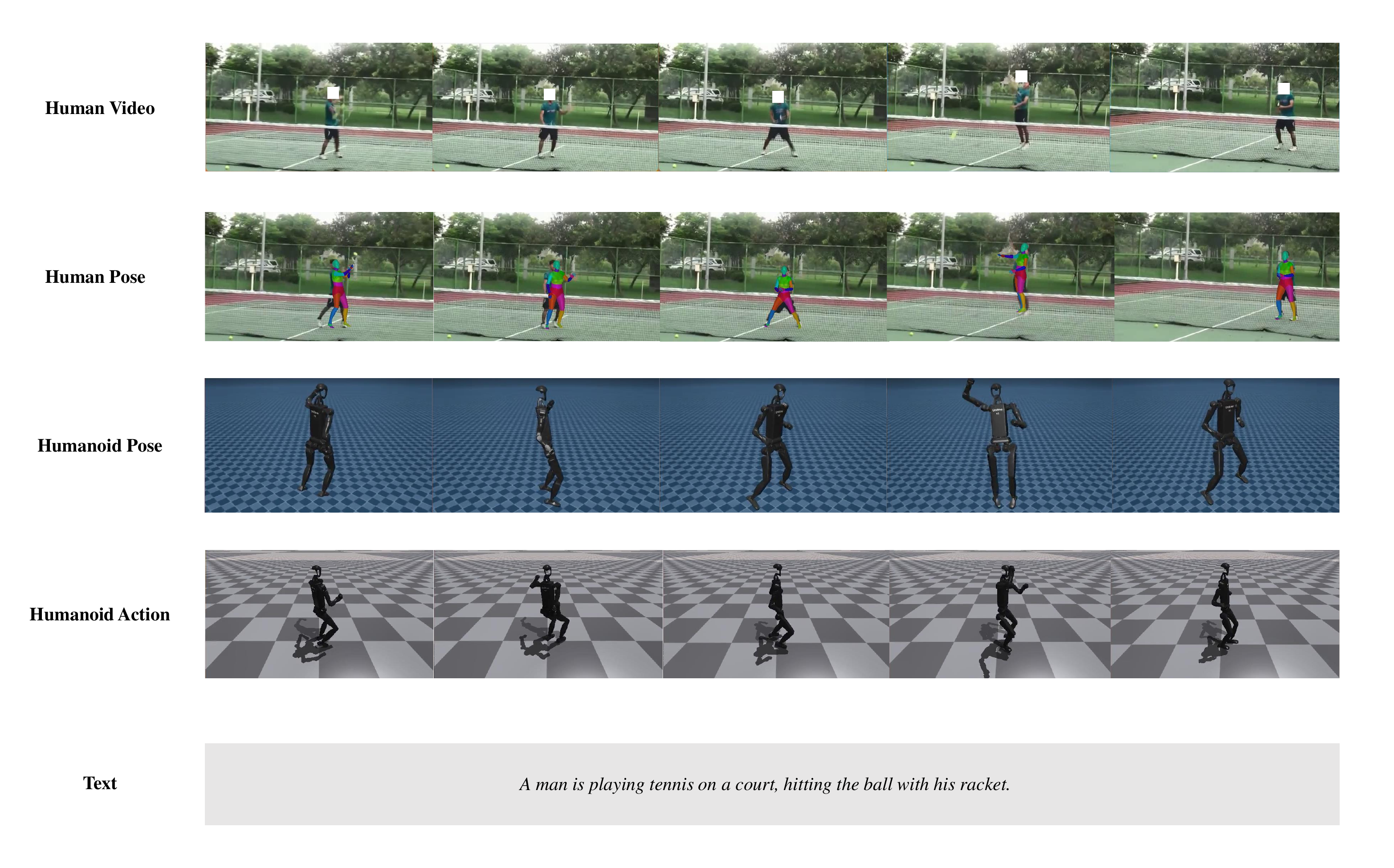}
   \caption{\textbf{Data samples in Humanoid-X}.}
   \label{fig:sample1}
\end{figure*}

\begin{figure*}[h]
  \centering
   \includegraphics[width=0.95\linewidth]{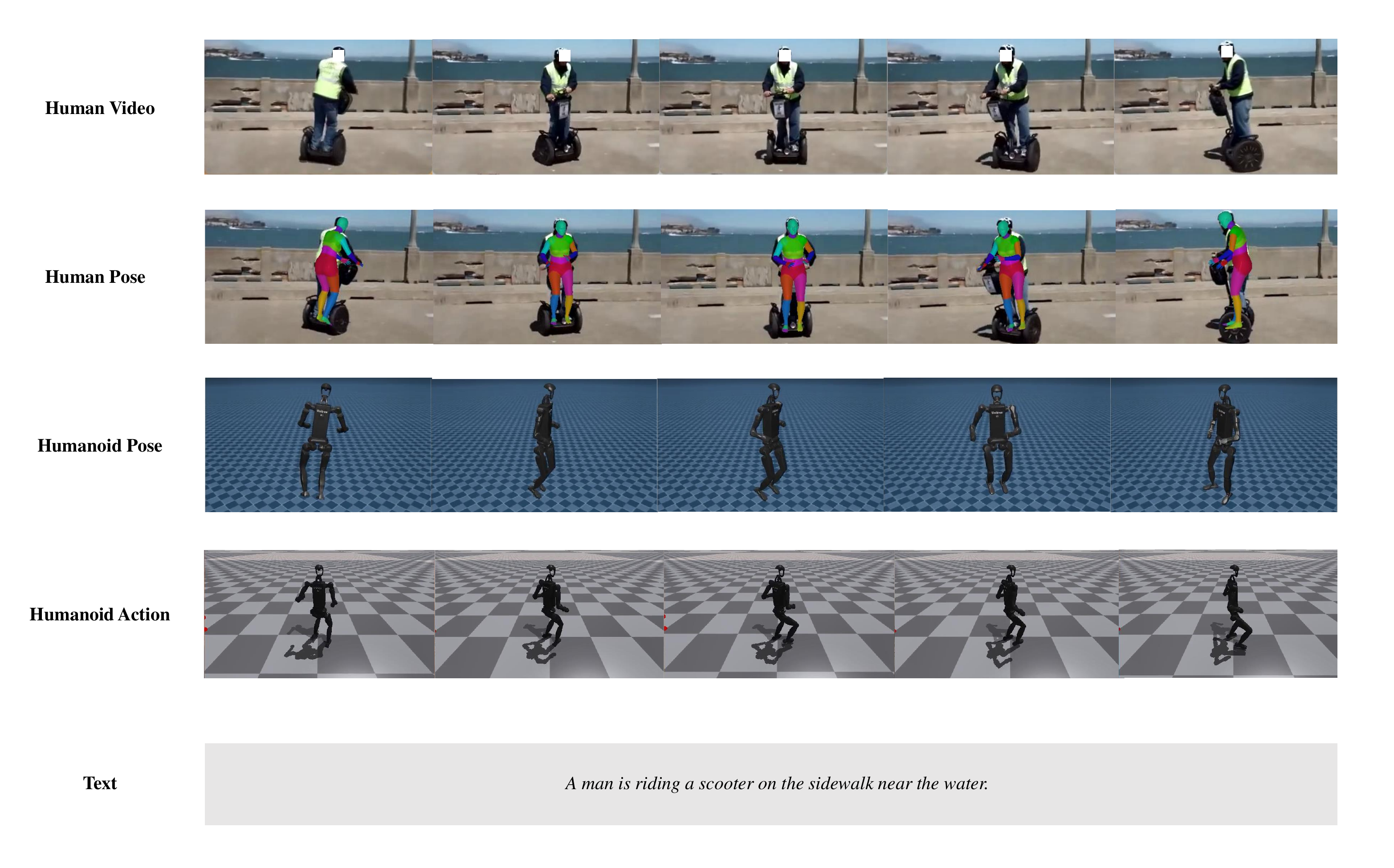}
   \caption{\textbf{Data examples in Humanoid-X}.}
   \label{fig:sample2}
\end{figure*}

\begin{figure*}[h]
  \centering
   \includegraphics[width=0.95\linewidth]{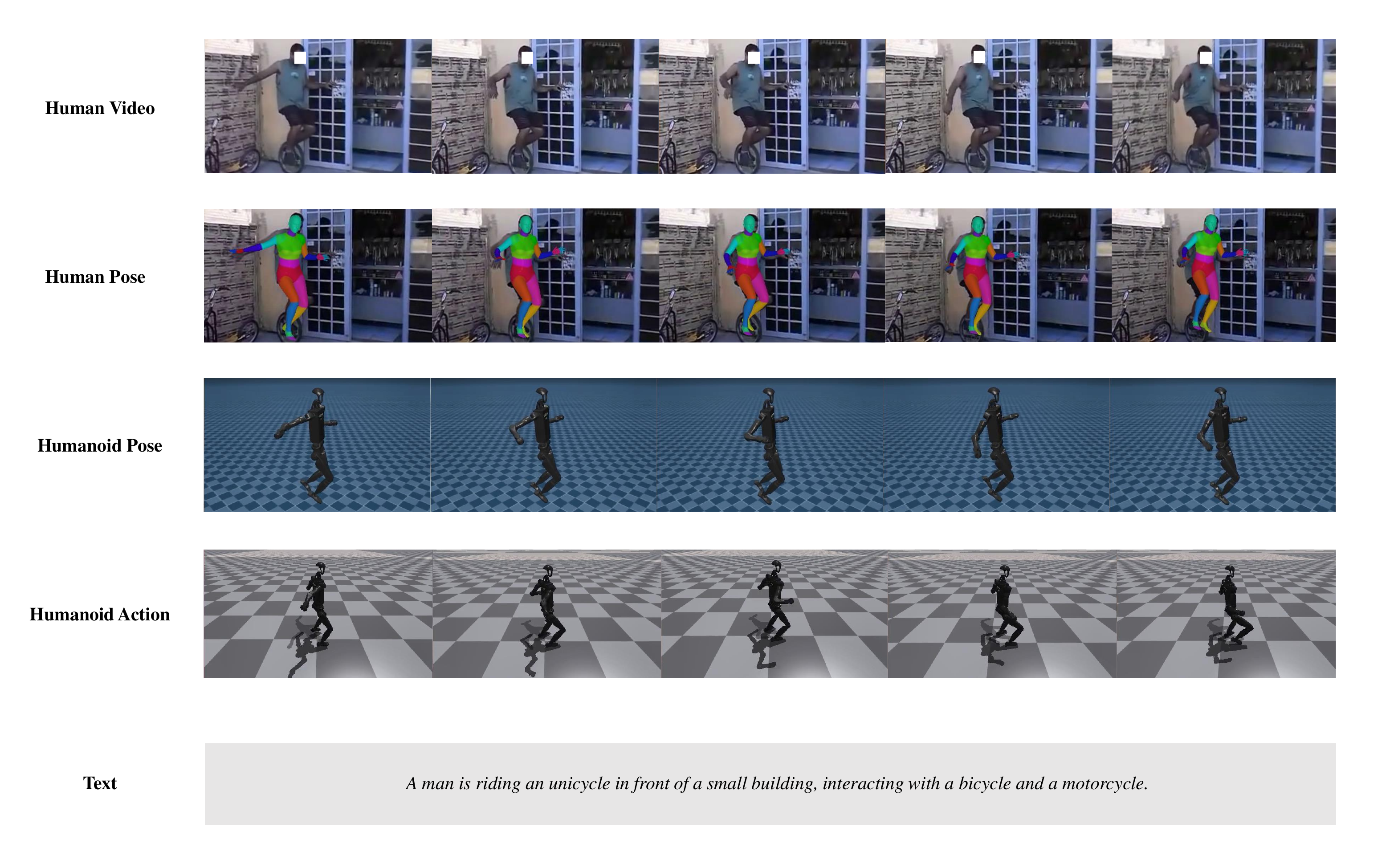}
   \caption{\textbf{Data examples in Humanoid-X}.}
   \label{fig:sample3}
\end{figure*}

\begin{figure*}[h]
  \centering
   \includegraphics[width=0.95\linewidth]{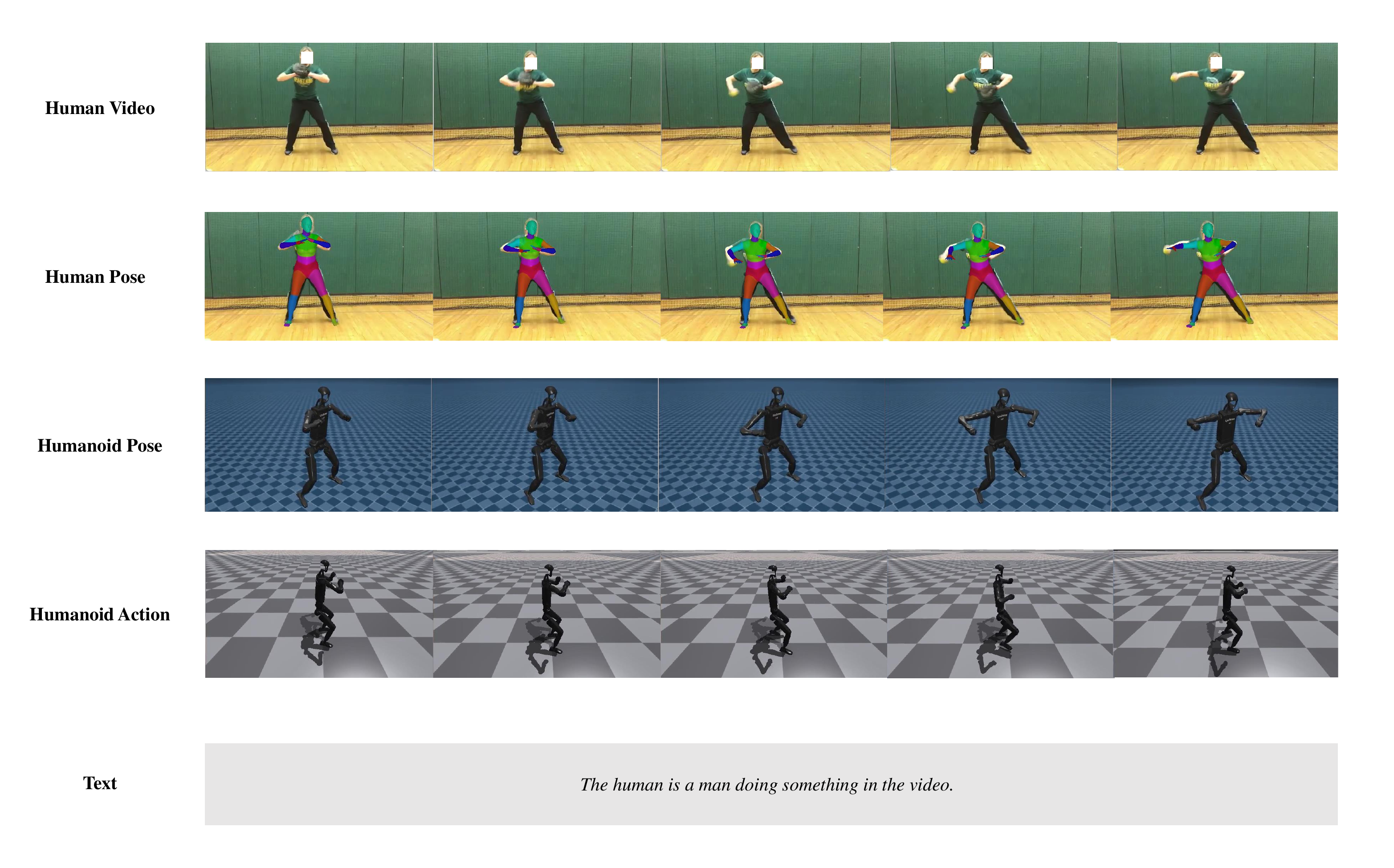}
   \caption{\textbf{Data examples in Humanoid-X}.}
   \label{fig:sample4}
\end{figure*}

\begin{figure*}[h]
  \centering
   \includegraphics[width=0.95\linewidth]{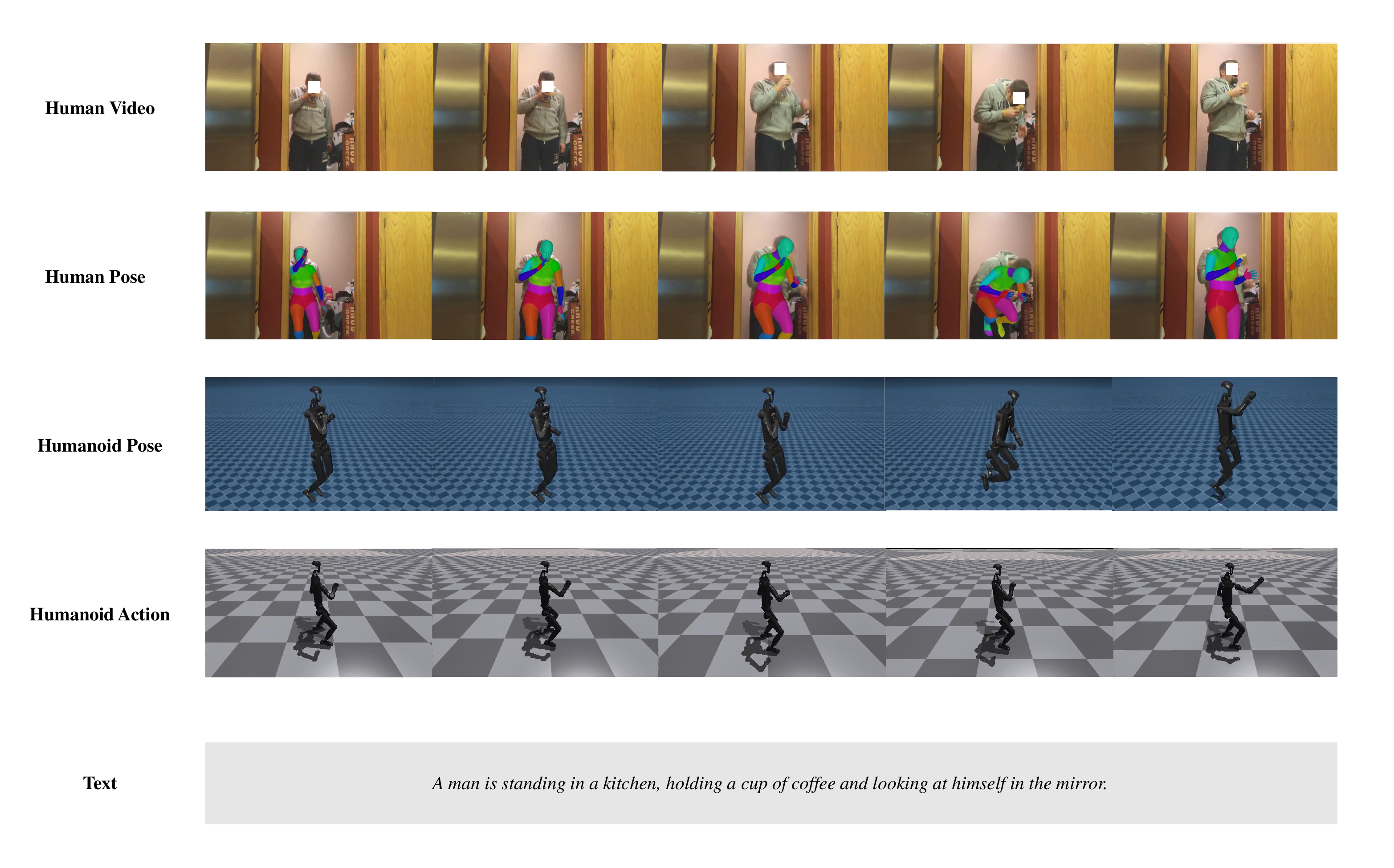}
   \caption{\textbf{Data examples in Humanoid-X}.}
   \label{fig:sample5}
\end{figure*}

\begin{figure*}[h]
  \centering
   \includegraphics[width=0.95\linewidth]{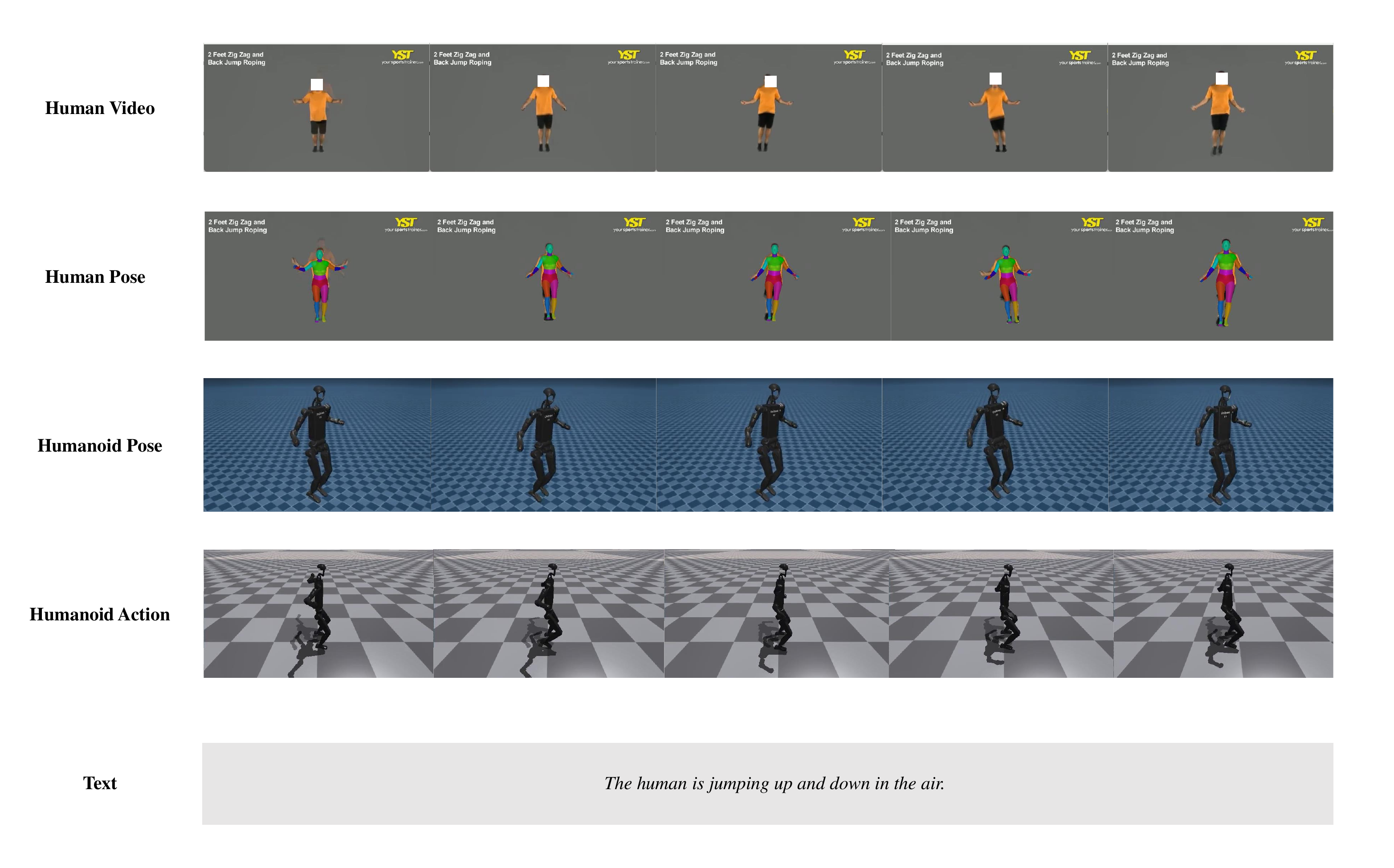}
   \caption{\textbf{Data examples in Humanoid-X}.}
   \label{fig:sample6}
\end{figure*}

\begin{figure*}[h]
  \centering
   \includegraphics[width=0.95\linewidth]{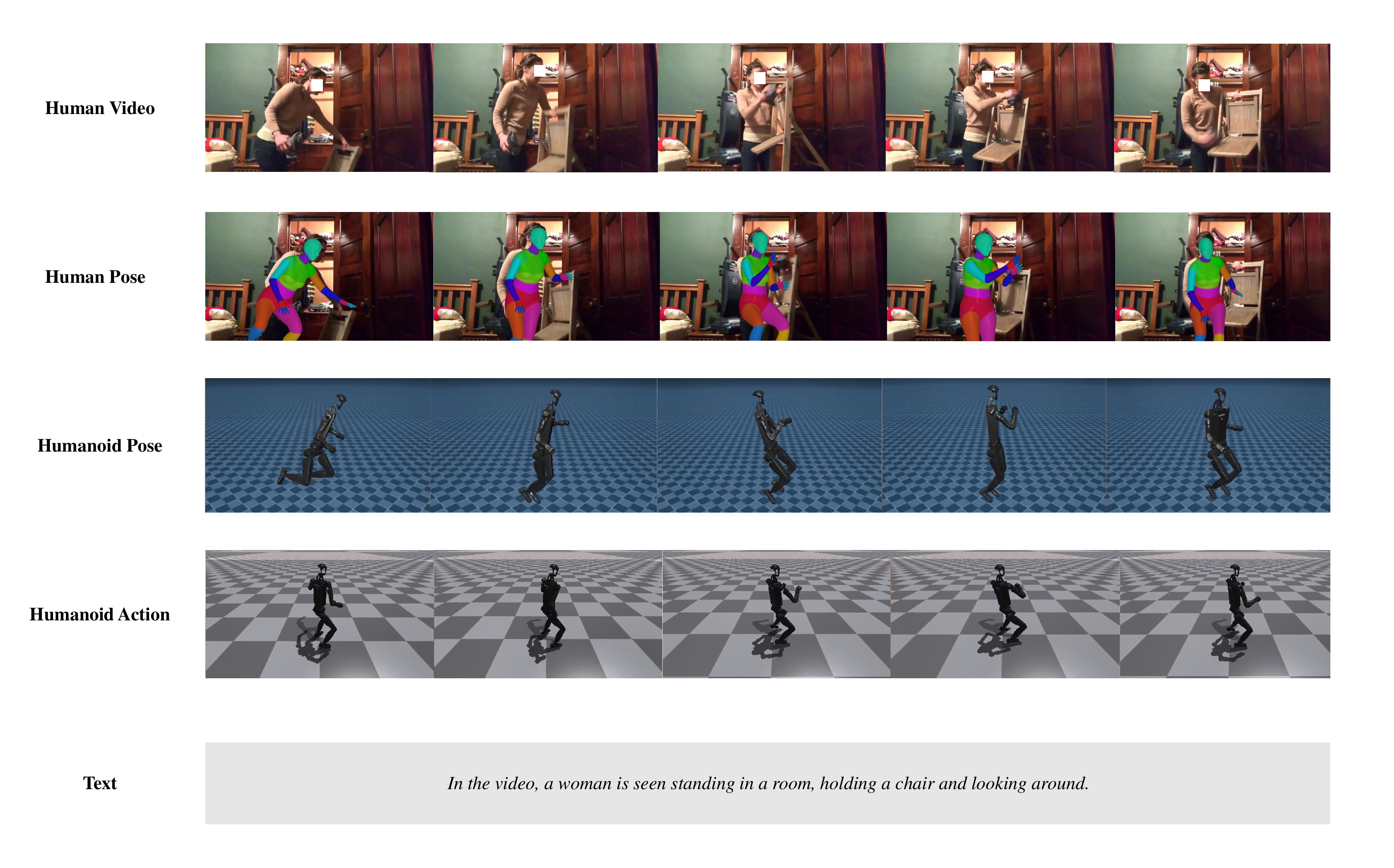}
   \caption{\textbf{Data examples in Humanoid-X}.}
   \label{fig:sample7}
\end{figure*}

\begin{figure*}[h]
  \centering
   \includegraphics[width=0.95\linewidth]{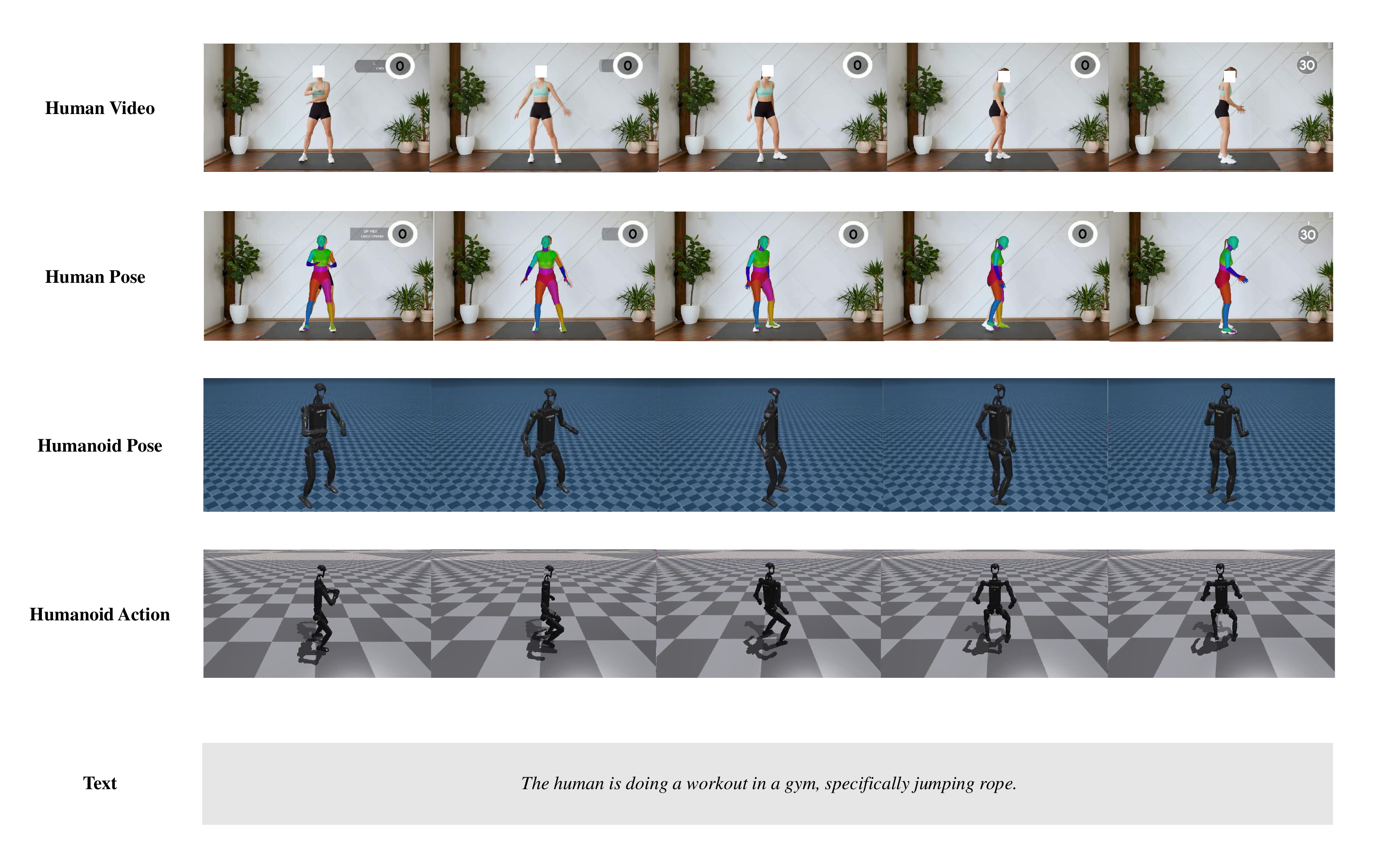}
   \caption{\textbf{Data examples in Humanoid-X}.}
   \label{fig:sample8}
\end{figure*}

\begin{figure*}[h]
  \centering
   \includegraphics[width=0.95\linewidth]{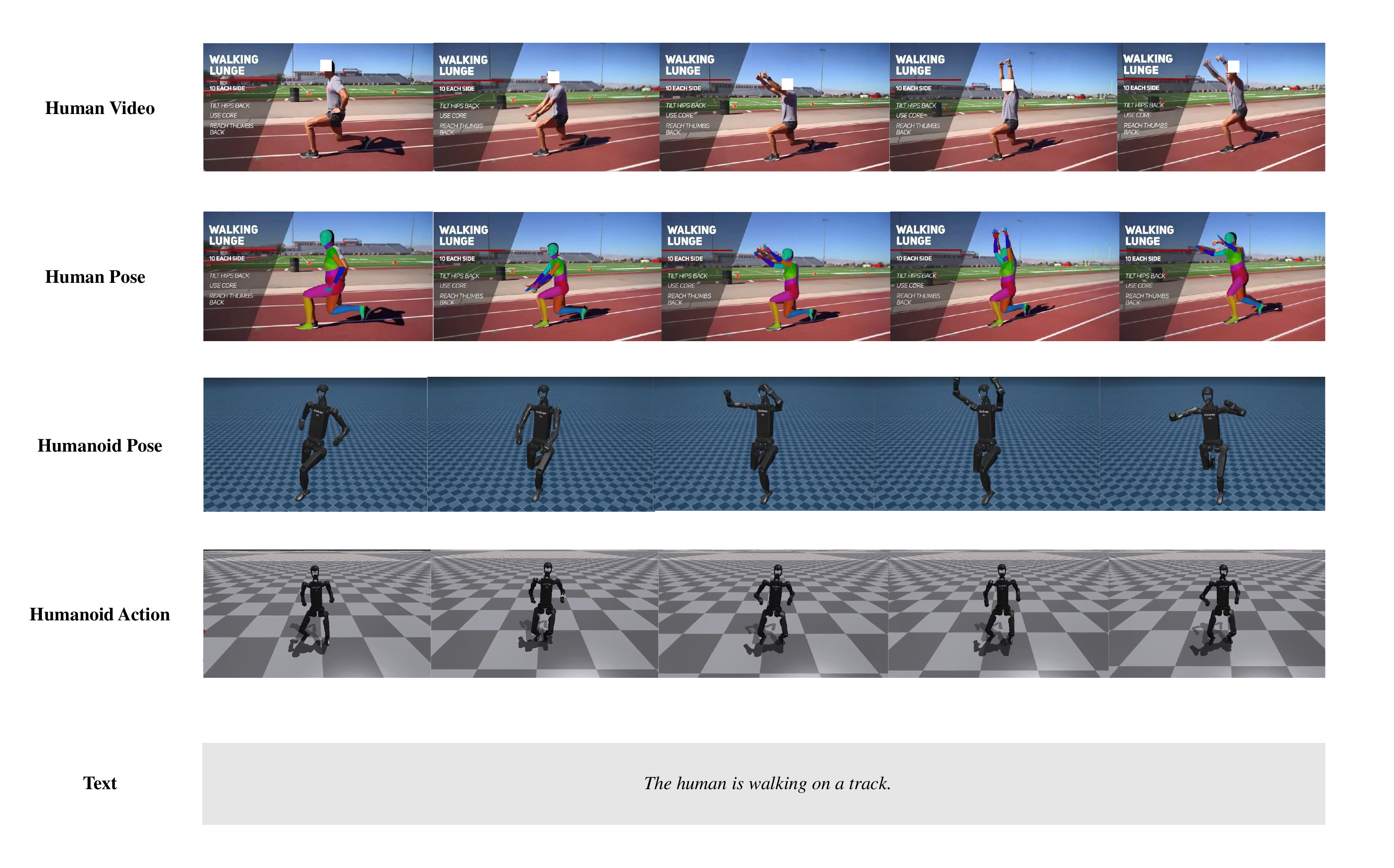}
   \caption{\textbf{Data examples in Humanoid-X}.}
   \label{fig:sample9}
\end{figure*}

\begin{figure*}[h]
  \centering
   \includegraphics[width=0.95\linewidth]{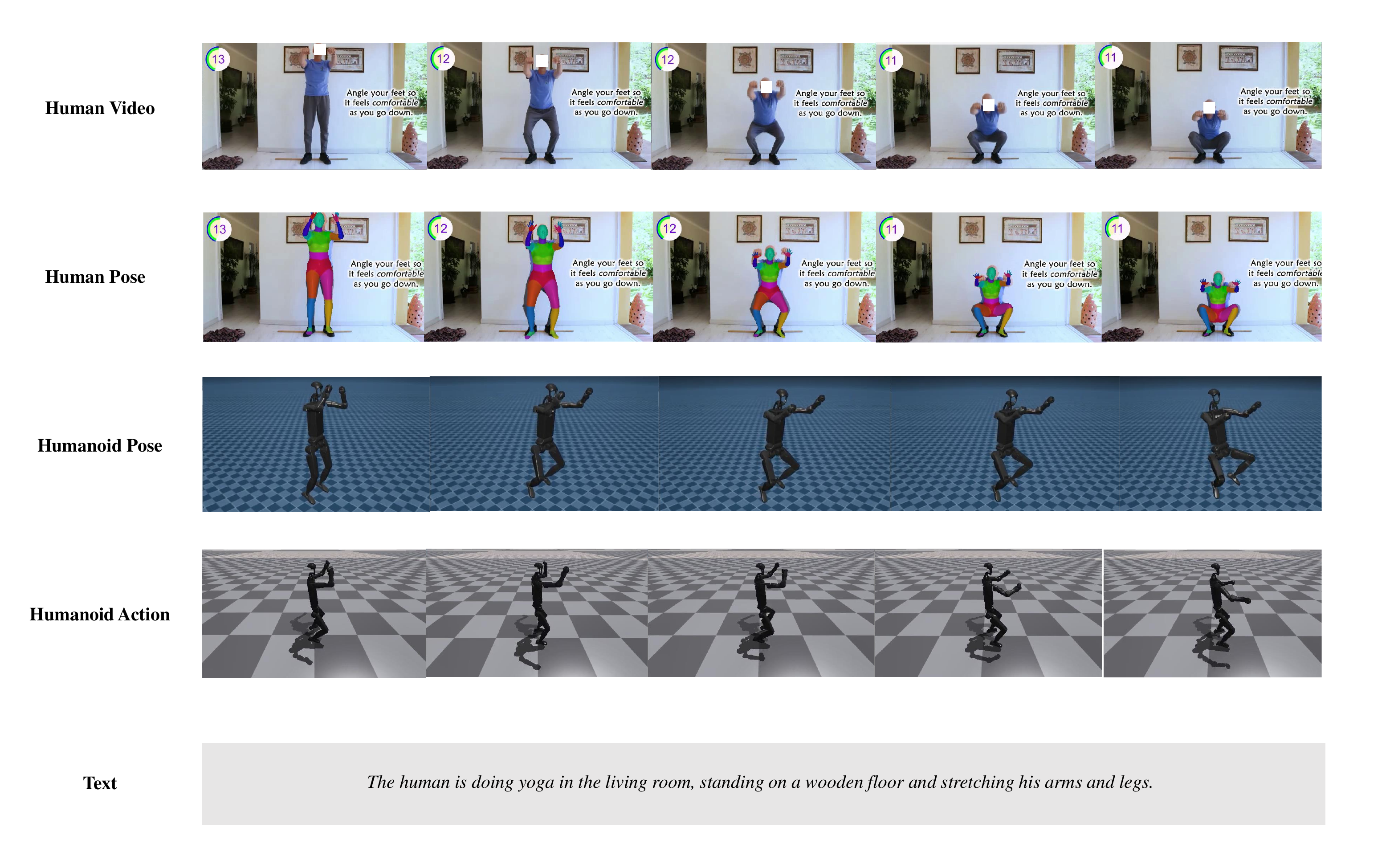}
   \caption{\textbf{Data examples in Humanoid-X}.}
   \label{fig:sample10}
\end{figure*}

\begin{figure*}[h]
  \centering
   \includegraphics[width=0.95\linewidth]{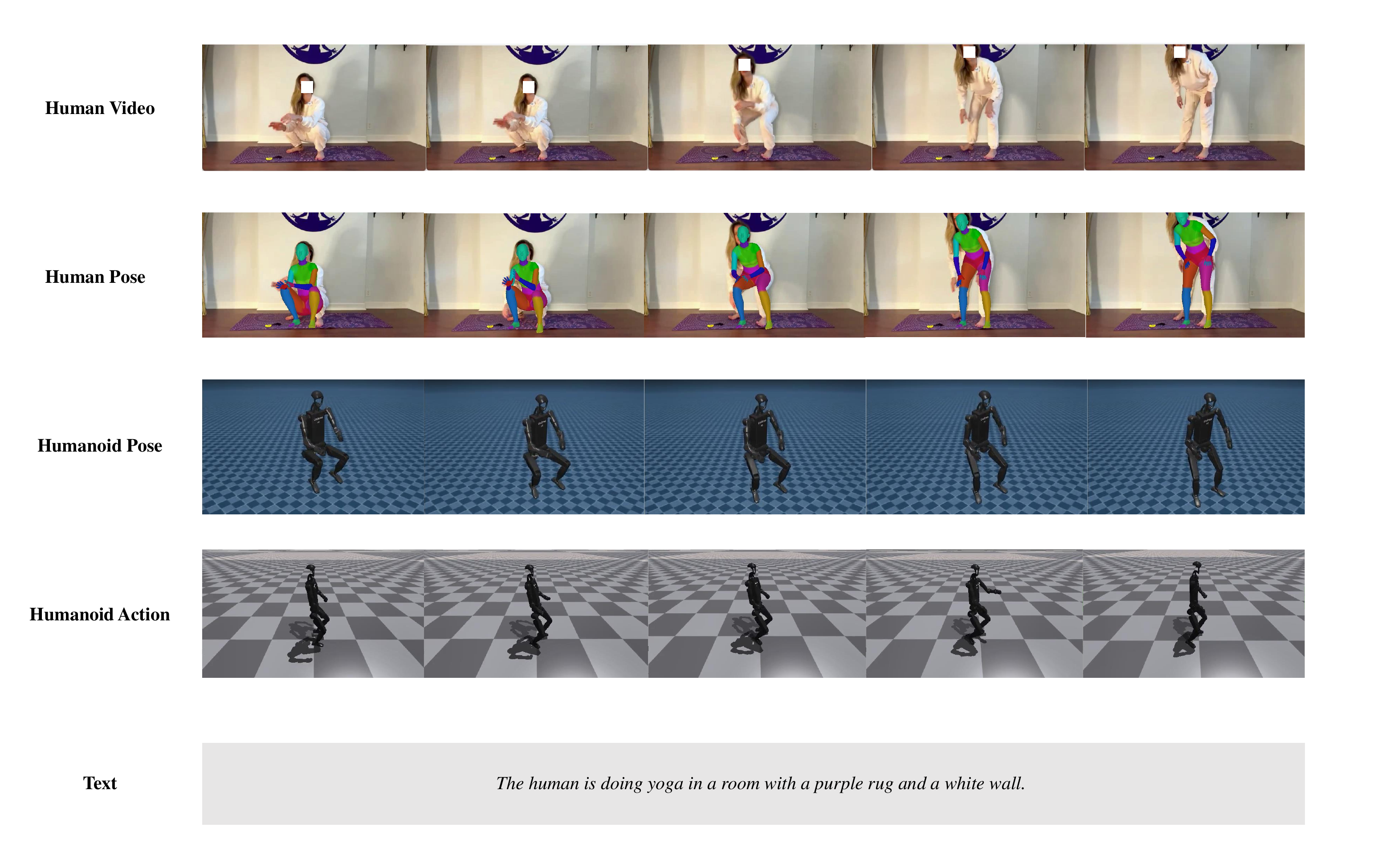}
   \caption{\textbf{Data examples in Humanoid-X}.}
   \label{fig:sample11}
\end{figure*}

\begin{figure*}[h]
  \centering
   \includegraphics[width=0.95\linewidth]{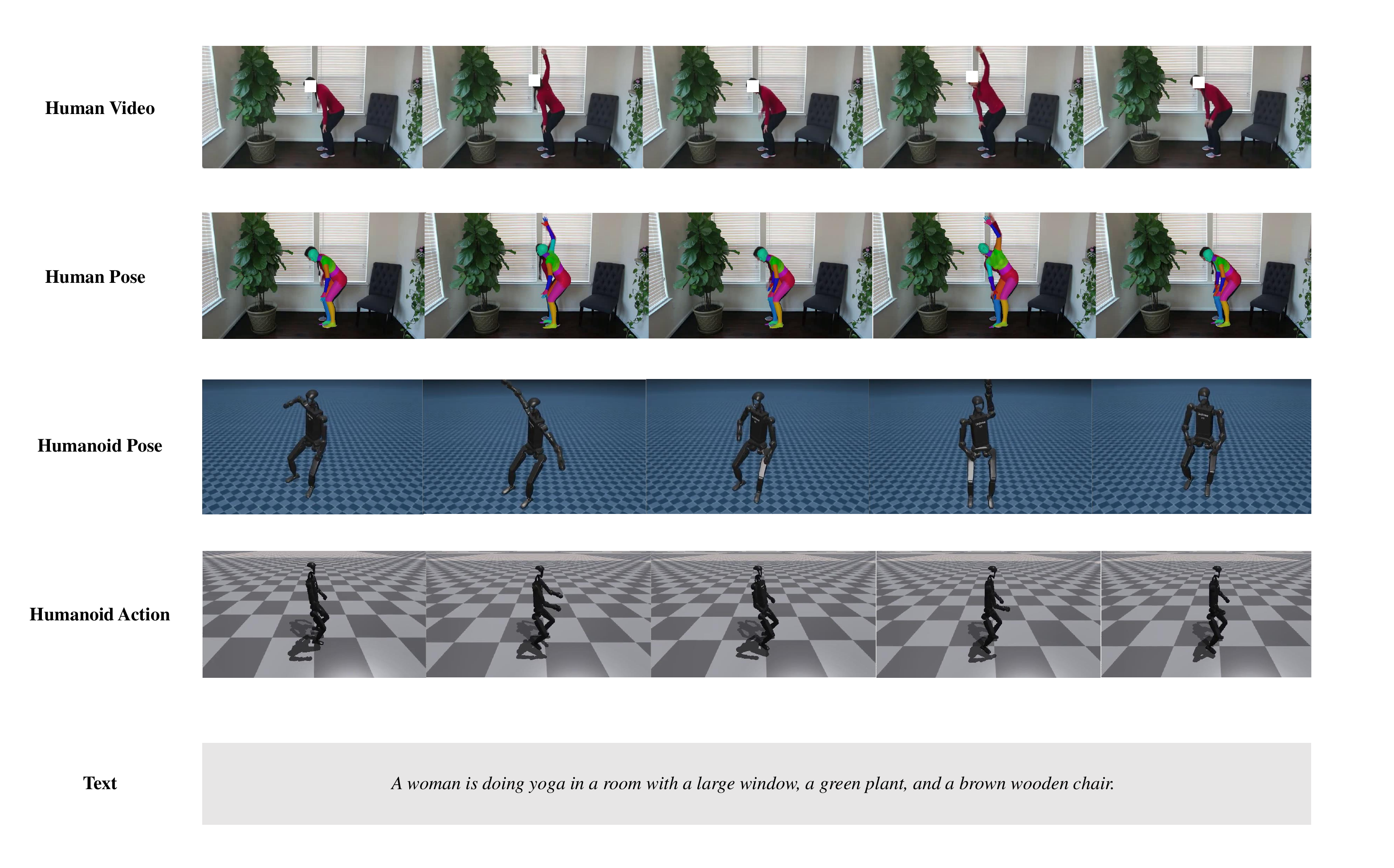}
   \caption{\textbf{Data examples in Humanoid-X}.}
   \label{fig:sample12}
\end{figure*}

\begin{figure*}[h]
  \centering
   \includegraphics[width=0.95\linewidth]{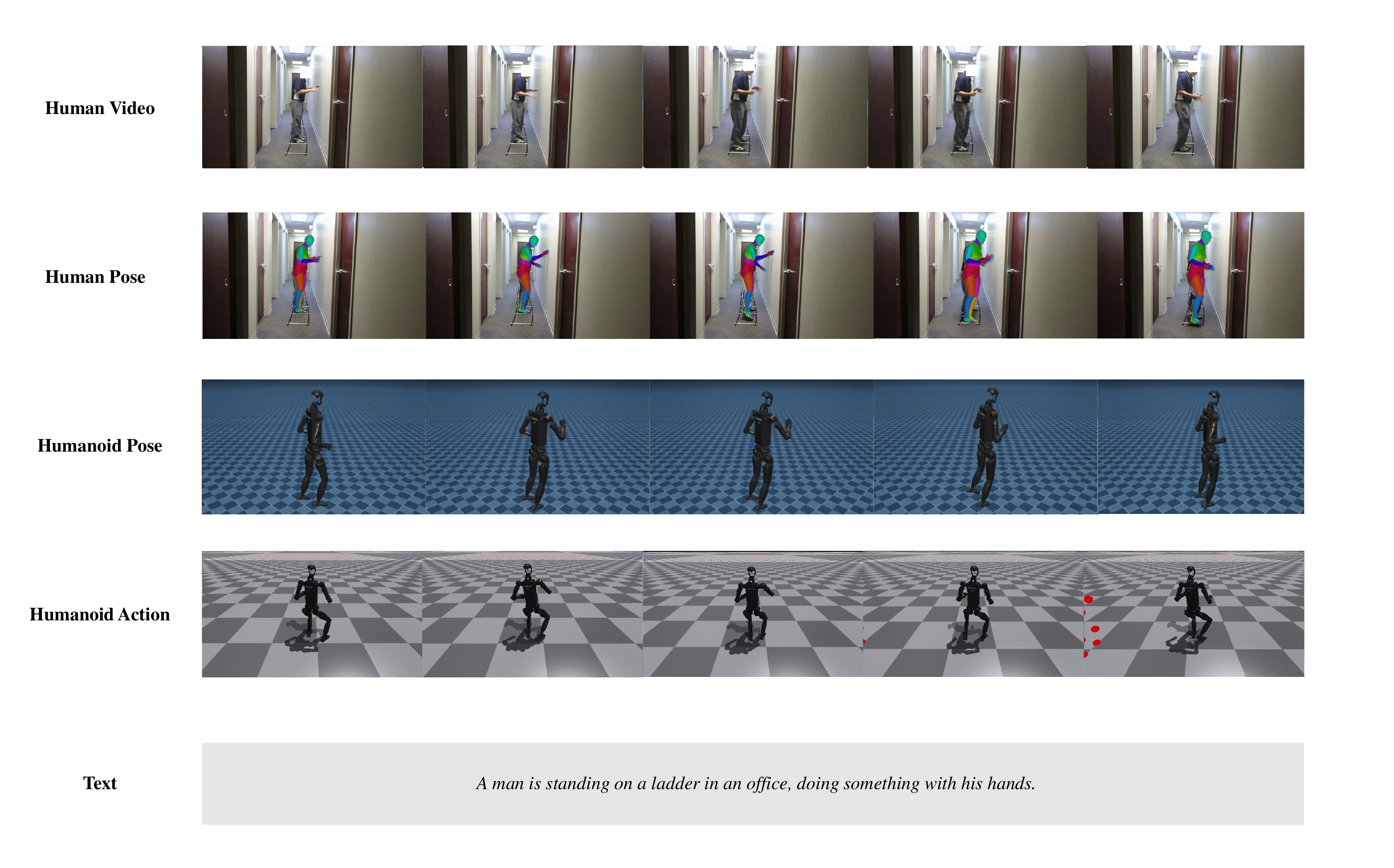}
   \caption{\textbf{Data examples in Humanoid-X}.}
   \label{fig:sample13}
\end{figure*}

\begin{figure*}[h]
  \centering
   \includegraphics[width=0.95\linewidth]{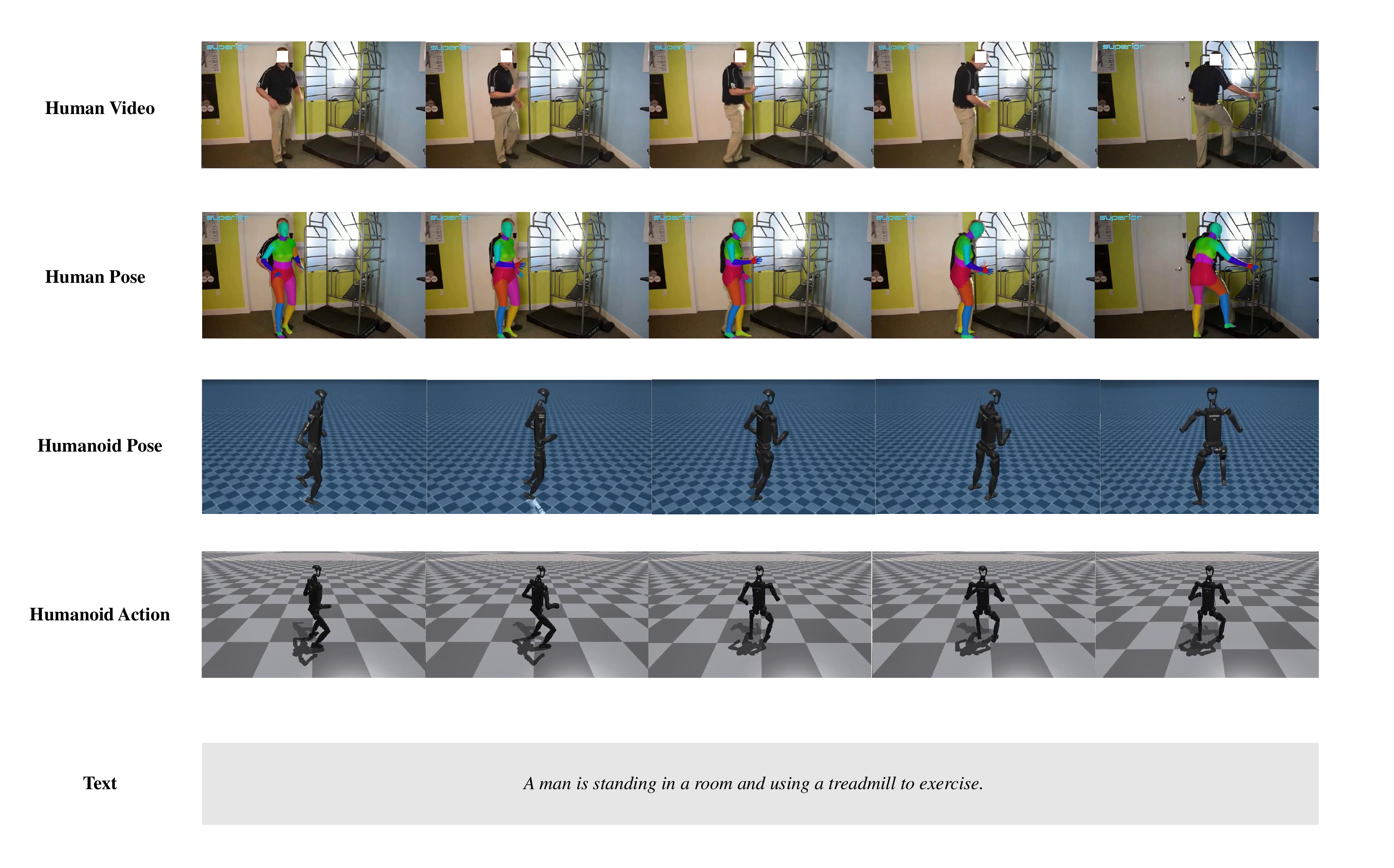}
   \caption{\textbf{Data examples in Humanoid-X}.}
   \label{fig:sample14}
\end{figure*}

\begin{figure*}[h]
  \centering
   \includegraphics[width=0.95\linewidth]{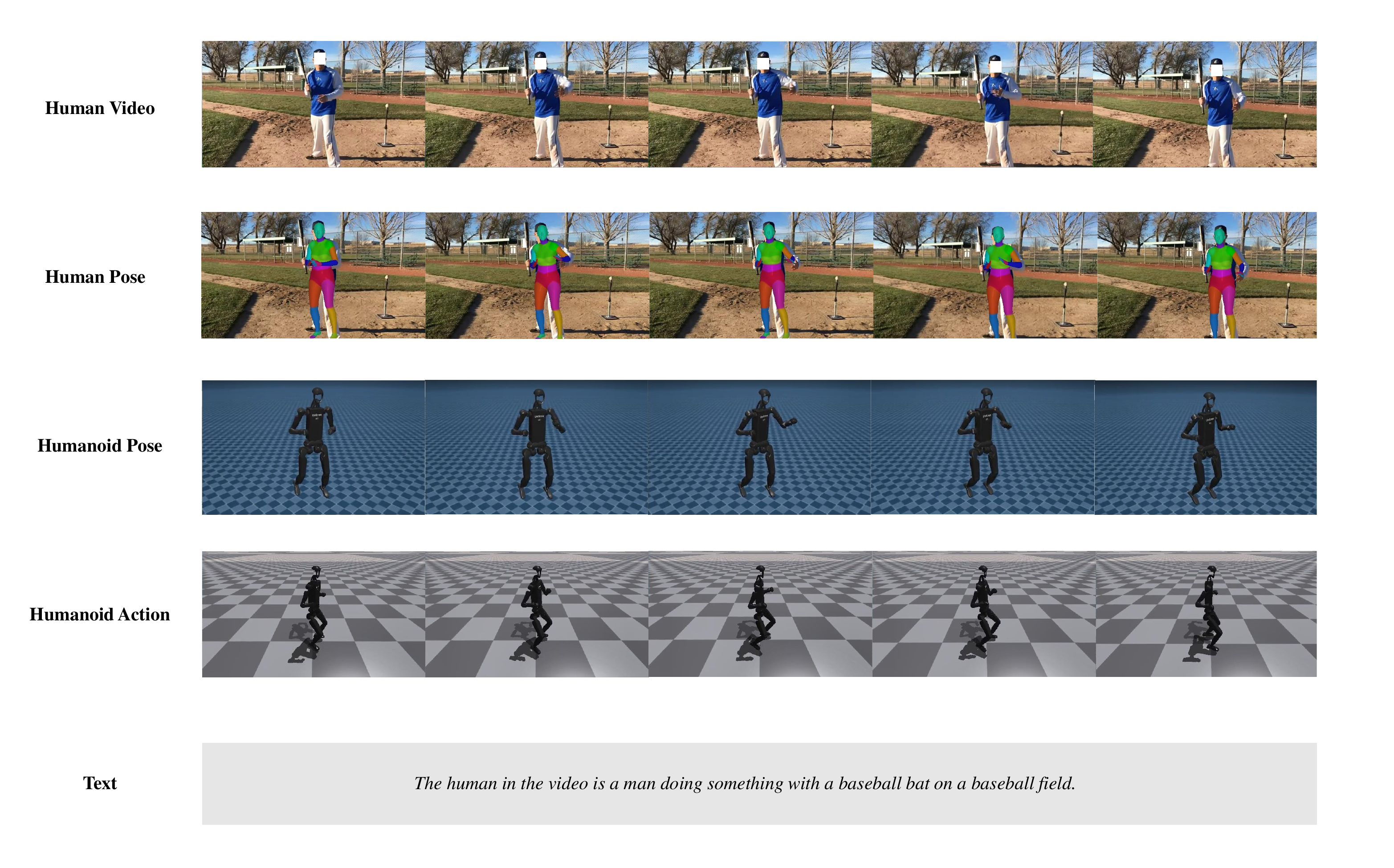}
   \caption{\textbf{Data examples in Humanoid-X}.}
   \label{fig:sample15}
\end{figure*}

\begin{figure*}[h]
  \centering
   \includegraphics[width=0.95\linewidth]{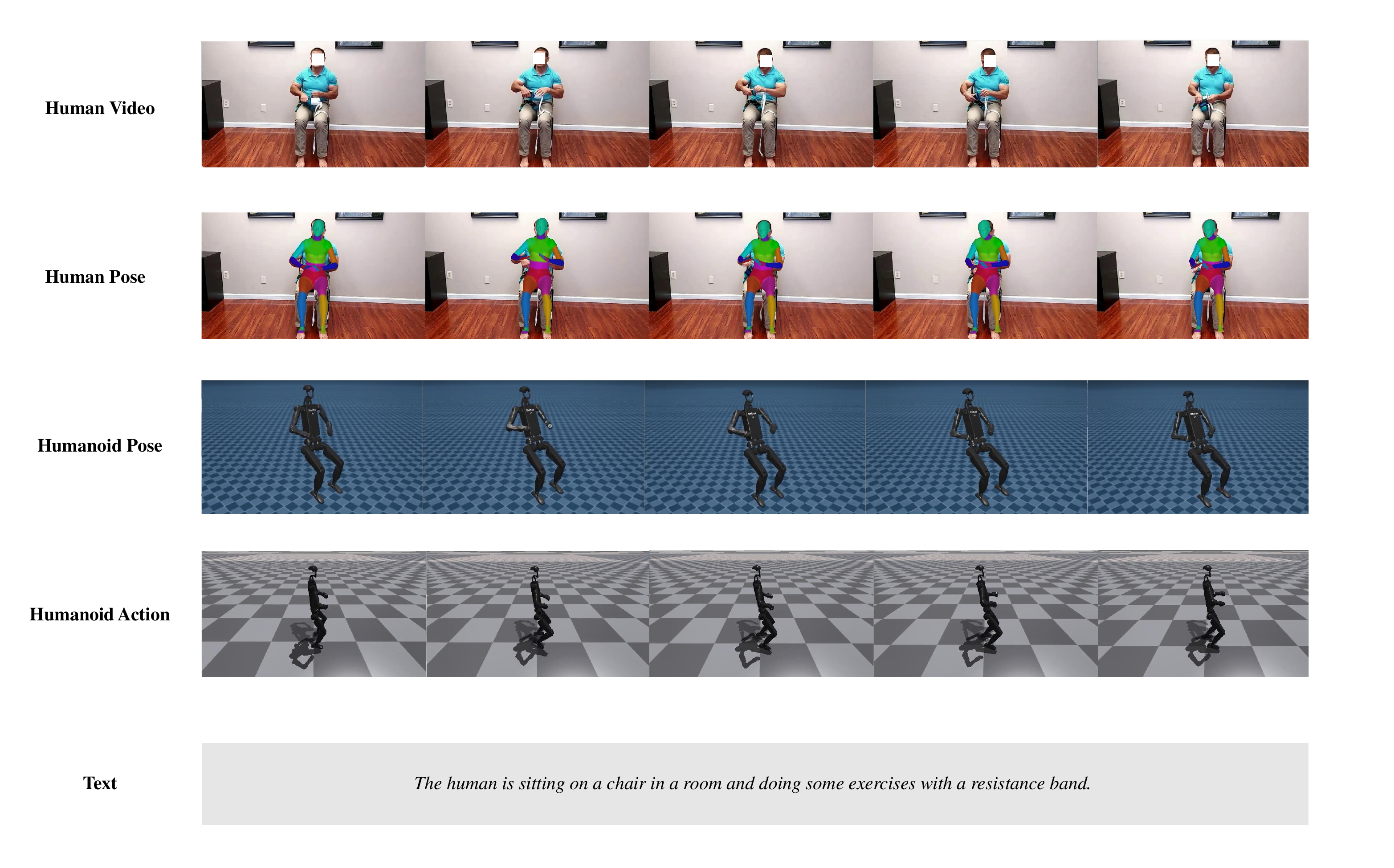}
   \caption{\textbf{Data examples in Humanoid-X}.}
   \label{fig:sample16}
\end{figure*}

\begin{figure*}[h]
  \centering
   \includegraphics[width=0.95\linewidth]{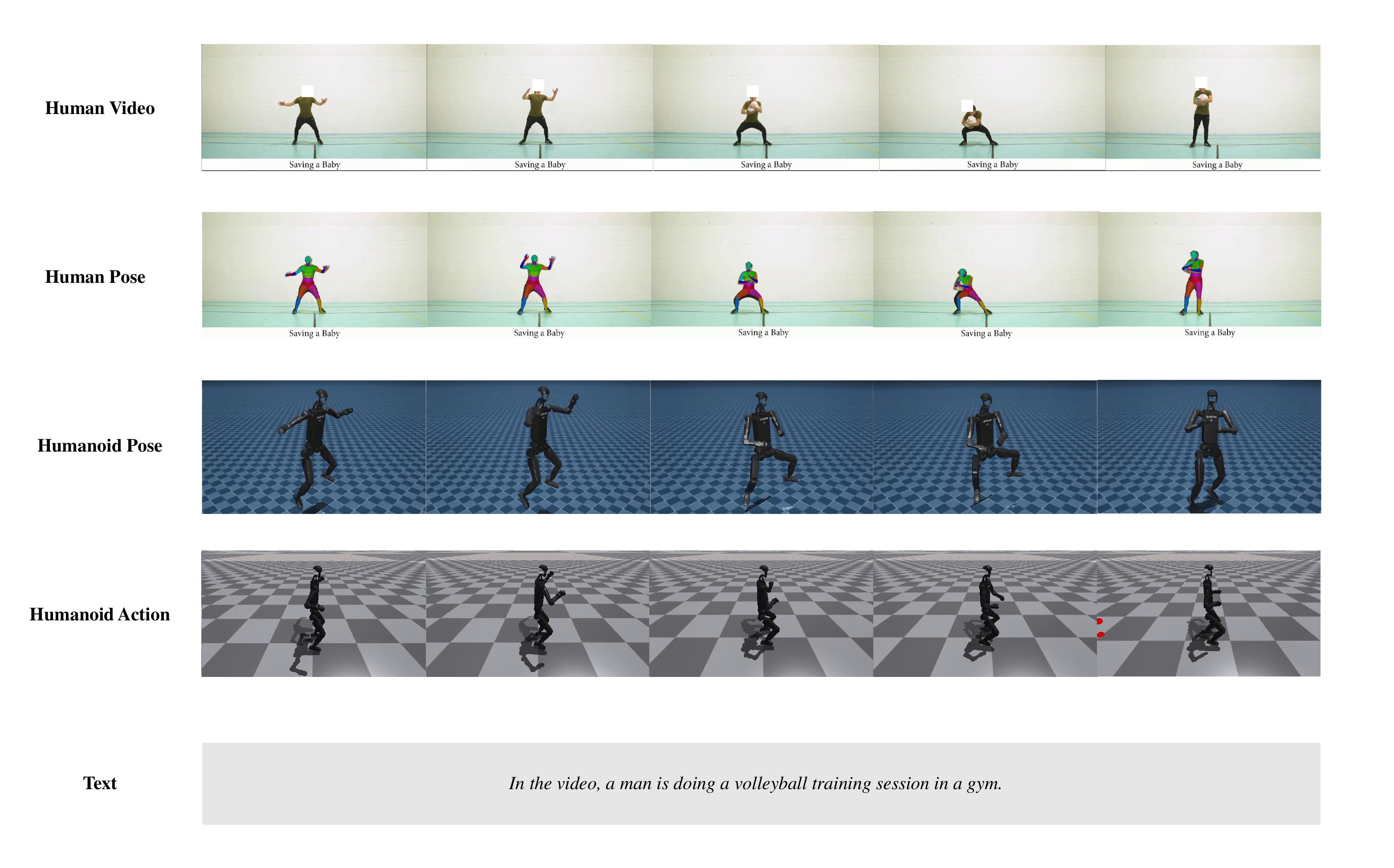}
   \caption{\textbf{Data examples in Humanoid-X}.}
   \label{fig:sample17}
\end{figure*}

\begin{figure*}[h]
  \centering
   \includegraphics[width=0.95\linewidth]{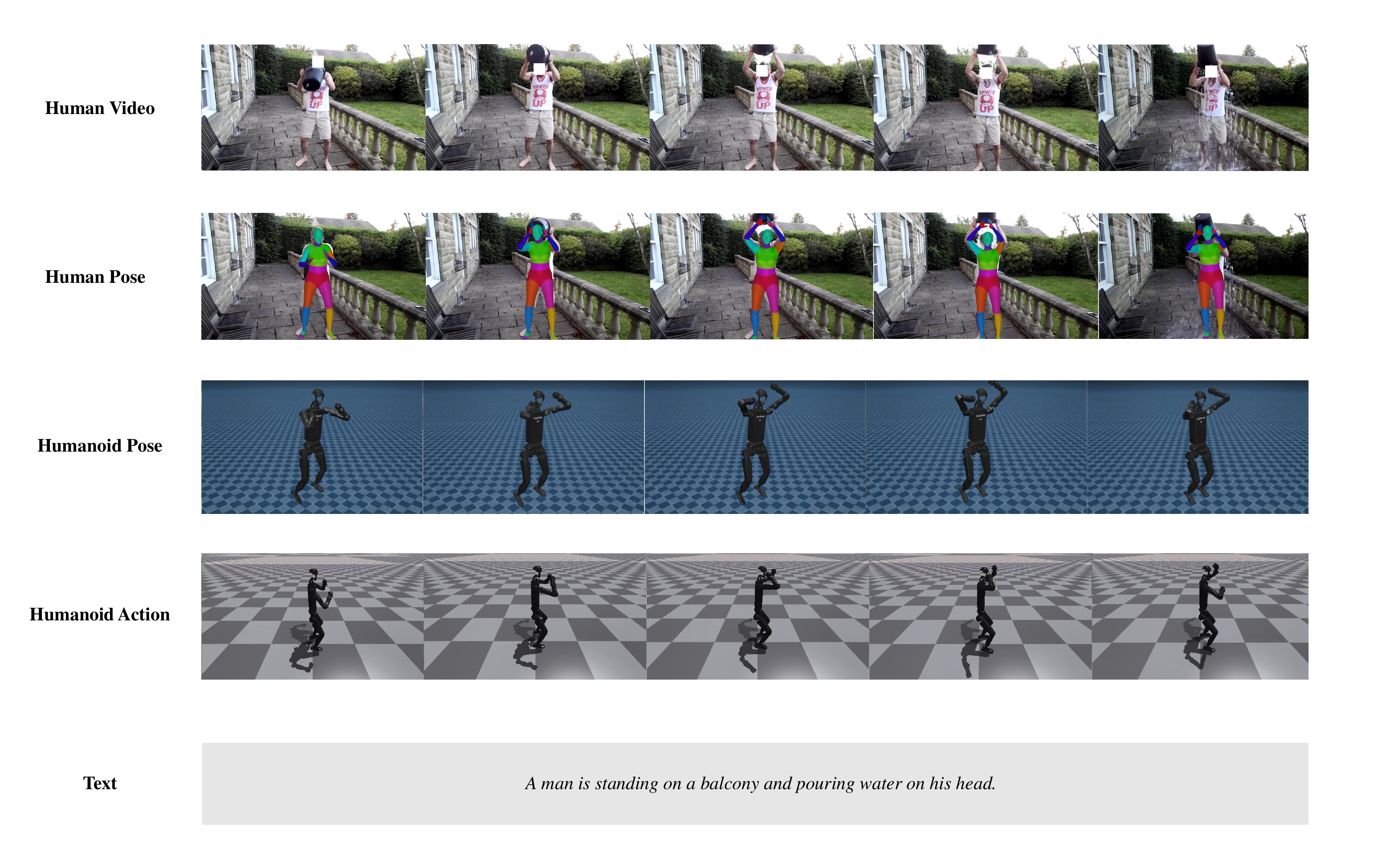}
   \caption{\textbf{Data examples in Humanoid-X}.}
   \label{fig:sample18}
\end{figure*}

\begin{figure*}[h]
  \centering
   \includegraphics[width=0.95\linewidth]{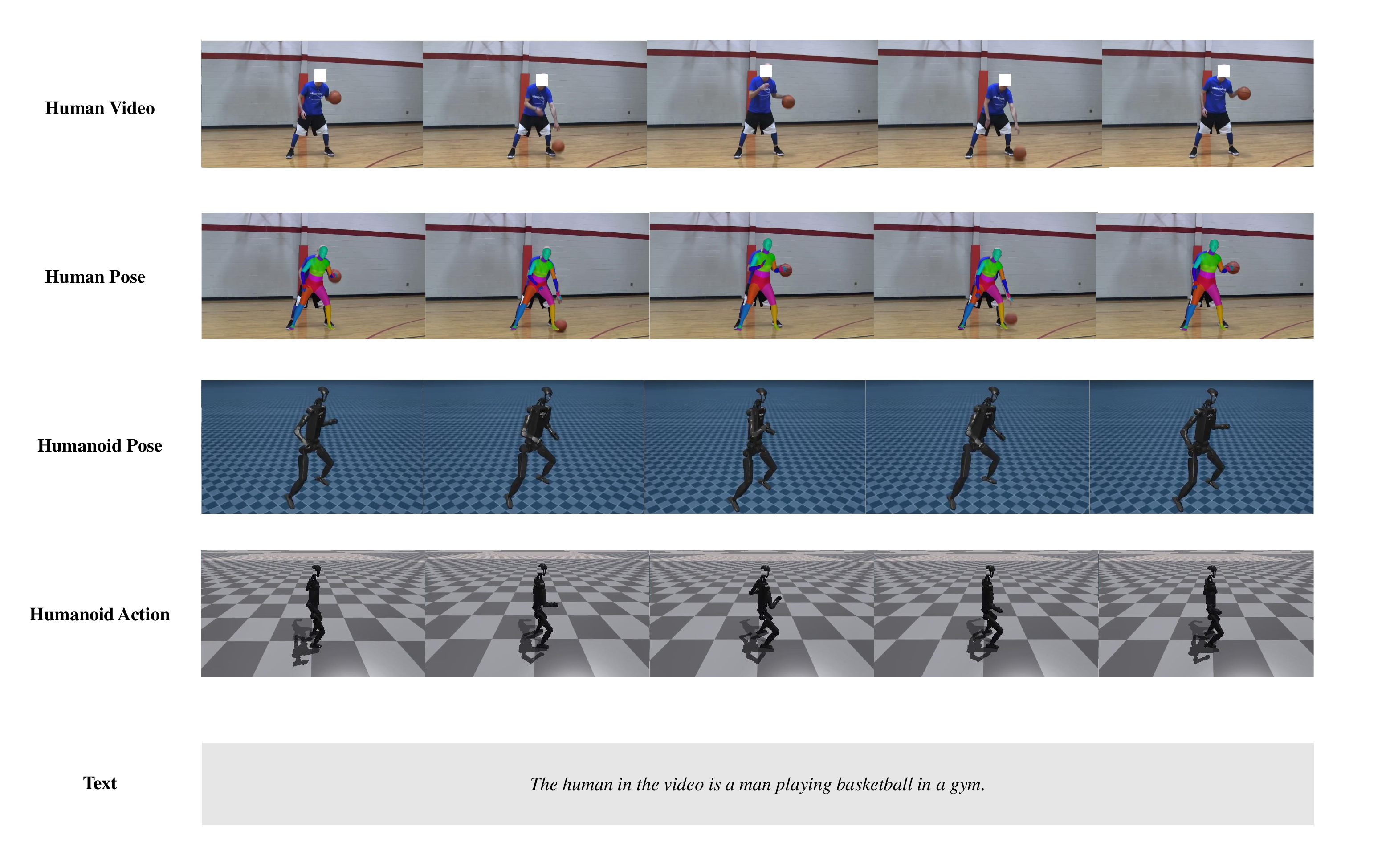}
   \caption{\textbf{Data examples in Humanoid-X}.}
   \label{fig:sample19}
\end{figure*}

\begin{figure*}[h]
  \centering
   \includegraphics[width=0.95\linewidth]{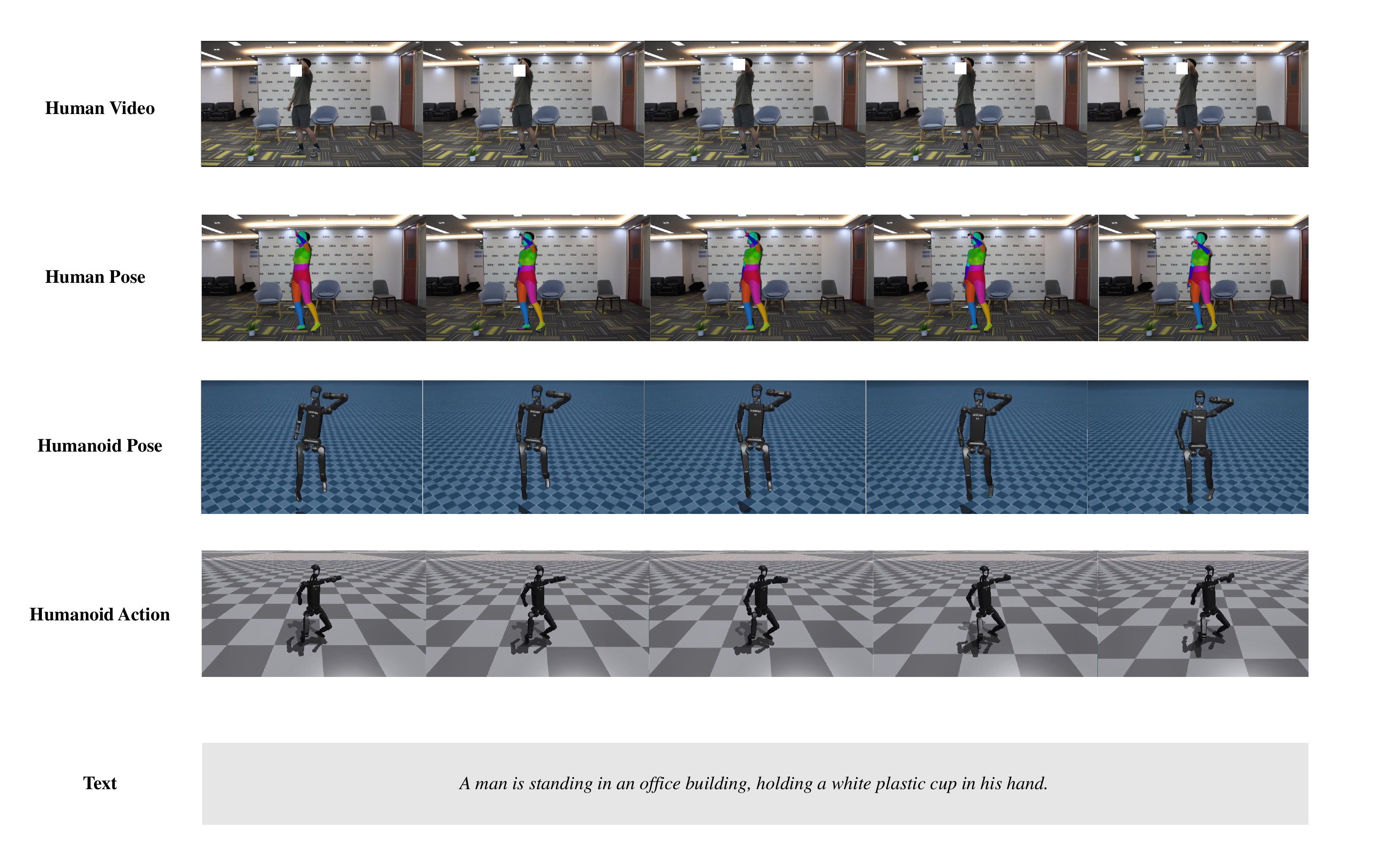}
   \caption{\textbf{Data examples in Humanoid-X}.}
   \label{fig:sample20}
\end{figure*}

\begin{figure*}[h]
  \centering
   \includegraphics[width=0.95\linewidth]{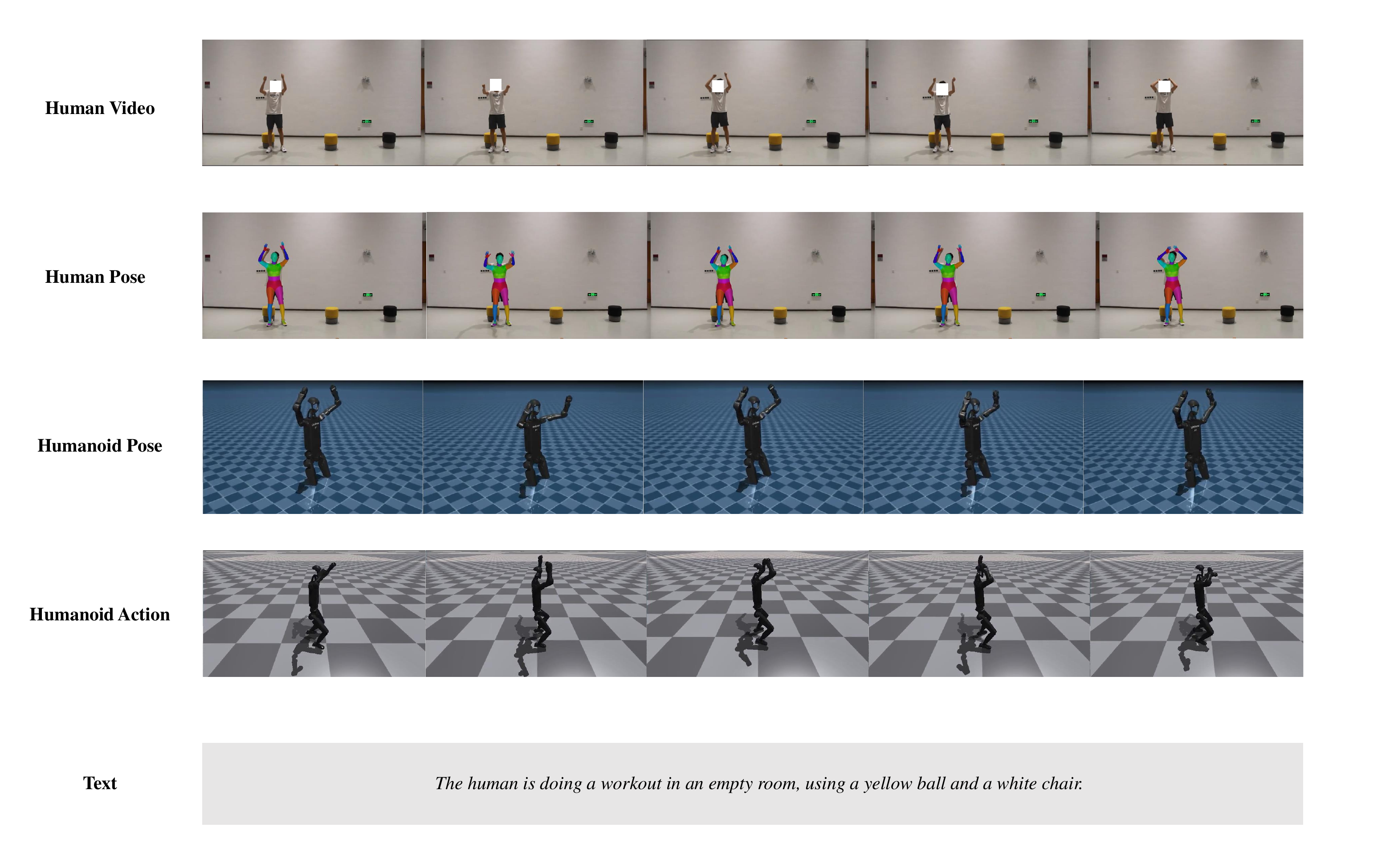}
   \caption{\textbf{Data examples in Humanoid-X}.}
   \label{fig:sample21}
\end{figure*}

\begin{figure*}[h]
  \centering
   \includegraphics[width=0.95\linewidth]{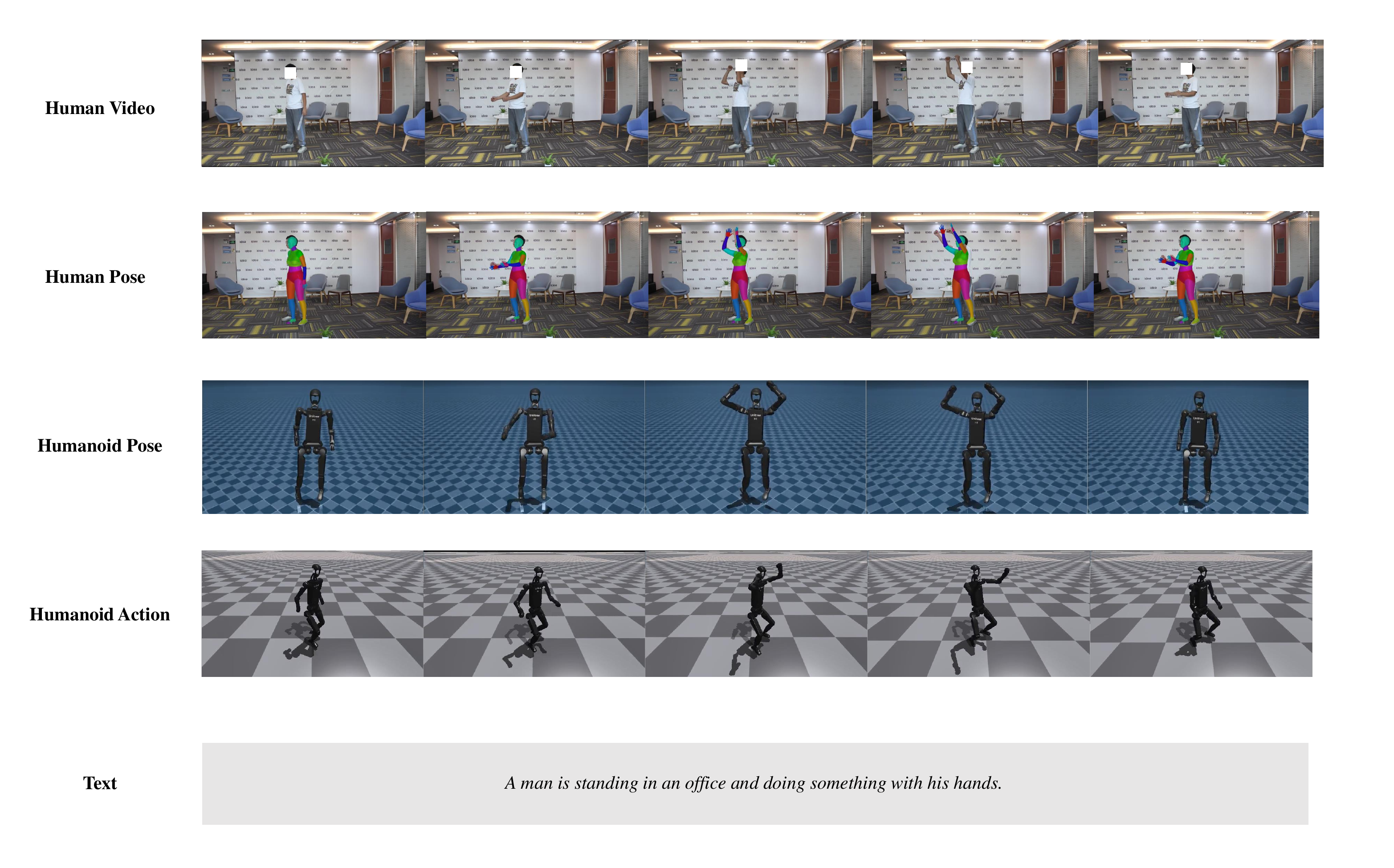}
   \caption{\textbf{Data examples in Humanoid-X}.}
   \label{fig:sample22}
\end{figure*}

We show visualized data examples from Humanoid-X in \cref{fig:sample1}, \cref{fig:sample2}, \cref{fig:sample3}, \cref{fig:sample4}, \cref{fig:sample5}, \cref{fig:sample6}, \cref{fig:sample7}, \cref{fig:sample8}, \cref{fig:sample9}, \cref{fig:sample10}, \cref{fig:sample11}, \cref{fig:sample12}, \cref{fig:sample13}, \cref{fig:sample14}, \cref{fig:sample15}, \cref{fig:sample16}, \cref{fig:sample17}, \cref{fig:sample18}, \cref{fig:sample19}, \cref{fig:sample20}, \cref{fig:sample21}, \cref{fig:sample22}.
For motion video clips, we sample 5 frames from each video clip shown.
For text, we directly present the text descriptions of the motions shown.
For the human pose, we sample the SMPL visualization of the corresponding frames in the video clip.
For the humanoid pose, we set the humanoid keypoints in MuJoCo and collected the MuJoCo-rendered images of the corresponding frames in the video clip.
For humanoid actions, we render the humanoid control policy rollout in IsaacGym and collect the rendered images of the corresponding frames in the video clip.

\section{Detailed UH-1 Model Architecture}

\subsection{UH-1 Action Tokenizer}
For a given input sequence $X = [x_1, x_2, \dots, x_T]$ with $x_t \in \mathbb{R}^{d_1}$ (representing either a humanoid keypoint or an action), UH-1 Action Tokenizer is designed to reconstruct this sequence using a learnable codebook $C = \{c_1, c_2, \dots, c_N\}$ with $c_n \in \mathbb{R}^{d_2}$ and a learnable autoencoder with an encoder $\mathbb{E}$ and a decoder $\mathbb{D}$. 
In this context, $T$ denotes the number of input frames, $N$ the codebook size, $d_1$ the dimension of input tokens, and $d_2$ the dimension of the codes. 
For sequence reconstruction, the encoder $\mathbb{E}$ maps the input sequence into latent representations $Z = \mathbb{E}(X) = [z_1, z_2, \dots, z_{T/k}]$ with $z_i \in \mathbb{R}^{d_2}$, where $k$ represents the temporal downsampling rate of the encoder. 
Each $z_i$ is subsequently quantized through the codebook into $\hat{z} \in C$ by selecting the nearest code $c_n \in C$, which can be formally expressed as:
\begin{equation}
 \hat{z} = \underset{c_n \in C}{\arg} \min \| z_i - c_n \|_2.
\end{equation}
Finally, the decoder $\mathbb{D}$ reconstructs the original input sequence as $X_{\text{re}} = \mathbb{D}(\hat{Z})$.
The temporal downsampling enables each code in the codebook to represent $k$ input tokens, encoding primitive humanoid motion skills and facilitating the generation of smooth actions.

In general, UH-1 Action Tokenizer is optimized by minimizing a standard objective function:
\begin{gather}
    \mathcal{L}_{\text{vqvae}} = \mathcal{L}_{\text{recon}} + \mathcal{L}_{\text{embed}} + \alpha \mathcal{L}_{\text{commit}}, \\
    \mathcal{L}_{\text{recon}} = \mathcal{L}_1(X, X_{\text{re}}), \\
    \mathcal{L}_{\text{embed}} = \| \text{sg}[Z] - \hat{Z} \|_2,
    \mathcal{L}_{\text{commit}} = \| Z - \text{sg}[\hat{Z}] \|_2. 
\end{gather}
In this formulation, $\alpha$ is a hyperparameter that regulates the relative influence of each loss term, and $\text{sg}[\cdot]$ denotes the stop gradient operator. The embedding loss $\mathcal{L}_{\text{embed}}$ promotes the quantized codebook embeddings to move closer to the continuous output of the encoder, while $\mathcal{L}_{\text{commit}}$ encourages the encoder to commit to particular codebook entries.

Given the unique properties of humanoid keypoints and actions, we propose an adjusted reconstruction loss, $\mathcal{L}_{\text{recon}}$, which integrates a forward difference loss and a root regularization term:
\begin{equation}
\mathcal{L}_1(X, X_{\text{re}}) + \beta \mathcal{L}_1(\Delta [X], \Delta [X_{\text{re}}]) + \gamma \mathcal{L}_1(X_{\text{re}}^\text{root}, \mathbf{0}),
\end{equation}
where $\beta$ and $\gamma$ are hyperparameters for balancing the additional loss components, and $\Delta[\cdot]$ represents the forward difference operator.

\subsection{UH-1 Transformer}
We formulate the language-conditioned humanoid keypoint or action generation tasks as auto-regressive prediction of the next codebook index. Formally, let $s_i \in \{1, 2, \cdots, N\} \cup \{\text{End}\}$ denote the current index to predict, $s_{1:i-1}$ represent the preceding context of indices, and $l$ the language instruction embedding encoded by CLIP~\cite{radford2021learning}. The UH-1 Transformer is then trained to model the conditional probability distribution $P(s_i| s_{1:i-1}, l)$. A special [End] token is incorporated into the indices set to signal the termination of sequence generation. For an input sequence $X = [x_1, x_2, \cdots, x_T]$, the encoder $\mathbb{E}$ and codebook $C$ of the UH-1 Action Tokenizer map this sequence into the codebook indices as $S = [s_1, s_2, \cdots, s_{T/k}, \text{End}]$; given this sequence of indices $S$, it can also be mapped back to $\hat{Z} = [c_{s_1}, c_{s_2}, \cdots, c_{s_{T/k}}]$, which is subsequently projected into the output space by the decoder $\mathbb{D}$ as $X_{\text{re}} = \mathbb{D}(\hat{Z})$.

To train this transformer model, we minimize the negative log-likelihood over the training dataset $\mathcal{D}$:
\begin{equation}
\mathcal{L}_{\text{trans}} = - \underset{S \in \mathcal{D}}{\sum} \log \underset{i = 1}{\overset{|S|}{\Pi}} p(s_i|s_{1:i-1}, l).
\end{equation}
This objective encourages accurate predictions of the next index in the context of previous indices and language instructions.

\subsection{Implementation Details}
The implementation of our model architecture follows previous work~\cite{t2m-gpt}.
For the UH-1 Action Tokenizer, we employ a straightforward convolutional architecture consisting of 1D convolutions, residual blocks, and ReLU activation functions. Temporal downsampling and upsampling are achieved using convolutions with a stride of 2 and nearest-neighbor interpolation, respectively. The codebook size is configured as $2048 \times 512$, with a downsampling rate $k = 4$. During training, action sequences are cropped to a temporal length of $T = 64$. For the UH-1 Transformer, it is based on an 18-layer transformer model featuring 16 attention heads and a dimensionality of $1,024$.

Training the UH-1 Action Tokenizer and the UH-1 Transformer on HumanoidML3D (a selected set of Humanoid-X) requires approximately 8 hours and 30 hours, respectively, on a single NVIDIA RTX$^{\text{TM}}$ 6000 Ada GPU, while training on the full set of Humanoid-X requires approximately 40 hours and 400 hours, respectively.

\section{Experiment Details}
\subsection{Real Robot Experiment}

\noindent\textbf{Success Rate.} The success rate of real robot pose control is evaluated using two criteria: (1) Stability: the humanoid robot must maintain stability while performing actions; any instance of falling or failing to maintain balance results in an unsuccessful trial. (2) Accuracy: the humanoid robot must accurately perform the desired actions based on text instructions. This is assessed by five human evaluators, and if the majority agree that the robot does not perform the actions correctly, the trial is considered unsuccessful.

\noindent\textbf{PD Controller.} The output actions $a$ of our model are the target DoF positions for controlling the humanoid robot. We use the PD control to transform actions $a$ into motor torques $\tau$, which can be represented as
\begin{equation}
    \tau = K_p (a - q) - K_d dq,
\end{equation}
where $K_p$ and $K_d$ are the proportional coefficients of the motor position and speed errors respectively, $q$ is the current angle position of the motor rotor, and $dq$ is the current rotor angular velocity of the motor rotor. We use the standard $K_p$ and $K_d$ provided in the official robot documents in our experiments. 

\begin{figure*}[h]
  \centering
   \includegraphics[width=0.95\linewidth]{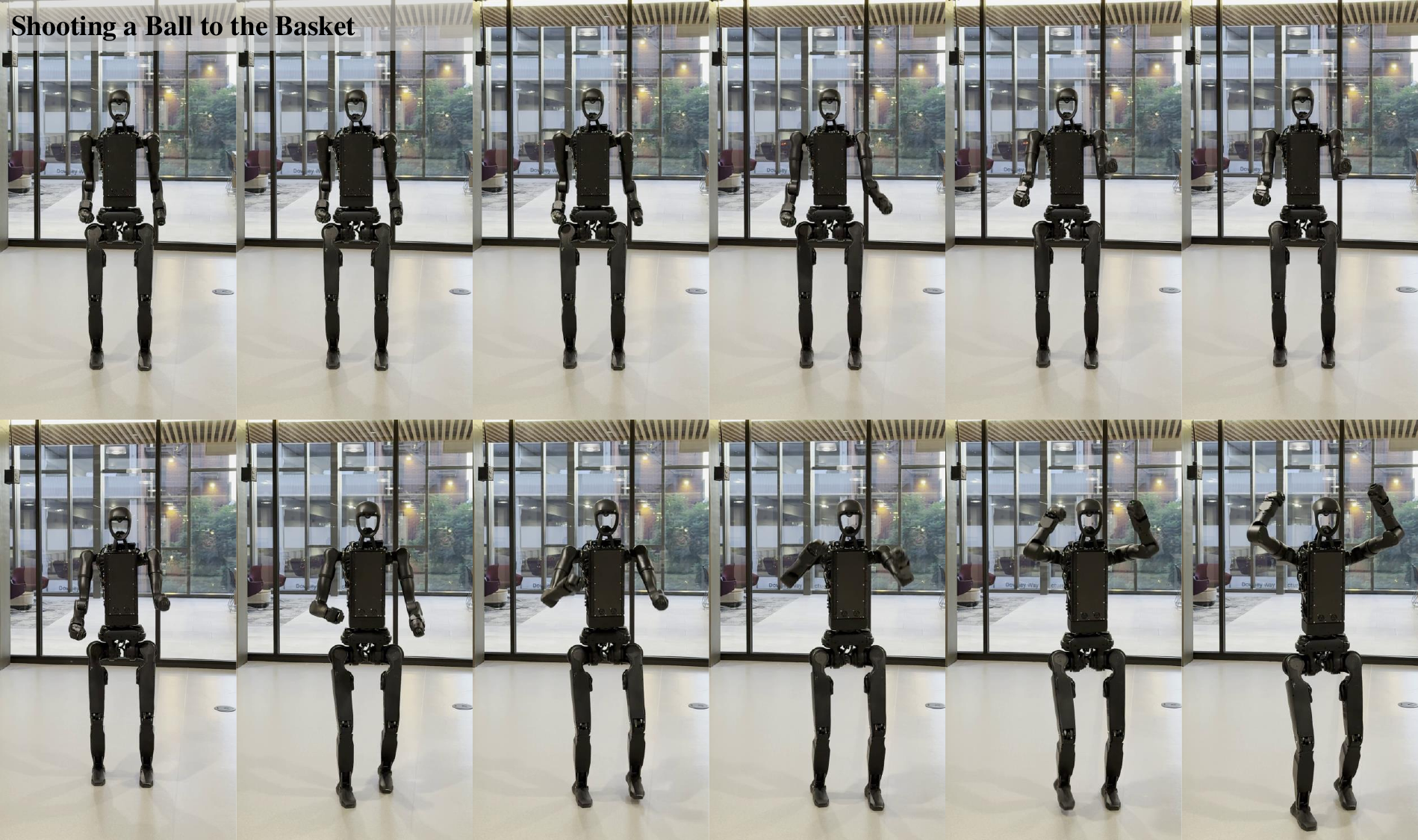}
   \caption{\textbf{Real robot demonstrations. Text instruction: \textit{Shooting a Ball to the Basket}}.}
   \label{fig:real1}
\end{figure*}

\begin{figure*}[h]
  \centering
   \includegraphics[width=0.95\linewidth]{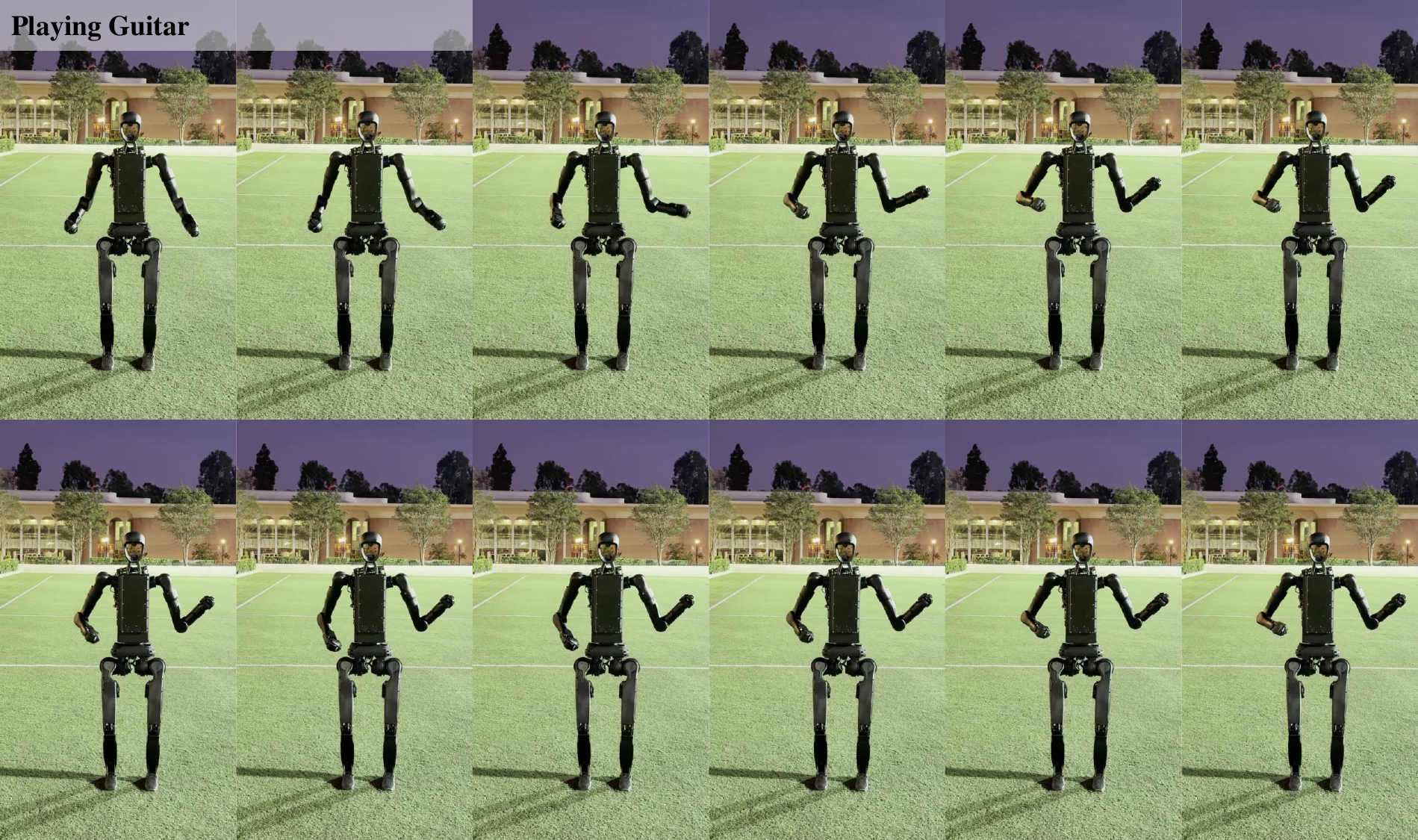}
   \caption{\textbf{Real robot demonstrations. Text instruction: \textit{Playing Guitar}}.}
   \label{fig:real2}
\end{figure*}

\begin{figure*}[h]
  \centering
   \includegraphics[width=0.95\linewidth]{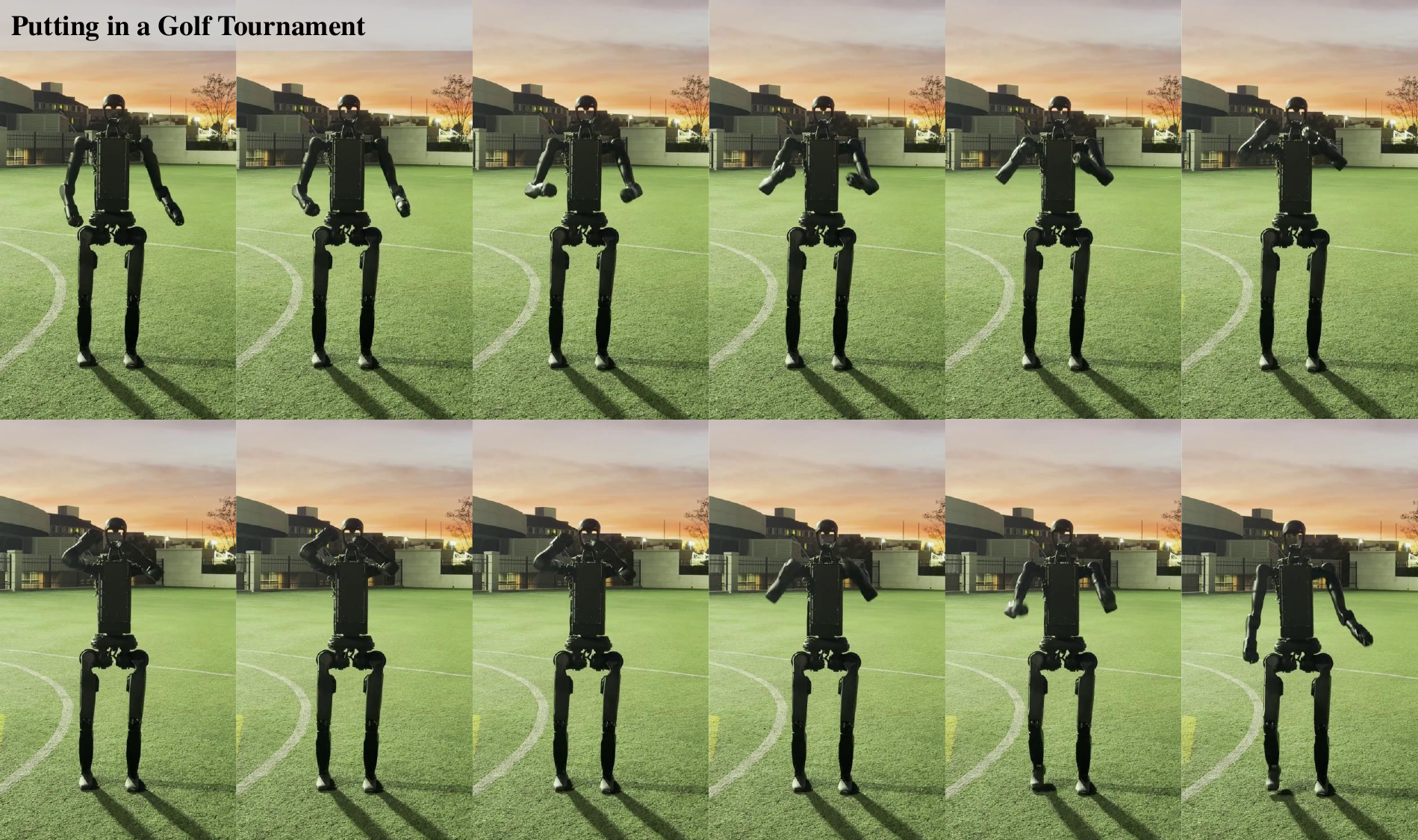}
   \caption{\textbf{Real robot demonstrations. Text instruction: \textit{Putting in a Golf Tournament}}.}
   \label{fig:real3}
\end{figure*}

\begin{figure*}[h]
  \centering
   \includegraphics[width=0.95\linewidth]{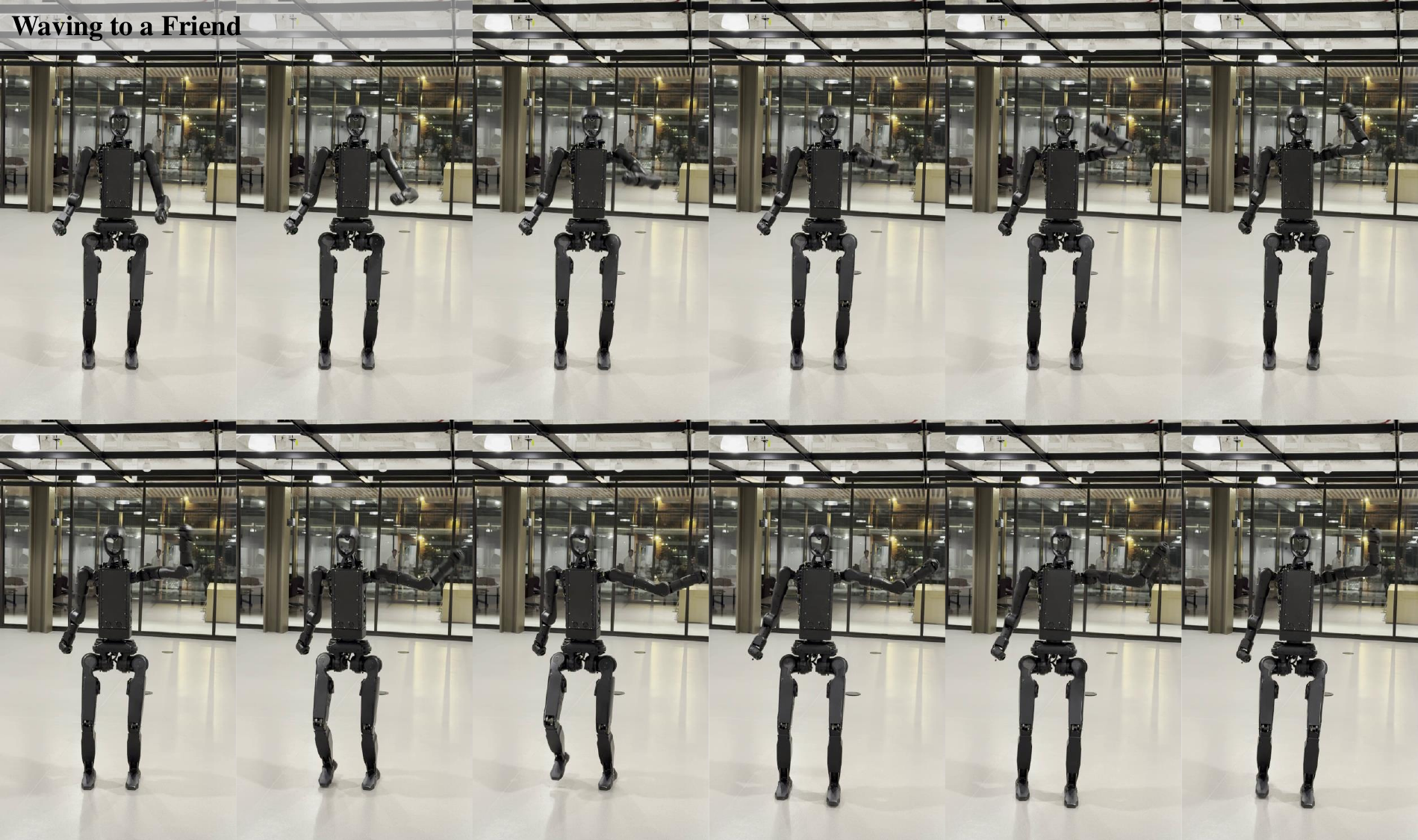}
   \caption{\textbf{Real robot demonstrations. Text instruction: \textit{Waving to a Friend}}.}
   \label{fig:real4}
\end{figure*}

\begin{figure*}[h]
  \centering
   \includegraphics[width=0.95\linewidth]{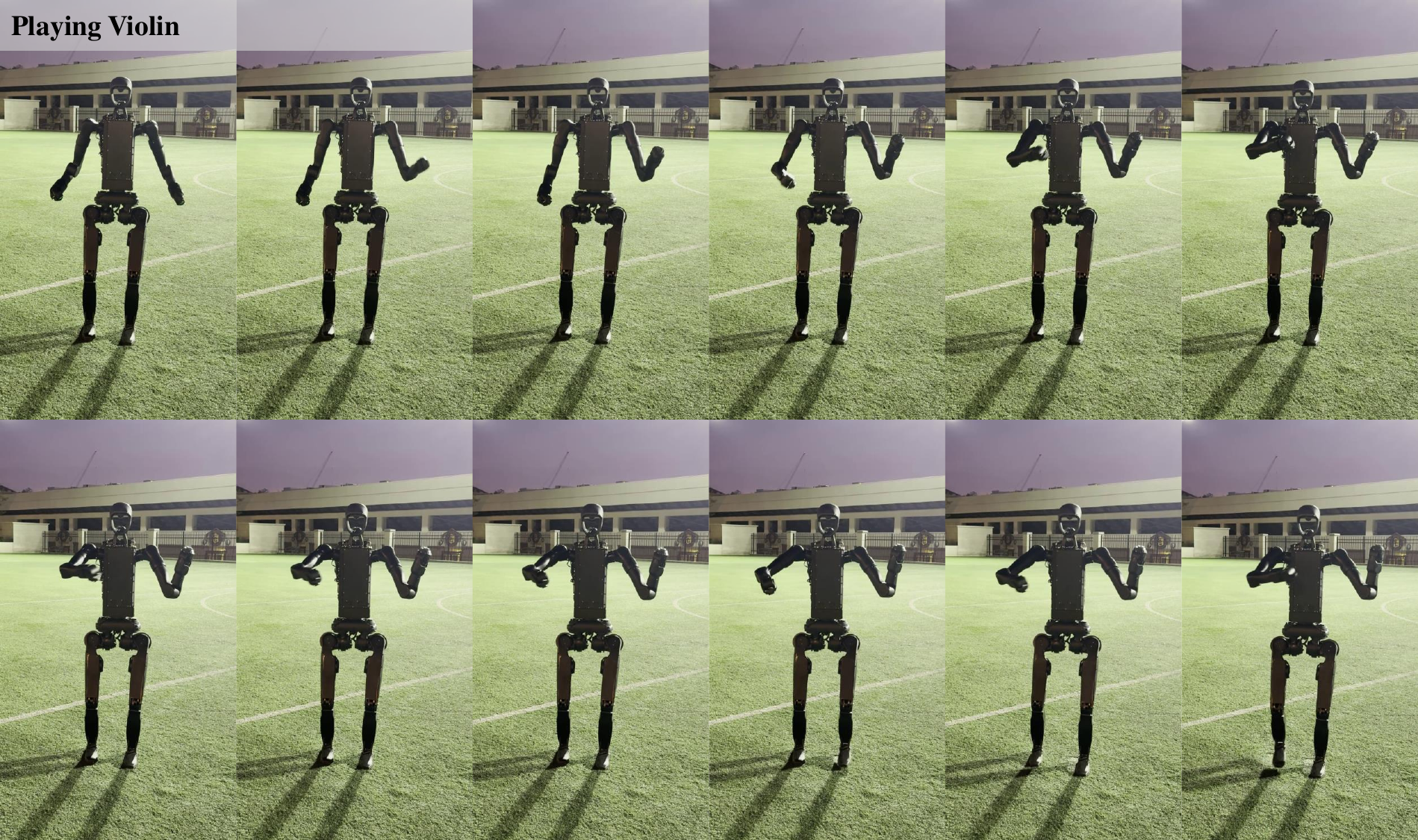}
   \caption{\textbf{Real robot demonstrations. Text instruction: \textit{Playing Violin}}.}
   \label{fig:real5}
\end{figure*}

\begin{figure*}[h]
  \centering
   \includegraphics[width=0.95\linewidth]{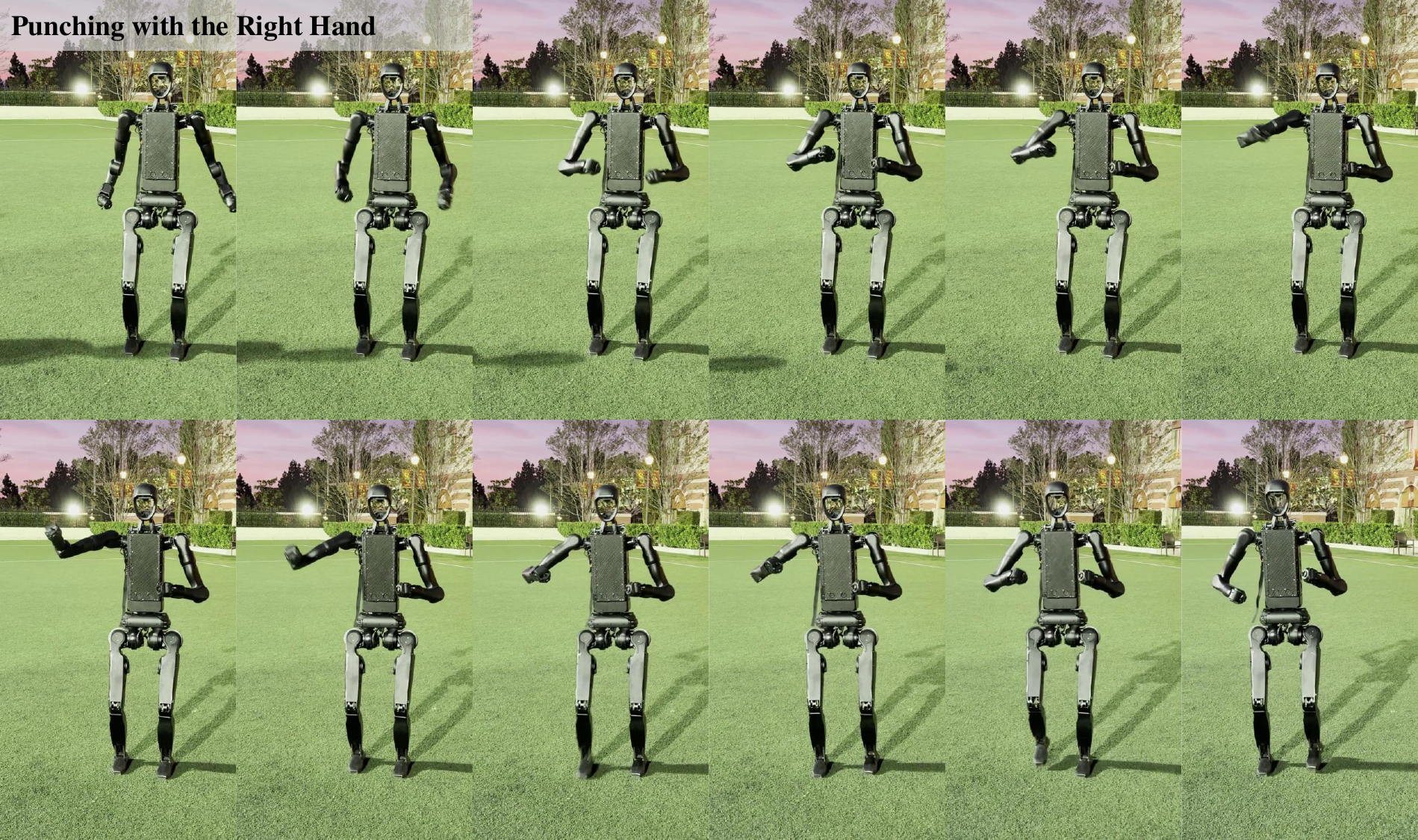}
   \caption{\textbf{Real robot demonstrations. Text instruction: \textit{Punching with the Right Hand}}.}
   \label{fig:real6}
\end{figure*}

\begin{figure*}[h]
  \centering
   \includegraphics[width=0.95\linewidth]{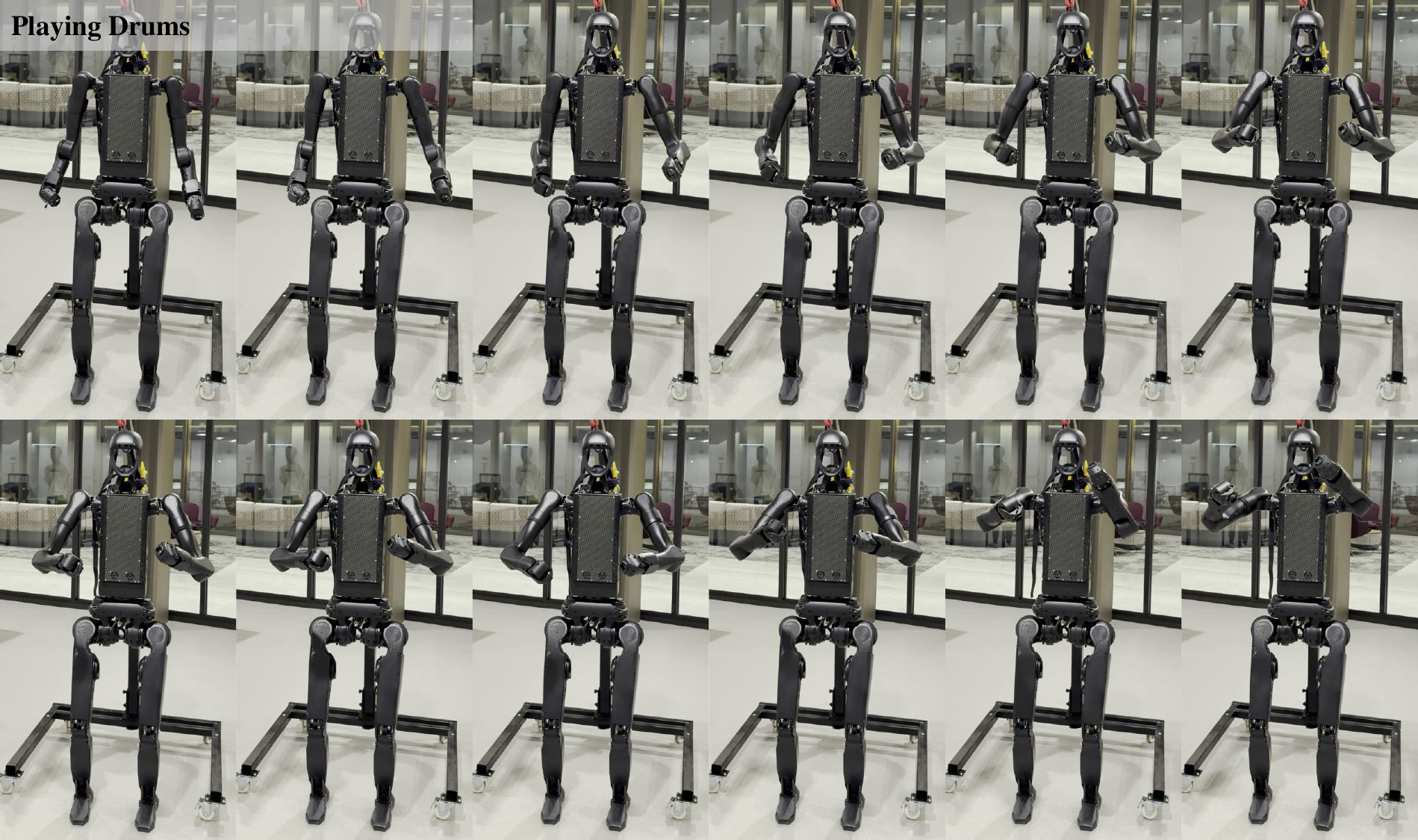}
   \caption{\textbf{Real robot demonstrations. Text instruction: \textit{Playing Drums}}.}
   \label{fig:real7}
\end{figure*}

\begin{figure*}[h]
  \centering
   \includegraphics[width=0.95\linewidth]{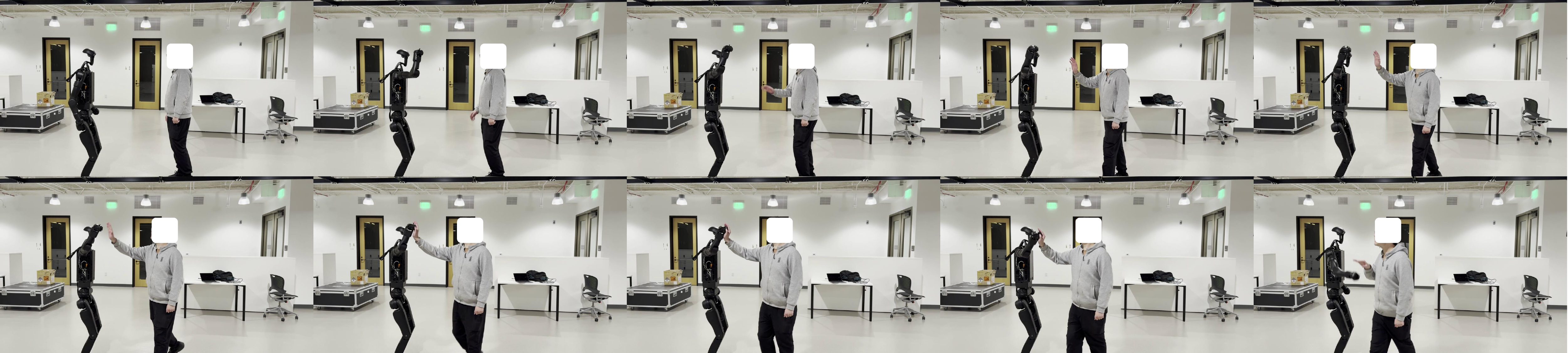}
   \caption{\textbf{Demonstration of human-humanoid interactions. Text instruction: \textit{High-Five}}.}
   \label{fig:real8}
\end{figure*}

\begin{figure*}[h]
  \centering
   \includegraphics[width=0.95\linewidth]{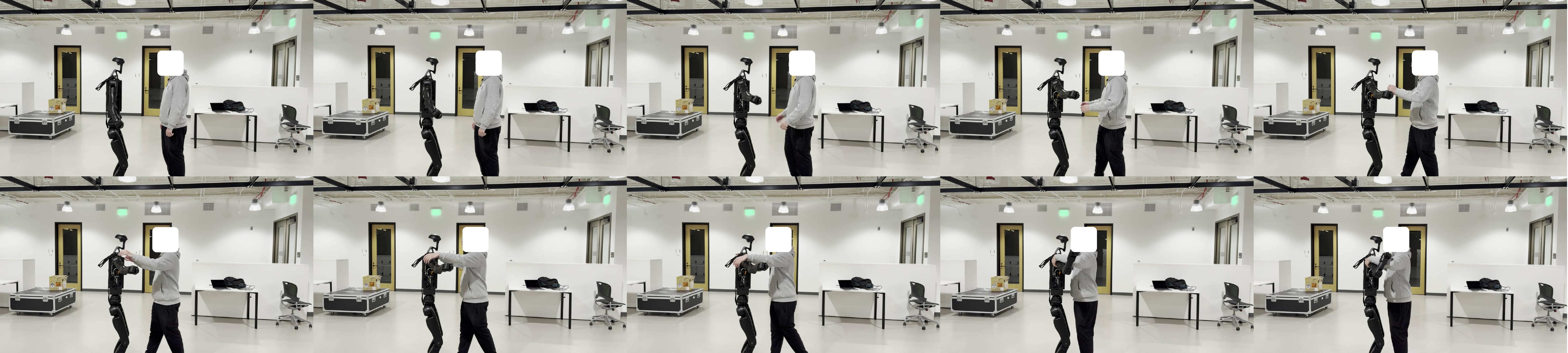}
   \caption{\textbf{Demonstration of human-humanoid interactions. Text instruction: \textit{Embrace}}.}
   \label{fig:real9}
\end{figure*}

\noindent\textbf{Real Robot Experiments.} We demonstrate the real humanoid robot pose control with text instructions in \cref{fig:real1}, \cref{fig:real2}, \cref{fig:real3}, \cref{fig:real4}, \cref{fig:real5}, \cref{fig:real6}, \cref{fig:real7}. We also demonstrate human-humanoid interactions in \cref{fig:real8} and \cref{fig:real9}. From these figures, we show that our method generates accurate and diverse poses to control the real humanoid robot with text instructions. 

\subsection{Ablation on Goal-conditioned Control Policy}
\begin{figure}[t]
  \centering
   \includegraphics[width=1.0\linewidth]{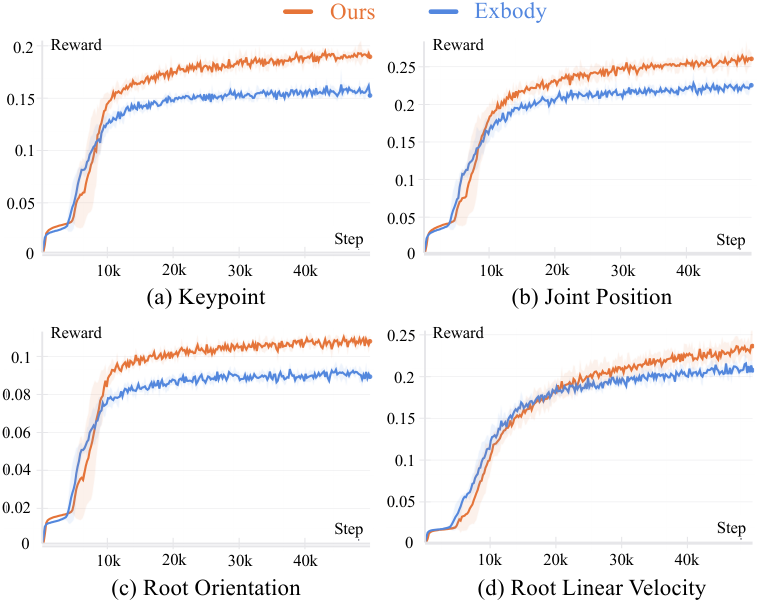}
   \captionof{figure}{\textbf{Ablation on different RL policies}, measured by task cumulative reward value. The solid line represents the mean return value, while the shaded regions correspond to the standard deviation, both calculated across five different random seeds. Our retargeted training data enhances the performance of the RL policy in tracking the imitation of body keypoints, joint positions, root orientation and root linear velocity.}
   \label{fig:rl_ablation}
\end{figure}

To investigate the impact of humanoid keypoints on the goal-conditioned RL policy, we compare our motion retargeting approach, originated from~\cite{omnih2o}, with another approach in~\cite{expressive}. We evaluate the quality of the humanoid keypoints generated by different motion retargeting methods by measuring the tracking rewards in the subsequent reinforcement learning step, maintaining other factors as the same. As illustrated in \cref{fig:rl_ablation}, we launch experiments in five random seeds for both methods.
We empirically found that our motion retargeting method improves the performance of the RL policy on the evaluation metrics in~\cite{expressive} tracking the imitation of body keypoints, joint positions, root orientation and root linear velocity in the form of training rewards.
The results show that our retargeted data enhances the performance of the RL policy, thus suggesting that our retargeting method can generate humanoid pose data more executable for humanoid robots.

\end{document}